\documentclass{article}

\PassOptionsToPackage{numbers, compress}{natbib}

\usepackage[preprint]{neurips_2025}

\usepackage[utf8]{inputenc} %
\usepackage[T1]{fontenc}    %
\usepackage{booktabs}       %
\usepackage{amsfonts}       %
\usepackage{nicefrac}       %
\usepackage{microtype}      %

\usepackage{graphicx}

\usepackage{amssymb, mathtools, amsmath}

\usepackage{multirow}

\usepackage{hyperref}
\usepackage{url}
\usepackage{amsfonts}       %
\usepackage{nicefrac}       %
\usepackage{microtype}      %
\usepackage{xcolor}         %

\usepackage[small,bf]{caption}

\usepackage{rotating}

\usepackage{wrapfig}
\usepackage{multicol}

\usepackage{algorithm}
\usepackage{algpseudocode}

\usepackage{adjustbox}

\usepackage{subcaption}
\usepackage{pifont}
\usepackage{framed}
\usepackage{fancyvrb}
\usepackage{xcolor}
\usepackage{bbm}

\hypersetup{
    colorlinks = true,
    citecolor=blue,
    linkcolor=blue,
    urlcolor=blue
}
\usepackage{amsthm}
\usepackage{listings}

\definecolor{darkgreen}{rgb}{0.0, 0.5, 0.0}
\definecolor{blue}{rgb}{0.0, 0.47, 0.75}
\definecolor{dartmouthgreen}{rgb}{0.05, 0.5, 0.06}
\definecolor{drab}{rgb}{0.59, 0.44, 0.09}
\definecolor{navyblue}{rgb}{0.0, 0.0, 0.5}

\definecolor{darkgreen}{rgb}{0.0, 0.5, 0.0}
\definecolor{blue}{rgb}{0.0, 0.47, 0.75}
\definecolor{dartmouthgreen}{rgb}{0.05, 0.5, 0.06}
\definecolor{drab}{rgb}{0.59, 0.44, 0.09}
\definecolor{navyblue}{rgb}{0.0, 0.0, 0.5}

\newcommand{\tb}[1]{\textbf{#1}}

\newcommand{\vc}{\mathbf{c}}

\newcommand{\vg}{\mathbf{g}}

\newcommand{\vx}{\mathbf{x}}
\newcommand{\vy}{\mathbf{y}}
\newcommand{\vz}{\mathbf{z}}

\newcommand{\vZ}{\mathbf{Z}}

\newcommand{\bE}{\mathbb{E}}

\newcommand{\bH}{\mathbb{H}}

\newcommand{\cmark}{ {\color{dartmouthgreen} \ding{51}} }%
\newcommand{\xmark}{ {\color{red} \ding{55}} }%

\usepackage{xcolor,colortbl}

\definecolor{Gray}{gray}{0.5}
\definecolor{LightCyan}{rgb}{0.88,1,1}

\newcolumntype{a}{>{\columncolor{Gray}}c}
\newcolumntype{b}{>{\columncolor{white}}c}

\usepackage[utf8]{inputenc}	%
\usepackage[T1]{fontenc}	%

\usepackage{graphicx}%
\usepackage{multirow}%
\usepackage{amsmath,amssymb,amsfonts}%
\usepackage{amsthm}%
\usepackage{mathrsfs}%
\usepackage[title]{appendix}%
\usepackage{xcolor}%
\usepackage{textcomp}%
\usepackage{manyfoot}%
\usepackage{booktabs}%
\usepackage{algorithm}%
\usepackage{algorithmicx}%
\usepackage{algpseudocode}%
\usepackage{listings}%
\usepackage[listings,theorems,breakable,skins]{tcolorbox}
\usepackage{xltabular}
\usepackage{soul}
\usepackage{makecell}
\usepackage{wasysym}
\usepackage{subcaption}

\usepackage[shortlabels,inline]{enumitem}

\definecolor{aiyellow}{HTML}{ebbe4d}
\definecolor{aigreen}{RGB}{210,244,211} 
\sethlcolor{aigreen}
\newcommand{\hle}[2][aiyellow]{{%
    \colorlet{foo}{#1}%
    \sethlcolor{foo}\hl{#2}}%
}

\theoremstyle{thmstyleone}%

\theoremstyle{thmstyletwo}%

\theoremstyle{thmstylethree}%

\definecolor{darkColor}{HTML}{D4E4E0}
\definecolor{lightGray}{HTML}{F9F6EE}

\newtcolorbox[auto counter,number within=subsection]{contextbox}[2][]{%
enhanced,
colback=white,
colframe=darkColor,
fonttitle=\bfseries\fontfamily{phv}\selectfont\color{black},
size=small, boxsep=0mm, top=1mm, boxrule=0.5mm, left=0mm, right=0mm,
title=,#1}

\NewTColorBox[]{promptbox}{O{}mO{VLM response:}}{%
enhanced,frame hidden,%
colback=white,
coltitle=black,
skin=bicolor, size=small,
colbacklower=lightGray, boxrule=0mm,%
before lower={\parbox[][5pt][t]{\linewidth}{\textit{\textbf{#3}}}},%
  fontupper=\fontsize{6}{7.2}\selectfont\fontfamily{phv}\selectfont, %
  #1,%
  fontlower=\fontsize{6}{7.2}\selectfont\fontfamily{phv}\selectfont
}

\newtcolorbox[auto counter,number within=subsection]{examplebox}[2][]{%
enhanced,
colback=white,
colframe=darkColor,
fonttitle=\bfseries\fontfamily{phv}\selectfont\color{black},
size=small, boxsep=0mm, top=1mm, boxrule=0.5mm, left=0mm, right=0mm,
title=,#1}

\NewTColorBox[]{vlmbox}{O{}mO{VLM response w/o \texttt{SVP}:}}{%
enhanced,frame hidden,%
colback=white,
coltitle=black,
skin=bicolor, size=small,
colbacklower=lightGray, boxrule=0mm,%
before lower={\parbox[][5pt][t]{\linewidth}{\textit{\textbf{#3}}}},%
  fontupper=\fontsize{6}{7.2}\selectfont\fontfamily{phv}\selectfont, %
  #1,%
  fontlower=\fontsize{6}{7.2}\selectfont\fontfamily{phv}\selectfont
}

\NewTColorBox[]{svpbox}{O{}mO{VLM response w/ \texttt{SVP}:}}{%
enhanced,frame hidden,%
colback=white,
coltitle=black,
skin=bicolor, size=small,
colbacklower=lightGray, boxrule=0mm,%
before lower={\parbox[][5pt][t]{\linewidth}{\textit{\textbf{#3}}}},%
  fontupper=\fontsize{6}{7.2}\selectfont\fontfamily{phv}\selectfont, %
  #1,%
  fontlower=\fontsize{6}{7.2}\selectfont\fontfamily{phv}\selectfont
}

\title{Feedback-Driven Vision-Language Alignment with Minimal Human Supervision}

\author{%
Giorgio Giannone\\
Amazon\\
\And
Ruoteng Li\\
Amazon\\
\And 
Qianli Feng\\
Amazon\\
\AND
Evgeny Perevodchikov\\
Amazon\\
\And
Rui Chen\\
Amazon\\
\And 
Aleix Martinez\\
Amazon\\
}

\begin{document}

\maketitle
\begin{abstract}
Vision-language models (VLMs) have demonstrated remarkable potential in integrating visual and linguistic information, but their performance is often constrained by the need for extensive, high-quality image-text training data. 
Curation of these image-text pairs is both time-consuming and computationally expensive.
To address this challenge, we introduce \texttt{SVP} (Sampling-based Visual Projection), a novel framework that enhances vision-language alignment without relying on manually curated text-image pairs or preference annotation. SVP leverages a small set of manually selected images, self-captioning and a pre-trained grounding model as a feedback mechanism to elicit latent information in VLMs. We evaluate our approach across six key areas: captioning, referring, visual question answering, multitasking, hallucination control, and object recall. Results demonstrate significant improvements, including a 14\% average improvement in captioning tasks, up to 12\% increase in object recall,  and significantly reduced hallucinations, while maintaining question-answering capabilities.
Using \texttt{SVP}, a small VLM achieves hallucination reductions similar to a model five times larger, while a VLM with initially poor referring capabilities more than doubles its performance, approaching parity with a model twice its size.
\end{abstract}

\section{Introduction}
\begin{wrapfigure}{r}{0.5\textwidth}
\vspace{-2cm}
    \centering
    \includegraphics[width=\linewidth]{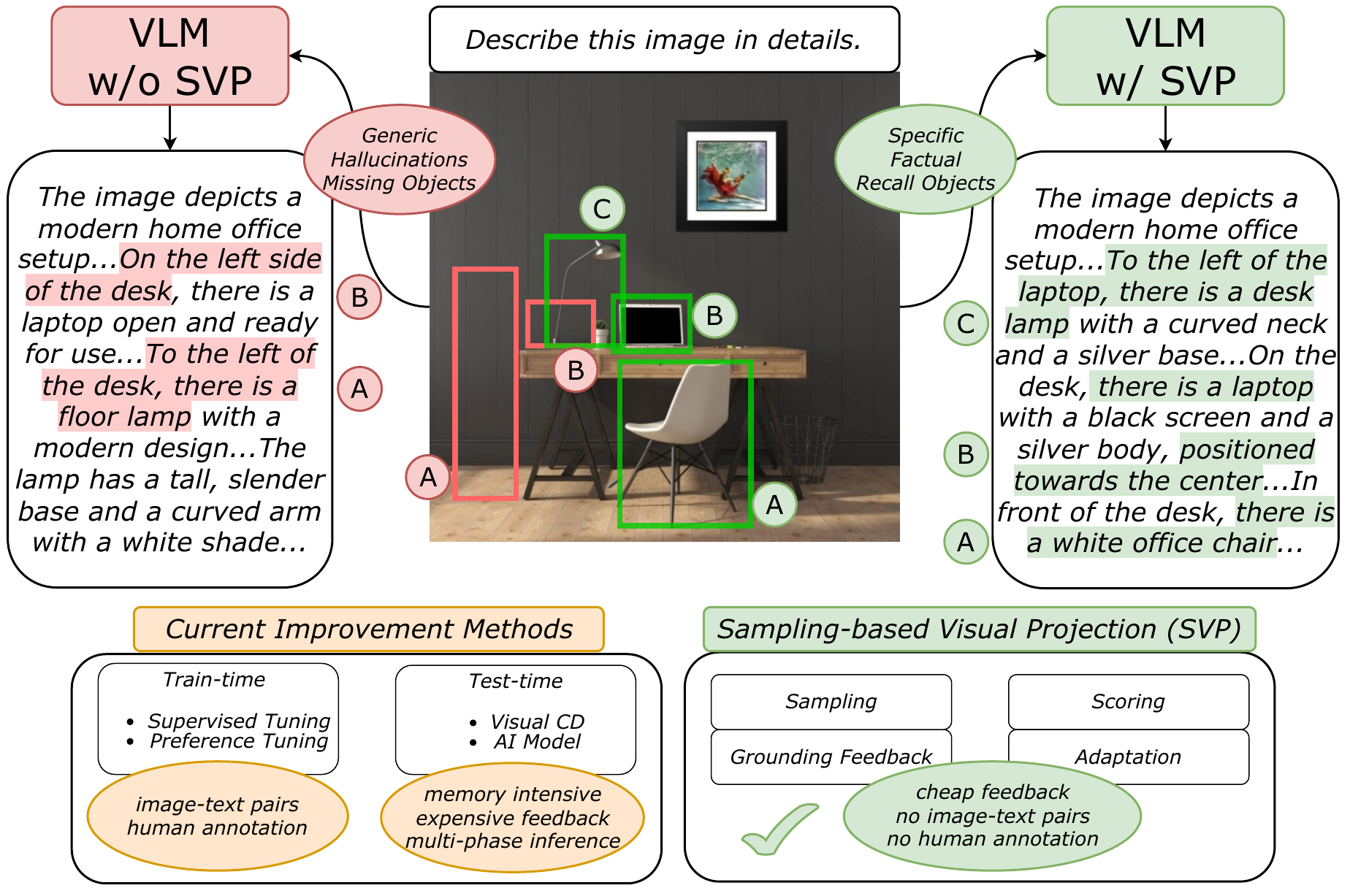}
    \caption{
    \textbf{Improving Vision-Language Alignment}. Vision-language models (VLMs) often produce descriptions lacking specificity and accuracy, frequently hallucinating objects or missing important elements (left). Our \emph{Sampling-based Visual Projection} (\texttt{SVP}) addresses these issues by leveraging self-captioning and grounding feedback. \texttt{SVP} enhances visual-language alignment without requiring human annotations, curated image-text pairs, or expensive AI feedback (right). This leads to models with greater contextual relevance, fewer hallucinations, and enhanced object recall. See Appx~\ref{fig:intro-figure} for details.
    }
    \label{fig:vp}
\vspace{-2cm}
\end{wrapfigure}
Vision-Language Models (VLMs~\cite{bordes2024introduction,zhang2024mm}) are essential to deploying expert level artificial intelligence, as human intelligence is predominantly multimodal.

Generative VLMs~\cite{li2024llava, li2022blip, ye2024mplug, chen2024far} built upon Large Language Models (LLMs) have shown great promises in zero-shot abilities on various downstream vision-linguistic tasks (Fig.~\ref{fig:main-loop-fig}.$(iv)$), unlocking new multimodal capacities and providing powerful generalization to specialized machine learning models. 
By learning a mapping between linguistic tokens and visual features, such VLMs enjoy the strong generation capabilities of LLMs~\cite{brown2020language,touvron2023llama} and the understanding of the physical world of computer vision models~\cite{radford2021learning,dosovitskiy2020image}.

However, VLMs derived from pretrained backbones are known to be impacted by the hallucinations and biases from LLMs~\cite{sasse2024mapping,rahmanzadehgervi2024vision}. 
It is frequently observed that these VLMs fail to produce text consistent with the visual content (left side Fig.~\ref{fig:vp}), i.e., the generated text describes entities not present in the input image or misses relevant entities altogether, generating content not grounded in the visual input~\cite{collerton2023understanding,bai2024hallucination}.
Addressing these shortcomings is crucial for future deployment of VLMs in high-stakes, real-world applications across the frontiers of scientific discovery~\cite{he2024foundation} and engineering~\cite{picard2023concept,song2024multi}.

\begin{figure}[ht]
    \centering
    \begin{minipage}[b]{\textwidth}
        \centering
        \includegraphics[width=.16\textwidth]{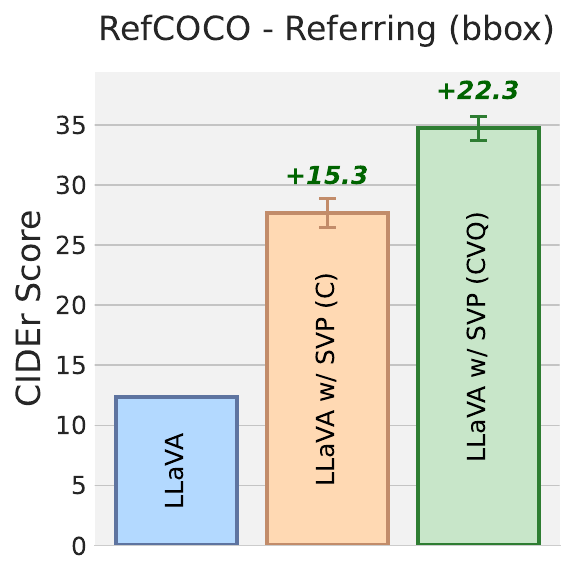}
        \includegraphics[width=.16\textwidth]{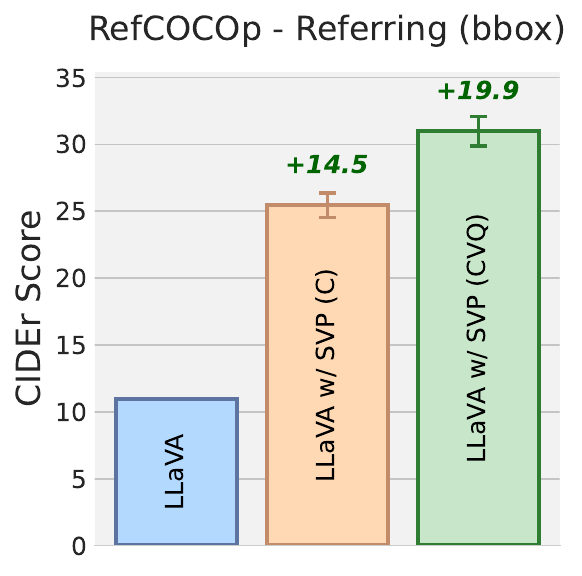}
        \includegraphics[width=.16\textwidth]{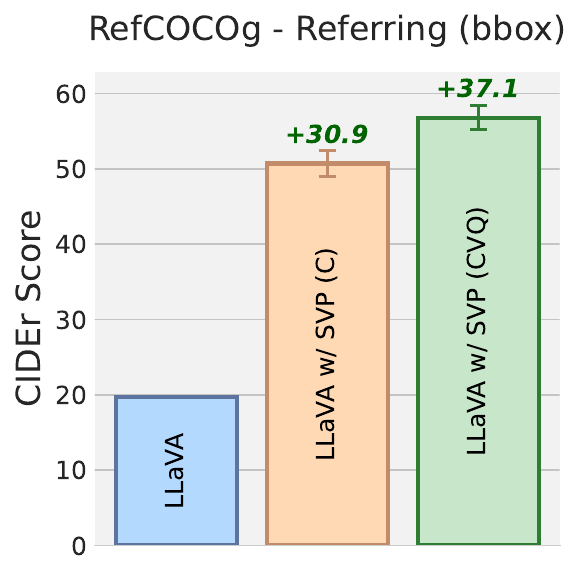}
        \vline
        \includegraphics[width=.16\textwidth]{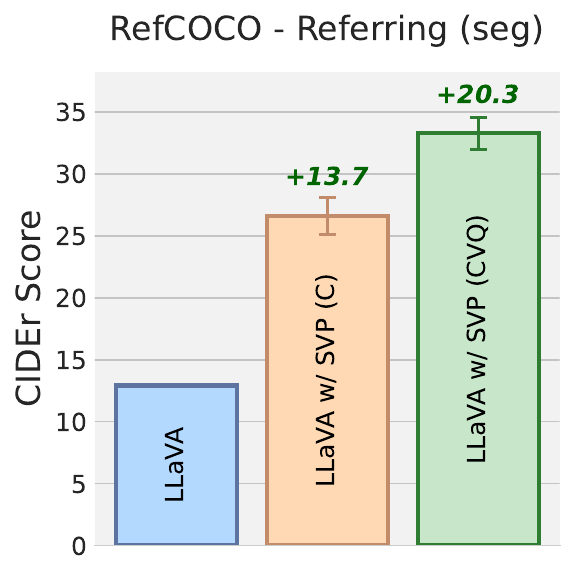}
        \includegraphics[width=.16\textwidth]{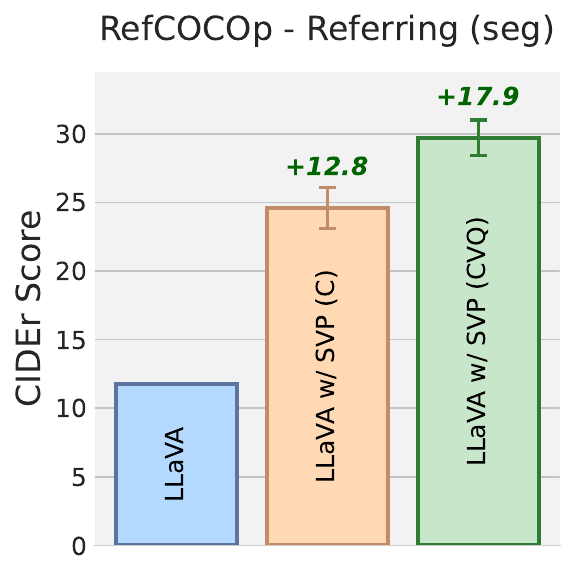}
        \includegraphics[width=.16\textwidth]{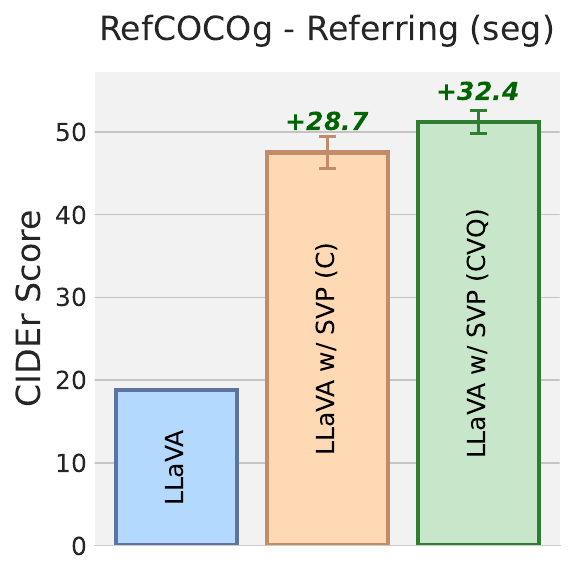}
        \caption{Referring w/ Bounding Box (left) and Segmentation Mask (right).}
        \label{bench:referring}
        \includegraphics[width=.16\textwidth]{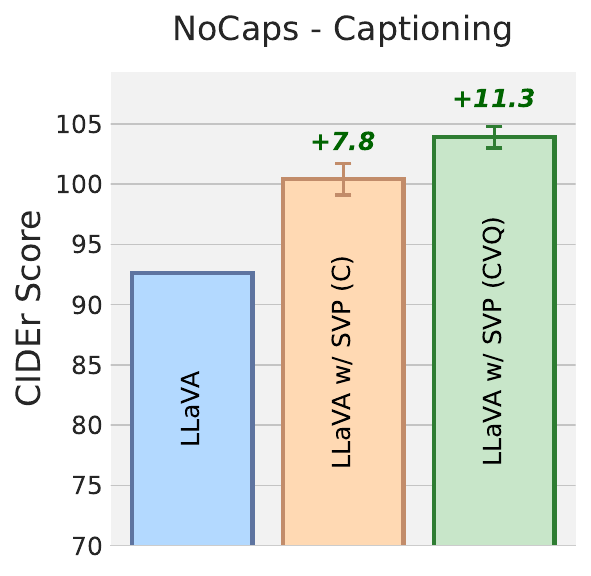}
        \includegraphics[width=.16\textwidth]{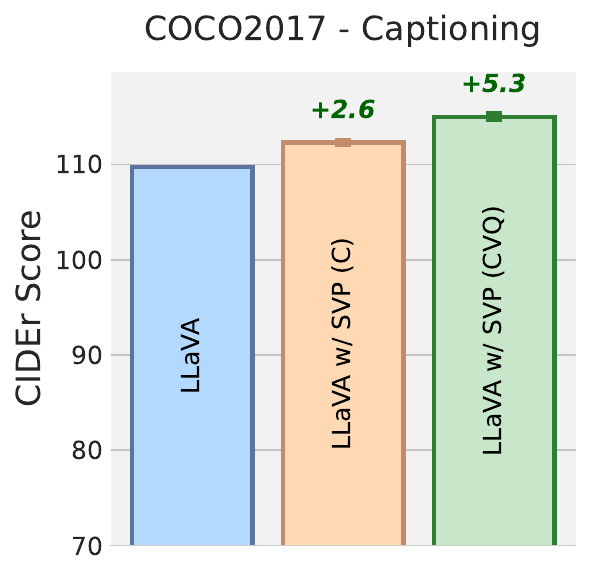}
        \includegraphics[width=.16\textwidth]{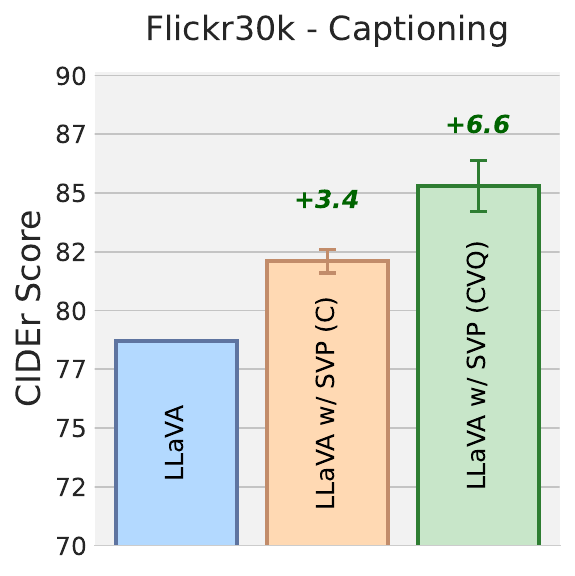}
        \vline
        \includegraphics[width=.16\textwidth]{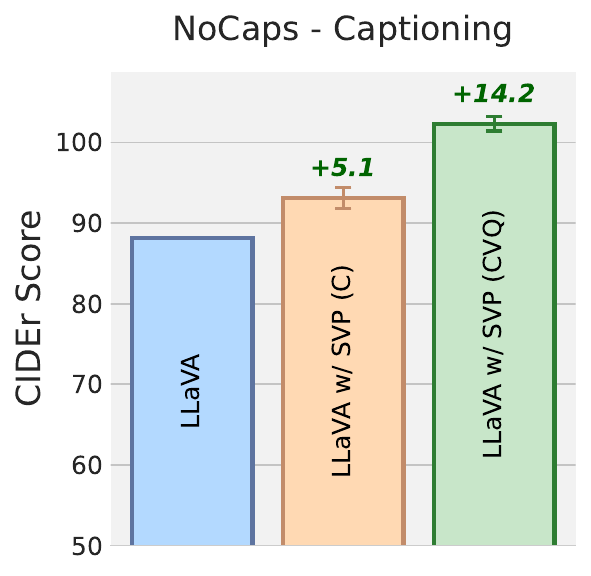}
        \includegraphics[width=.16\textwidth]{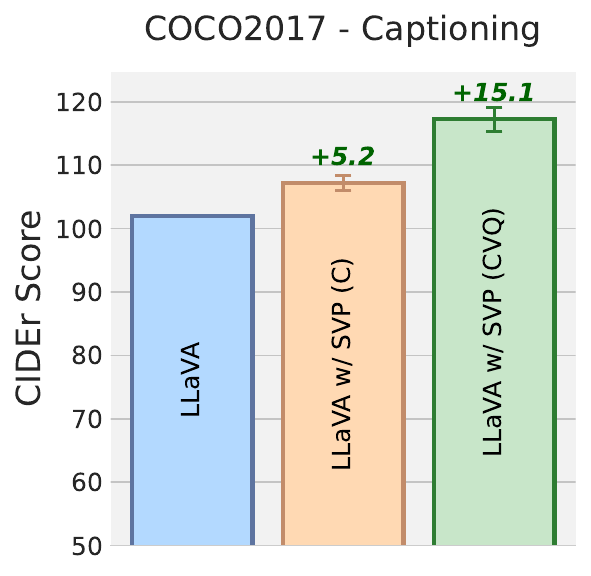}
        \includegraphics[width=.16\textwidth]{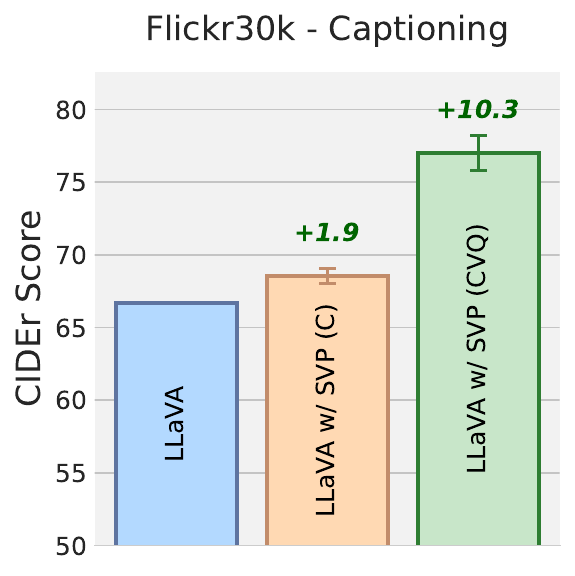}
        \caption{Captioning w/ 7b (left) and 13b (right) models.}
        \label{bench:caption}
    \end{minipage}
    \hfill
    \begin{minipage}[b]{\textwidth}
        \centering
        \includegraphics[width=.36\textwidth]{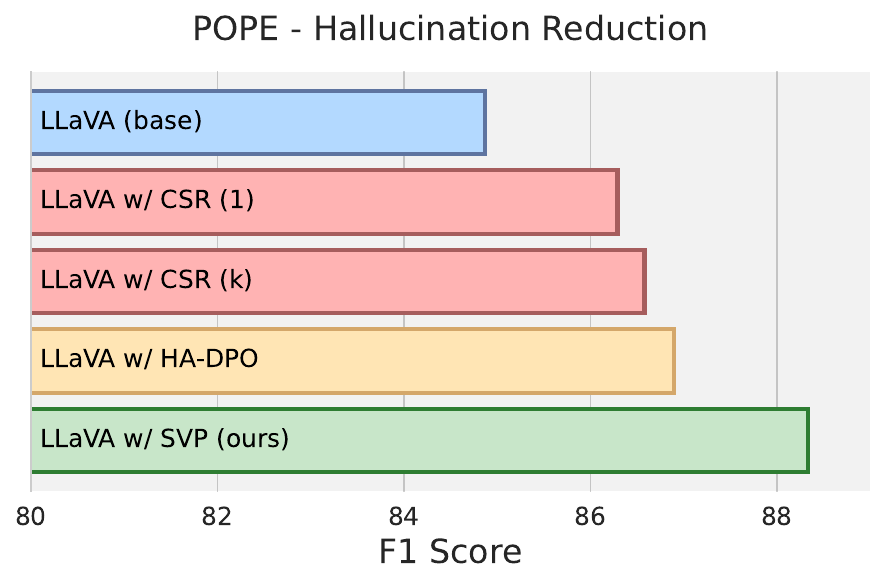}
        \includegraphics[width=.26\textwidth]{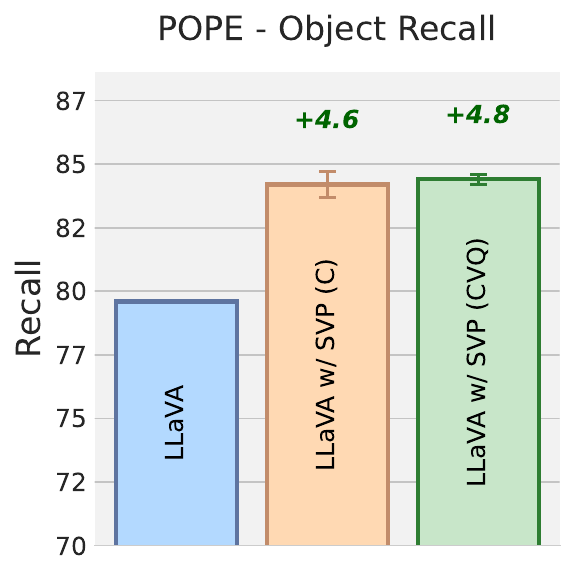}
        \includegraphics[width=.36\textwidth]{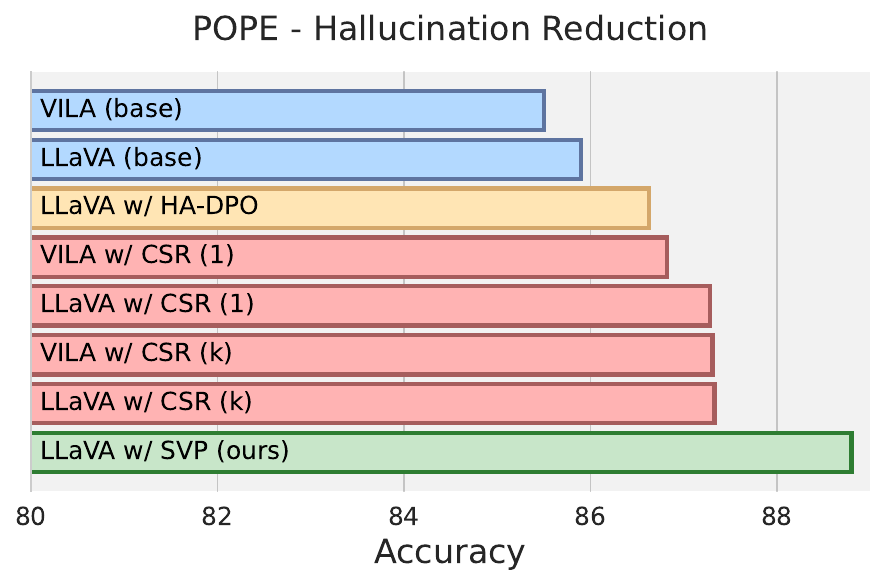}
        \caption{Object Recall and Hallucination Reduction.}
        \label{bench:hallucination}
    \end{minipage}
        \caption{
\textbf{Benchmark Results} comparing base models to our \texttt{SVP}-adapted model on captioning (CIDEr), referring (CIDEr), hallucination control (F1), and object recall (R). Models were adapted using three sets of 1,000 images from the \texttt{COCO2014} training set, with self-captioning and grounding feedback. Higher scores indicate better performance. \texttt{SVP} demonstrates significant improvements in captioning, referring, object recall, and hallucination reduction.}
\label{fig:barplot-results}
\end{figure}

Researchers have explored various approaches to solve the above problem in VLMs (bottom Fig.~\ref{fig:vp}). 
Most of these works focus on  fine-tuning VLMs with supervised (carefully curated) data to improve grounding~\cite{peng2024grounding,beyer2024paligemma, yuan2021florence,you2023ferret,zhang2025llava} and vision-language alignment~\cite{liu2024improved, sun2023aligning}. 
Unfortunately, this data approach tends to be costly and sample-inefficient, requiring large amounts of image-text annotations even for small models to resolve the above stated problem~\cite{yuan2021florence}. 

Preference-based post-training methods~\cite{ouyang2022training,christiano2017deep,rafailov2024direct} as another popular approach align VLM outputs with visual inputs~\cite{zhou2024aligning,sun2023aligning} but require curated preference pairs~\cite{sun2023aligning, favero2024multi}. And, test-time approaches~\cite{wan2024contrastive,leng2024mitigating,favero2024multi,yin2023woodpecker} improve grounding without architectural changes, yet their computational demands and model-specific heuristics limit broad applicability.

To address the significant challenges posed by the extensive data and annotation requirements of modern VLMs, we propose to leverage external feedback to enhance the alignment between visual and linguistic modalities in a task-agnostic manner (right side Fig.~\ref{fig:vp}). %

\emph{Drawing inspiration from human learning, we propose to emulate the way humans efficiently align sensory experiences with language by \emph{grounding} new information in tangible visual examples leveraging feedback}~\cite{hattie2007power, tenenbaum1989meta, tenenbaum2011grow}.
We hypothesize that spatial and positional reasoning is the key for connecting the low-level visual elements and high-level linguistic representations~\cite{peirce2015understanding,oquab2014learning,vallar2007spatial}, and that an external visual grounding model~\cite{liu2023grounding}, agnostic to the VLM's shortcomings, can be used as feedback to extract latent information in the models.

Specifically, in this work, we introduce \texttt{SVP} (\emph{Sampling-based Visual Projection}, Fig.~\ref{fig:main-loop-fig}), an algorithm founded on two core principles: self-improvement and grounding feedback. The self-improvement approach~\cite{zelikman2022star,anthony2017thinking,gulcehre2023reinforced} utilizes the model's own outputs to enhance its performance. And, the grounding feedback provides the VLM with a mechanism to improve its output and select informative samples. Our goal is not to directly build a specialist grounding model, but to \emph{leverage grounding as feedback to elicit latent information in the model}, with the aim of better aligning language and visual representations without the need of costly image-text annotations~\cite{sun2023aligning,peng2024grounding}, preference data~\cite{ouyang2022training,rafailov2024direct}, or multi-step inference workflows~\cite{yin2023woodpecker,wan2024contrastive}.
See Sec~\ref{appx:related_work} for extended related work.

\texttt{SVP} is a three-step process: \emph{(i)} \emph{Inner-Loop Sampling:} A base VLM generates detailed and comprehensive image descriptions. These descriptions are then processed by a pre-trained grounding model~\cite{liu2023grounding}. The resulting spatially enriched grounding output serves as feedback, conditioning the same VLM to generate text tokens that better align with the visual information (Fig.~\ref{fig:main-loop-fig}.$(i)$).
\emph{(ii)} \emph{Scoring:} This step employs a scoring and ranking mechanism to select grounded samples that are more informative and better aligned with the visual input (Fig.~\ref{fig:main-loop-fig}.$(ii)$).
\emph{(iii)} \emph{Outer-Loop Adaptation:} The base VLM undergoes adaptation~\cite{hu2021lora} on the filtered dataset. Importantly, the grounding information is not shown during the fine-tuning process but is utilized during inference (Fig.~\ref{fig:main-loop-fig}.$(iii)$).

\paragraph{Contributions} 
Our key contributions are:
\begin{itemize}%
\item We introduce \emph{Sampling-based Visual Projection} (\texttt{SVP}), a novel framework that enhances vision-language alignment through iterative self-improvement, leveraging self-captioning and visual grounding techniques without requiring expensive image-text annotations or preference data.

\item We develop a principled formulation based on hierarchical sampling, and feedback-driven optimization, where grounding guides the sampling process toward better vision-language alignment. Our design ensures easy applicability across various VLM architectures and scales while providing interpretable vision-language alignment.

\item We demonstrate \texttt{SVP}'s effectiveness through comprehensive experiments across 10 diverse vision-language benchmarks, including captioning, referring expressions, visual question answering, and hallucination control, using only a small set of curated images and a pretrained grounding model.
\end{itemize}

\section{Background}
\begin{figure}[ht]
    \centering
    \begin{minipage}[b]{0.48\textwidth}
        \centering
    \includegraphics[width=\linewidth]{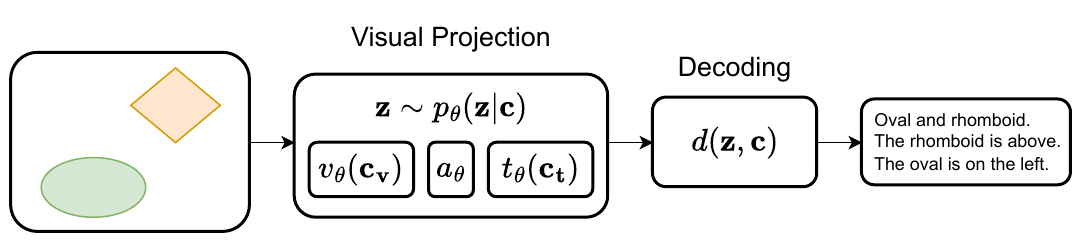}
    \end{minipage}
    \hfill
    \begin{minipage}[b]{0.48\textwidth}
        \centering
        \includegraphics[width=\linewidth]{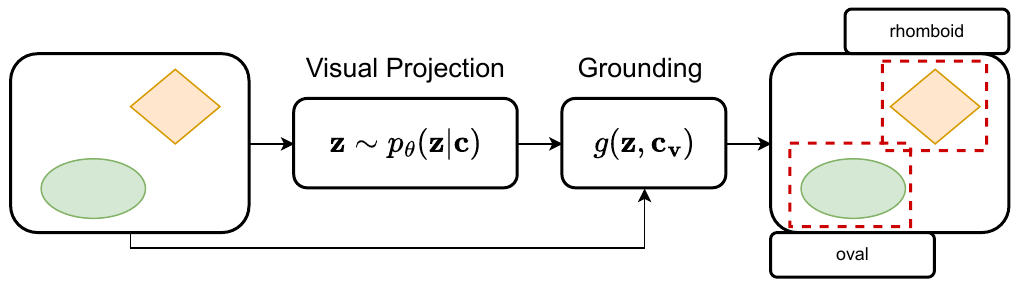}
    \end{minipage}
\caption{Vision-Language Generative Model (left) and Vision-Language Grounding (right)}
\label{fig:vision-language-models}
\end{figure}
\paragraph{Notation}
We use $p(\vx | \vc)$ and $p(\vz | \vc)$ to denote auto-regressive distributions, where $\vc$ is the conditioning information (image and prompt), $\vz$ is a visual projection using grounding feedback, and $\vx$ is the task-specific output. These distributions follow $p(\vx | \vc) = p(\vx_T | \vc) \prod^{T}_{t=1} p(\vx_{t-1} | \vx_{t}, \vc)$, with similar form for $p(\vz | \vc)$. For latent variables, $\vz$ represents trajectories $\vz_{1:T_z}$. We assume a deterministic output distribution $p(\vx | \vz, \vc) = \delta(\vx - d(\vz, \vc))$, as is common in tokenization-based models. Given context $\vc = (\mathbf{c_{v}}, \mathbf{c_{t}})$ with visual input $\mathbf{c_{v}}$ and text prompt $\mathbf{c_{t}}$, we define a Visual Projection as $p(\vz | \vc)$ and its grounded version as $q(\vz | \vc, \vg)$ when conditioning on grounding $\vg$. The conditional entropy is $\bH[\vz | \vc] = -\int_{\vz} p(\vz | \vc) \log p(\vz | \vc)$.

\paragraph{Vision-Language Models}
Generative VLMs are multimodal systems processing both text and images. LLaVA-like architectures (Fig.~\ref{fig:vision-language-models}, left) integrate a visual encoder $v_{\theta}(\mathbf{c_{v}})$, text encoder $t_{\theta}(\mathbf{c_{t}})$, visual-text alignment adapter $a_{\theta}$, and large language model. The model $p_{\theta}$ generates token trajectories $\vz$ from conditioning $\vc$ for various downstream tasks. These systems undergo three training phases: multimodal pre-training, visual-text alignment, and instruction tuning~\cite{zhu2023minigpt,liu2024llava,li2022blip}, enabling broad cross-modal capabilities.

\paragraph{Vision-Language Grounding}
Grounding links language descriptions to spatial regions in images. A grounding model $g(\vz, \mathbf{c_{v}})$ processes visual $\mathbf{c_{v}}$ and textual $\vz$ inputs to produce open-set detection labels and bounding boxes (Fig.~\ref{fig:vision-language-models}, right). While traditional object detection uses fixed-class classification, modern approaches like GLIP and GroundingDINO reframe detection as text-guided grounding. This flexibility enables broader applications in detection and spatial understanding tasks.

\section{Method}
We present Sampling-based Visual Projection (\texttt{SVP}), a general method to sample, score, and adapt a vision-language model (VLM) in the absence of paired image-text data and extrinsic environmental feedback. \texttt{SVP} draws inspiration from self-improving iterative techniques for reasoning in language models~\cite{zelikman2022star,zelikman2024quiet,gulcehre2023reinforced} and sampling in latent variable models~\cite{jordan1999introduction,hoffman2024training}. 
Our approach combines an inner-loop sampling process with an outer-loop adaptation mechanism to improve vision-language alignment.
The core idea of \texttt{SVP} is to generate a task-agnostic language-based representation $\vz$, referred to as Visual Projection (VP), for the visual input $\vc$. 
\begin{figure}
\centering
    \includegraphics[width=\linewidth]{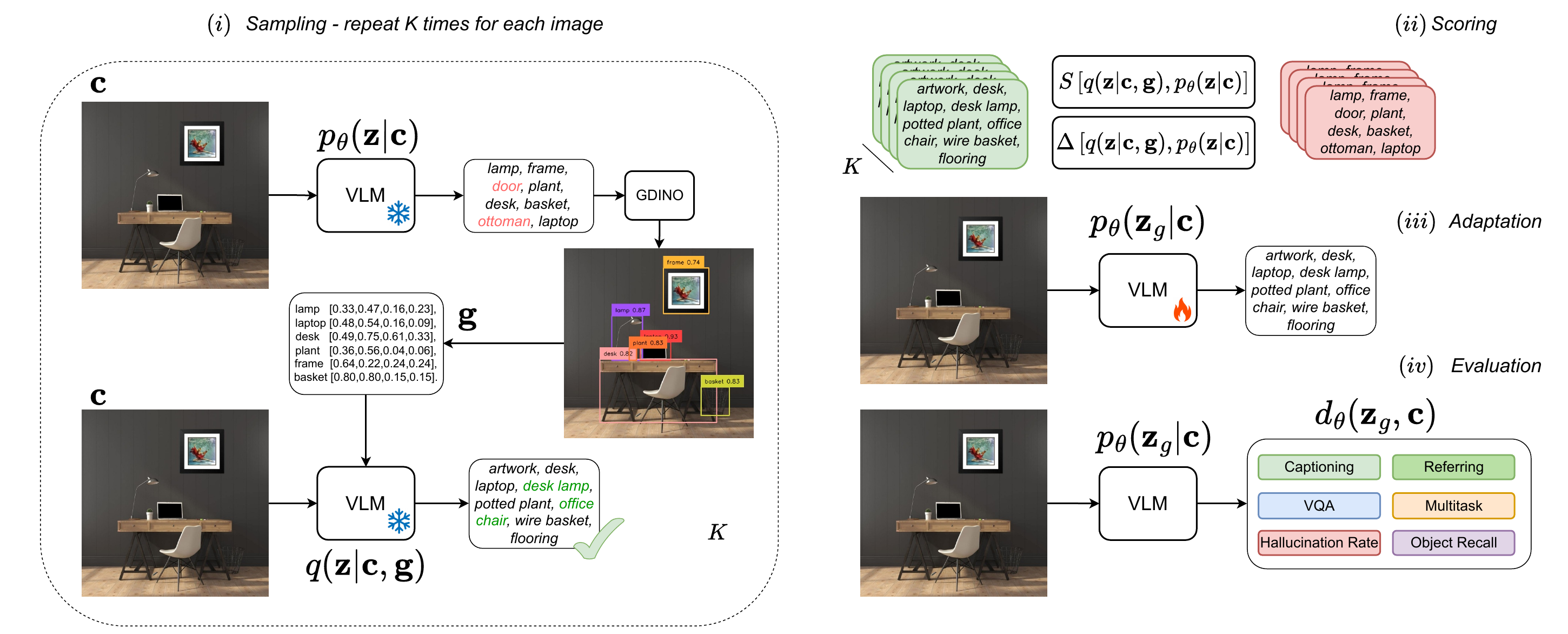}
\caption{
\textbf{\texttt{SVP} Overview.} The inner-loop (left) generates $K$ samples per input $C$ with and without grounding, then scores and ranks them, selecting the top $20\%$ (right side). 
Instead of visually representing the grounding, we transform it into textual form and incorporate it into the prompt as context.
This process includes $(i)$ data generation with grounding feedback and $(ii)$ sample scoring. The outer-loop (right) uses selected samples to $(iii)$ adapt the base model. Post-\texttt{SVP} adaptation, we evaluate on ten benchmarks and six tasks  $(iv)$. Full VLM output in Appx~\ref{fig:intro-figure}. Prompt structure in Appx~\ref{appx:prompting}.
}
\label{fig:main-loop-fig}
\end{figure}
These VPs function as latent variables or generalized captions, and \texttt{SVP} aims to refine them through self-improving iterative methods, strengthening the alignment between vision and language modalities to enhance the base VLM's performance across diverse tasks.
We now present our sampling procedure, scoring mechanisms, and adaptation strategy for improving vision-language alignment in VLMs.

\paragraph{Problem Formulation}
\begin{wrapfigure}{r}{0.4\textwidth}
\vspace{-.5cm}
    \centering
    \includegraphics[width=0.3\linewidth]{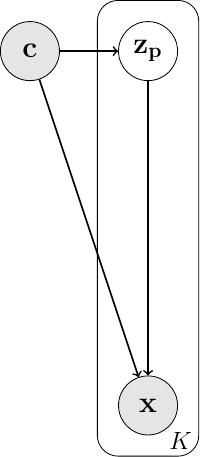}
    \includegraphics[width=0.3\linewidth]{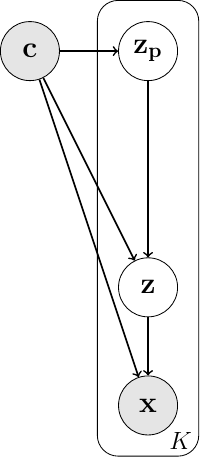}
    \includegraphics[width=0.3\linewidth]{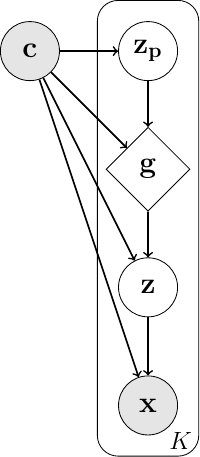}
    \caption{Graphical Models for the sampling processes. Left: standard sampling. Center: hierarchical sampling. Right: hierarchical sampling with internal structure.}
    \label{fig:sampling-process}
\end{wrapfigure}
For a VLM with conditional model $p_{\theta}(\vx | \vc)$, where $\vc = (\vc_v, \vc_t)$ contains visual input and optional text prompt, direct sampling often yields poor alignment between visual and textual modalities. To address this, we introduce a visual projection as a latent variable (Fig~\ref{fig:sampling-process}, left)
\begin{equation}
p_{\theta}(\vx, \vz | \vc) = p(\vx | \vz, \vc) p_{\theta}(\vz | \vc),
\label{eq:x_c}
\end{equation}
where $\vz$ acts as an intermediate visual projection bridging vision and language, similar to chain-of-thought approaches in LLMs. To enhance flexibility and control through ancestral sampling, we extend to a hierarchical structure (Fig~\ref{fig:sampling-process}, center)
\begin{equation}
p_{\theta}(\vx, \vz, \vz_p | \vc) = p(\vx | \vz, \vc) p(\vz | \vz_p, \vc) p_{\theta}(\vz_p | \vc).
\label{eq:x_zp_c}
\end{equation}
While this hierarchical structure offers more flexibility, it provides minimal improvement without proper optimization. Simply iterating through the same visual input and refining projections without feedback can lead to model collapse. To address this limitation, we incorporate a grounding model $\vg = g(\vz_p, \vc)$ into the hierarchical projection (Fig~\ref{fig:sampling-process}, right)
\begin{equation}
p^{g}_{\theta}(\vx, \vz, \vz_p | \vc) = p(\vx | \vz, \vc) q(\vz | g(\vz_p, \vc), \vc) p_{\theta}(\vz_p | \vc).
\label{eq:x_zq_c}
\end{equation}

Here, $q$ is a guided distribution utilizing the grounding model $g$, which provides specialized feedback for vision-language alignment. This feedback mechanism is particularly effective for improving spatial relationships and object attributes, where grounding helps correct the base model's initial predictions. The discrepancy between base model predictions and grounded outputs serves as a valuable signal for enhancing vision-language alignment, especially in cases where grounding information conflicts with initial model predictions.

\paragraph{Sampling}
We implement a guided three-step sampling process to generate improved visual projections: (1) Prior Sampling, where we generate initial projections $\vz_p \sim p_{\theta}(\vz_p | \vc)$ from the base model; (2) Grounding, where we apply the grounding model to obtain feedback $\vg \leftarrow g(\vz_p, \vc)$; and (3) Guided Sampling, where we generate guided visual projections $\vz \sim q(\vz | g(\vz_p, \vc), \vc)$. This process repeats $K$ times for each visual input $\vc$.
For each guided sample, we evaluate the guided distribution $q(\vz | \vc, \vg)$ with grounding feedback $\vg$ and the prior distribution $p_{\theta}(\vz | \vc)$ using the base model. This computation allows us to quantify grounding effects by comparing guided and prior distributions token-wise over the vocabulary, revealing how visual context influences model predictions. For practical implementation, we convert visual grounding to textual form and include it in the prompt as context, rather than using direct visual representation. 
The complete prompt structure and examples are detailed in Appx~\ref{appx:prompting}.

\begin{figure}[ht!]
    \centering
    \includegraphics[width=\linewidth]{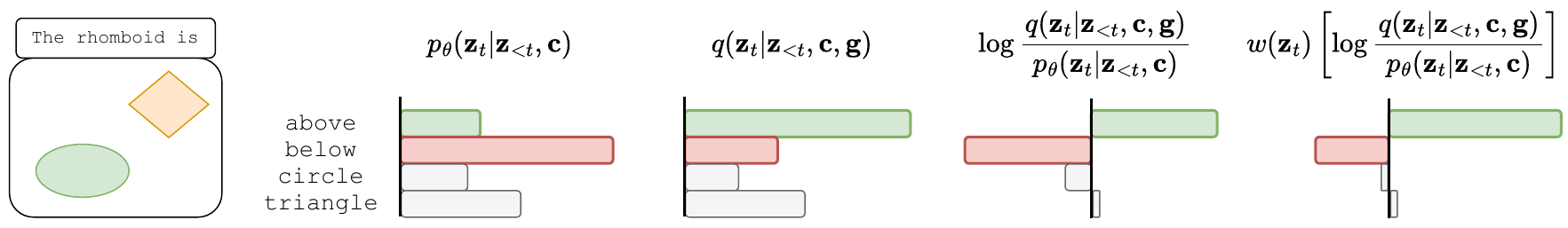}
    \caption{\textbf{Visualization of prior and guided distribution} for token $t$ over vocabulary $V = \{\texttt{above}, \texttt{below}, \texttt{circle}, \texttt{rhomboid}\}$. The base model $p_{\theta}$ incorrectly predicts ``\texttt{below}'' for the circle-rhomboid spatial relationship. With grounding feedback, $q$ correctly assigns higher likelihood to ``\texttt{above}''. Using log-ratio and re-weighting with $w(\vz_t) \propto q(\vz_t | \vz_{< t}, \vc, \vg)$ emphasizes grounding-relevant tokens while down-weighting tokens with similar likelihoods in both distributions.}
    \label{fig:distro-scoring}
\end{figure}

\paragraph{Scoring}
We evaluate sample quality by viewing alignment as a feedback-driven process inspired by policy optimization~\cite{rafailov2024direct,peters2007reinforcement,go2023aligning}. We define a scoring function\footnote{if we assume that $q$ is the optimal alignment policy, we can write $q(\vz | \vc, \vg) \propto p_{\theta}(\vz | \vc) \exp(S(\vz)/ w)$} that measures the \emph{alignment gap} between the guided and prior distributions:
\begin{equation} 
    S(\vz) \propto \log q(\vz | \vc, \vg) - \log p_{\theta} (\vz| \vc), \quad \vz \sim q(\vz | \vc, \vg).
\label{eq:kl-sample}
\end{equation}
This score quantifies the effect of grounded visual projection on the model. When grounding provides no additional information, $q(\vz | \vg, \vc) \approx p(\vz | \vz_p, \vc)$, and Eq.~\ref{eq:x_zq_c} reduces to~\ref{eq:x_c}.
The score approximates the one sample KL divergence between $q$ and $p_{\theta}$. Low values indicate token trajectories well-known to the base model, while high values reveal surprising trajectories that offer learning opportunities. As shown in Fig.~\ref{fig:distro-scoring}, the guided distribution $q$ helps correct misaligned predictions of the base model. We implement two scoring approaches. First, a log-ratio scoring:
\begin{align}
\begin{split}
S(q, p)_{\vz} = \sum\nolimits^{T}_{t=1}  \sum\nolimits^{V}_{v=1}w_{v,t} \left[ \log q_{v,t} - \log {p_{\theta}}_{v,t}\right]
\end{split}
\label{eq:score-kl}
\end{align}
where $w_{v,t} \propto q(\vz_t | \vz_{< t}, \vc, \vg)$ over-emphasizes grounding-relevant tokens. Second, a weighted-difference scoring:
\begin{equation}
    \Delta(q,p)_{\vz} = 
    \sum\nolimits^{T}_{t=1}  \sum\nolimits^{V}_{v=1} w^{q}_{v,t} \log q_{v,t} 
    - \sum\nolimits^{T}_{t=1}  \sum\nolimits^{V}_{v=1} w^{p}_{v,t} \log {p_{\theta}}_{v,t}
\label{eq:score-entropy}
\end{equation}

The weighted-difference score~\cite{settles2009active} is inspired by the fact that grounding should reduces prediction uncertainty: $\bH[\vz | \vc, \vg] < \bH[\vz | \vc]$. Both scoring methods provide similar signals for grounding and diversity (correlation analysis in Appx~\ref{fig:score-corr-appx}). Importantly, generic surprise alone (pure exploration) does not enhance vision-language alignment. Our hypothesis is that informative grounding conditioning makes surprising instances statistically valuable for learning and alignment.

\paragraph{Adaptation}
Inspired by re-weighted regression~\cite{peng2019advantage} and off-policy policy optimization~\cite{roux2025tapered,ahmadian2024back,gulcehre2023reinforced}, we propose an iterative optimization where $q(\vz | \vg, \vc)$ serves as a behavioral policy providing high-quality demonstrations, while $p_{\theta}(\vz | \vc)$ is our target model. We maximize:
\begin{equation}
\mathcal{\tilde F}(\vc; \theta) =
\dfrac{1}{|k(\vc)|} \sum\nolimits^{K}_{i=1} 
    \left[ \mathbbm{1}\{\vz^i : S(q(\vz^i | \vc, \vg), p_{\theta}(\vz^i | \vc)) \geq S_{k(\vc)}\}\right]
\log p_{\theta}(\vz^i | \vc)
\label{eq:loss}
\end{equation}
where $S_{k(\vc)}$ is the $k$-th highest score among $K$ samples generated for image $\vc$ from the guided distribution, $\{\vz^i\}^{K}_{i=1} \sim q(\vz | \vc, \vg)$. This objective can be interpreted as both re-weighted maximum likelihood and greedy off-policy optimization (detailed in Appx~\ref{appx:objective}). While not necessarily optimal for likelihood or policy metrics, this approach prioritizes vision-language alignment by selectively optimizing better-aligned samples. The final training loss averages this objective over a batch of visual inputs $\vc$.

\paragraph{Inner/Outer-loop Interpretation}
Our approach follows a meta-learning framework~\cite{li2017meta} with nested optimization loops. The inner loop learns task-specific policies through guided sampling and scoring, while the outer loop adapts model parameters using high-quality samples via re-weighted loss. This structure mirrors meta-learning strategies that leverage learned behaviors to enhance overall performance, naturally balancing exploration through guided sampling with exploitation via model adaptation.
Though \texttt{SVP} supports iterative refinement (Fig.~\ref{fig:iterations-recall}), significant improvements emerge after just one iteration, highlighting the effectiveness of our scoring and selection mechanisms.

\section{Experiments}

\paragraph{Base Model Selection}
Our study centers on the LLaVA family~\cite{liu2024visual} instead of larger state-of-the-art alternatives for three main reasons:
\emph{(i) Capability Gap}. LLaVA’s straightforward supervised fine-tuning approach reveals clear performance limitations (Table~\ref{tab:main-results}), providing an ideal benchmark for validating SVP’s ability to bootstrap fundamental visual-language skills from scratch.
\emph{(ii) Transparent Dataset}. LLaVA’s open-source, compact training datasets allow for precise evaluation and ensure there is no overlap with benchmark evaluation sets.
\emph{(iii) Controlled Progress}. The incremental dataset expansions within the LLaVA family facilitate unambiguous assessment of performance improvements, free from confounding factors such as proprietary data or complex post-training interventions.

This approach provides a clearer validation than improving already capable models with inherent grounding mechanisms or extensive reinforcement fine-tuning. We further strengthen our analysis through comprehensive comparisons with existing models, with a particular focus on hallucination reduction (Table~\ref{tab:hallucination-f1-results} and~\ref{tab:hallucination-acc-results}).

\paragraph{Seed Images and Models}

\begin{table}[htbp]
\begin{minipage}{0.49\textwidth}
    \centering
    \caption{\textbf{Hallucination Mitigation - F1} scores on POPE benchmark comparing LLaVA variants across adversarial, popular, random, and overall splits. Results show how hallucination avoidance is influenced by model size, fine-tuning approach, encoder selection, and SVP adaptation. See~\ref{appx:model_size} for analysis of model scaling effects.
    }
    \resizebox{\linewidth}{!}{%
    \setlength{\tabcolsep}{4pt}
    \begin{tabular}{lccccccc}
        \toprule
        &&&&  \multicolumn{4}{c}{POPE ($F1$ score $\uparrow$)} \\
        \multirow{1}{*}{Model} & Size &  $v_{\theta}$  & $t_{\theta}$                       & adv & pop & random & all \\ 
        \midrule
        \multirow{1}{*}{LLaVA}~\cite{liu2024visual} & 7b  & CLIP  & Vicuna & 72.0 & 75.3 & 80.7  & 76.0 \\
        \multirow{1}{*}{LLaVA-SFT$^{+}$}~\cite{sun2023aligning} & 7b & CLIP  & Vicuna & 80.1   & 82.4 & 85.5 & 82.7 \\
        \multirow{1}{*}{LLaVA-RLHF}~\cite{sun2023aligning} & 7b & CLIP  & Vicuna & 79.5   & 81.8 & 83.3 & 81.5 \\
        \multirow{1}{*}{LLaVA}~\cite{liu2024visual} & 13b & CLIP  & Vicuna & 74.4   & 78.2 & 78.8 & 77.1 \\ 
        \multirow{1}{*}{LLaVA-SFT$^{+}$}~\cite{sun2023aligning} & 13b & CLIP  & Vicuna & 81.1   & 82.6 & 84.8 & 82.8 \\
        \multirow{1}{*}{LLaVA-RLHF}~\cite{sun2023aligning} & 13b & CLIP  & Vicuna & 80.5   & 81.8 & 83.5 & 81.9 \\
        \midrule
        LLaVA-NeXT-DPO~\cite{liu2024llava} & 7b & CLIP   & Qwen2  & 83.43 & 83.78   & 84.73 & 83.98  \\
        LLaVA-OV-DPO~\cite{li2024llava}   & 7b & SigLIP & Qwen2  & 85.12 & 86.24   & 87.37 & 86.24  \\
        LLaVA-HA-DPO~\cite{zhao2023beyond}   & 7b & CLIP   & Vicuna & 82.54 &   \underline{87.89} & \textbf{90.25} &  86.90 \\
        \midrule
        \multirow{1}{*}{LLaVA-1.5}~\cite{liu2024improved}          & 13b & CLIP  & Vicuna &  84.53 & 86.31 & 87.17 & 86.00 \\ 
        \multirow{1}{*}{LLaVA-1.5 w/ \texttt{SVP}} & 13b & CLIP  & Vicuna & 84.66  & 86.84 & 87.44 & 86.31 \\ 
        \midrule
        \multirow{1}{*}{LLaVA-1.6}~\cite{liu2024llava}       & 7b  & CLIP  & Mistral &  \underline{85.43} & 86.87 & 88.05 & 86.73 \\ 
        \multirow{1}{*}{LLaVA-1.6 w/ \texttt{SVP}} & 7b & CLIP & Mistral & \textbf{85.93} & \textbf{89.04} & \underline{90.02} & \textbf{88.33}\\
        \midrule
        \multirow{1}{*}{LLaVA-1.6}~\cite{liu2024llava}          & 13b  & CLIP &Vicuna & 85.17  & 86.36 & 87.20 & 86.24 \\
        \multirow{1}{*}{LLaVA-1.6 w/ \texttt{SVP}} & 13b & CLIP & Vicuna & 85.15 & 87.50 & 89.23 & \underline{87.30} \\
        \midrule
        LLaVA-OV~\cite{li2024llava} & 0.5b & SigLIP & Qwen2 & 82.28 & 83.19 & 83.89 & 83.12 \\
        LLaVA-OV w/ \texttt{SVP} & 0.5b & SigLIP & Qwen2 & 83.45 & 84.70 & 85.46 & 84.53\\
        \midrule
        \emph{\color{Gray} Bigger VLMs} \\
        \color{Gray}{LLaVA-1.6}~\cite{liu2024llava} & \color{Gray}34b  & \color{Gray}CLIP & \color{Gray}Yi-2 &  \color{Gray}- & \color{Gray}- & \color{Gray}- & \color{Gray}87.7   \\
        \midrule
        \color{Gray}{InternVL}~\cite{chen2024internvl} & \color{Gray}19b & \color{Gray}IViT & \color{Gray}Vicuna & \color{Gray}- & \color{Gray}- & \color{Gray}- & \color{Gray}87.6 \\
        \color{Gray}InternVL-1.2~\cite{chen2024internvl} & \color{Gray}40b & \color{Gray}IViT & \color{Gray}Yi-2 & \color{Gray}- & \color{Gray}- & \color{Gray}- & \color{Gray}88.0 \\
        \color{Gray}InternVL-1.2$^{+}$~\cite{chen2024internvl} & \color{Gray}40b & \color{Gray}IViT & \color{Gray}Yi-2 & \color{Gray}- & \color{Gray}- & \color{Gray}- & \color{Gray}88.7 \\
        \midrule
        \color{Gray}VILA-1.5~\cite{lin2024vila} & \color{Gray}8b & \color{Gray}SigLIP & \color{Gray}LLaMA3 & \color{Gray}- & \color{Gray}- & \color{Gray}- & \color{Gray} 85.6 \\
        \color{Gray}VILA-1.5~\cite{lin2024vila} & \color{Gray}8b & \color{Gray}SigLIP & \color{Gray} Vicuna  & \color{Gray}- & \color{Gray}- & \color{Gray}- & \color{Gray} 86.3 \\
        \color{Gray}VILA-1.5~\cite{lin2024vila} & \color{Gray}40b & \color{Gray}IViT & \color{Gray}Yi2 & \color{Gray}- & \color{Gray}- & \color{Gray}- & \color{Gray} 87.3 \\
        \color{Gray}VILA-1.5-AWQ~\cite{lin2024vila} & \color{Gray}40b & \color{Gray}IViT & \color{Gray}Yi2 & \color{Gray}- & \color{Gray}- & \color{Gray}- & \color{Gray}88.2 \\
        \bottomrule
    \end{tabular}
    }
\label{tab:hallucination-f1-results}
\end{minipage}
\hfill
\begin{minipage}{0.49\textwidth}
    \centering
    \caption{\textbf{Hallucination Mitigation - Accuracy} across VLMs using fine-tuning, train-time, and test-time adaptation approaches. Higher scores indicate better performance. Size (Eff) indicates total parameters for multi-phase inference, e.g., Woodpecker (Wp)~\cite{yin2023woodpecker} requires multiple models for response processing.}
    \resizebox{\linewidth}{!}{%
    \setlength{\tabcolsep}{4pt}
    \begin{tabular}{lcccccc}
        \toprule
        &&&&  \multicolumn{3}{c}{POPE ($Acc$ score $\uparrow$)} \\
        \multirow{1}{*}{Model} & Size (Eff) &  $v_{\theta}$  & $t_{\theta}$                       & adv & pop & random \\ 
        \midrule
        \emph{Fine-tuning} \\
        \multirow{1}{*}{InstructBLIP}~\cite{dai2023instructblipgeneralpurposevisionlanguagemodels} & 7b & ViT & FlanT5 & 72.1 & 82.7 & 88.6 \\
    \multirow{1}{*}{LLaVA-SFT$^{+}$}~\cite{sun2023aligning} & 7b & CLIP  & Vicuna & 80.2 & 82.9 & 86.1   \\
    \multirow{1}{*}{mPLUG-Owl2}~\cite{ye2024mplug} & 8b & ViT & LLaMA2 & 84.1 & 86.2 & 88.3 \\
    \multirow{1}{*}{InstructBLIP}~\cite{dai2023instructblipgeneralpurposevisionlanguagemodels} & 13b & ViT  & Vicuna & 74.5 & 81.4 & 88.7\\
    \multirow{1}{*}{LLaVA-SFT$^{+}$}~\cite{sun2023aligning} & 13b & CLIP  & Vicuna & 82.3 & 83.9 & 85.2 \\
        \midrule
        \emph{Test-time adaptation} \\
        \multirow{1}{*}{QwenVL w/ VCD}~\cite{leng2024mitigating} & 7b (14b) & CLIP & Vicuna & 84.3 & 87.1 & 88.6 \\
        \multirow{1}{*}{LLaVA w/ M3ID}~\cite{favero2024multi} & 7b (14b) & CLIP & Vicuna & 65.8 & 69.3 & 76.0  \\
        \multirow{1}{*}{Otter w/ Wp}~\cite{yin2023woodpecker} & 7b (14b+) & CLIP & LLaMA & 83.0 & 84.3 & 86.7\\
        \multirow{1}{*}{mPLUG-Owl w/ Wp}~\cite{yin2023woodpecker} & 7b (14b+) & ViT & LLaMA & 81.0 & 84.1 & 
        86.3 \\
        \multirow{1}{*}{LLaVA w/ M3ID}~\cite{favero2024multi} & 13b (26b) & CLIP & Vicuna & 71.3 & 77.0 & 84.3 \\
        \midrule
        \emph{Train-time adaptation} \\
        \multirow{1}{*}{LLaVA-M3ID-DPO}~\cite{favero2024multi}   & 7b & CLIP  & Vicuna & 68.2 & 73.9 & 81.2  \\
        \multirow{1}{*}{LLaVA-RLHF}~\cite{sun2023aligning} & 7b & CLIP  & Vicuna & 80.7 & 83.3 & 84.8 \\
        \multirow{1}{*}{LLaVA-NeXT-DPO}~\cite{rafailov2024direct} & 7b  & CLIP & Qwen2 & 85.2 & 85.6 & 86.6 \\
        \multirow{1}{*}{LLaVA-OV-DPO}~\cite{rafailov2024direct} & 7b  & SigLIP & Qwen2 & 86.3 & 87.5 & 88.7 \\
        \multirow{1}{*}{LLaVA-HA-DPO}~\cite{zhao2023beyond} & 7b & CLIP & Vicuna & 81.5 & 87.9 & \textbf{90.5} \\
        SeVa~\cite{zhu2024self} & 7b & CLIP & Vicuna & 83.6 & 87.4 & 89.4 \\
        \multirow{1}{*}{LLaVA-M3ID-DPO}~\cite{favero2024multi} & 13b & CLIP  & Vicuna & 73.2 & 79.1 & 85.2 \\
        \multirow{1}{*}{LLaVA-RLHF}~\cite{sun2023aligning} & 13b & CLIP  & Vicuna & 82.3 & 83.9 & 85.2 \\
        \multirow{1}{*}{InstructBLIP-HA-DPO}~\cite{zhao2023beyond} & 13b & ViT  & Vicuna & 80.7 & 85.8 & 89.8 \\
        \midrule
        \multirow{1}{*}{LLaVA-1.6}~\cite{liu2024llava} & 7b  & CLIP  & Mistral & 86.4 & 87.9 & 89.2  \\
        \multirow{1}{*}{LLaVA-1.6 w/ \texttt{SVP}} & 7b & CLIP & Mistral & 86.2 & \textbf{89.6} & \textbf{90.6} \\ 
        \midrule
        \multirow{1}{*}{LLaVA-1.6}~\cite{liu2024llava} & 13b  & CLIP &Vicuna &  86.4 & 87.7  & 88.5 \\
        \multirow{1}{*}{LLaVA-1.6 w/ \texttt{SVP}} & 13b & CLIP & Vicuna & \textbf{86.7} & 88.4 & 89.2 \\ 
        \midrule
        LLaVA-OV & 0.5b & SigLIP & Qwen2 & 84.3 & 85.2 & 86.0 \\
        LLaVA-OV w/ \texttt{SVP} & 0.5b & SigLIP & Qwen2 & 85.0 & 86.3 & 87.2 \\
        \bottomrule
    \end{tabular}
    }
    \label{tab:hallucination-acc-results}
\end{minipage}
\end{table}

We utilize a pre-trained grounding model~\cite{liu2023grounding} to provide the external feedback signals. 
For our core experiments, we randomly sampled a subset of $C=1000$ natural images from the \texttt{COCO2014} training set~\cite{lin2014microsoft}. 
We conduct a comprehensive comparison against various baselines, including models fine-tuned with self-captioning without grounding and preference-based adaptation methods. Our evaluation encompasses a wide range of model scales (.5, 7, 8, 13, 19, 40 billion parameters), architectures (LLaVA-1.5~\cite{liu2024improved}, LLaVA-1.6~\cite{liu2024llava}, LLaVA-OV~\cite{li2024llava}, VILA~\cite{lin2024vila}, InternVL~\cite{chen2024internvl}), visual encoders (CLIP~\cite{radford2021learning}, SigLIP~\cite{zhai2023sigmoid}, ViT~\cite{dosovitskiy2020image}), language encoders (Vicuna~\cite{chiang2023vicuna}, Mistral~\cite{jiang2023mistral}, Qwen2~\cite{yang2024qwen2}, Yi-2~\cite{young2024yi}), and scoring mechanisms $S(q,p)$ and $\Delta(q,p)$.

\paragraph{Implementation Details}
We implement two SVP variants: SVP (C) using only grounded self-generated captions, and SVP (CVQ) which additionally incorporates visual queries from the model's training history to prevent over-specialization on descriptive tasks.
For the inner-loop sampling, we generate K=20 samples per image from both base and grounded VLMs, selecting the top 10\% using our scoring mechanisms (Eq.~\ref{eq:score-kl}, ~\ref{eq:score-entropy}). With $C=1000$ images, we collect ~4000 samples for SVP (C) and double this for SVP (CVQ) by including visual queries, yielding ~8000 total training pairs. While smaller than typical supervised datasets, this proves sufficient for effective model adaptation~\cite{sun2023aligning,zhu2024self}. We use normalized xyxy bounding boxes and filter out degenerate samples ($<0.5$\% for LLaVA-1.5/1.6, ~5\% for LLaVA-OV), with $w_{v,t} = q_{v,t}$.
For outer-loop adaptation, we fine-tune using LoRA~\cite{hu2021lora} ($\alpha=16$, $r=64$ for $\leq$ 7b models; $\alpha=256$, $r=128$ for 13b models) for one epoch on 8-A100 GPUs with batch size $B=20$. Following~\cite{li2024llava,liu2024llava}, we run up to 3 iterations of SVP. Our evaluation uses sample-wise, zero-shot testing without prompt engineering or batching to ensure fair comparison across model variants.

\paragraph{Metrics}
We use the CIDEr score~\cite{vedantam2015cider} for captioning and referring tasks; accuracy for VQA and multitasking. F1, Accuracy and Recall for hallucination and object recall. We also consider standard metrics for language translation like BLEU~\cite{papineni2002bleu}, METEOR~\cite{banerjee2005meteor}, and ROUGE~\cite{lin2004rouge} scores. We re-compute metrics for LLaVA baselines and variants (1.5, 1.6, OV) up to 13b parameters.

\subsection{Vision-Language Benchmarks}
\paragraph{Datasets}
We evaluate SVP across six tasks using ten standard VLM benchmarks: COCO2017~\cite{lin2014microsoft}, NoCaps~\cite{agrawal2019nocaps}, and Flickr30k~\cite{plummer2015flickr30k} for captioning; RefCOCO variants~\cite{kazemzadeh2014referitgame} for referring expression generation; ScienceQA~\cite{saikh2022scienceqa} and GQA~\cite{hudson2019gqa} for VQA; MMBench~\cite{liu2025mmbench} and MMMU~\cite{yue2024mmmu} for multitasking; and POPE~\cite{li2023evaluating} for hallucination assessment.
Following lmms-eval~\cite{zhang2024lmms}, we use both full and lite evaluation sets for captioning and VQA tasks to demonstrate result stability across sample sizes. For MMMU, POPE, and all RefCOCO variants, we use the complete evaluation sets.

\paragraph{General Results}
Across the 10 datasets and 6 tasks evaluated (Fig.~\ref{fig:barplot-results} and Table~\ref{tab:main-results}), our method demonstrates significant improvements in captioning, referring expression generation, hallucination control, and object recall. We maintain comparable or improved performance on multitasking benchmarks and VQA tasks. The most substantial gains appear in captioning, with nearly 20\% improvement, while performance remains stable even in challenging tasks like visual question answering. 
The impact of \texttt{SVP} is especially dramatic for models with initial weaknesses in specific tasks. For instance, when applied to LLaVA with Mistral, which originally shows poor referring capabilities, SVP improves referring expression generation performance by a factor of three (Fig.~\ref{bench:referring}).

The preservation of VQA performance is particularly significant, as it indicates that our method \emph{enhances vision-language alignment without compromising existing capabilities} or requiring task-specific knowledge injection. This balanced improvement highlights SVP's ability to strengthen fundamental cross-modal understanding while maintaining the model's broader base capabilities.

\begin{table}[ht!]
    \centering
    \caption{\textbf{Benchmark Performance} across LLaVA variants (7B/13B) with same visual encoder (CLIP) and varying the text encoders (Mistral and Vicuna) evaluated using \texttt{lmms-eval} (\texttt{lite} split, full MMMU, POPE, and ScienceQA). Results show \texttt{SVP} and \texttt{SVP}+VQ improve captioning, referring tasks, and object recall while reducing hallucinations, maintaining strong performance on multitask benchmarks. Higher scores are better.} 
    \resizebox{\textwidth}{!}{%
    \setlength{\tabcolsep}{4pt}
    \begin{tabular}{l c c | c c c c c c c cc c c c}
    \toprule
    & & & \multicolumn{2}{c}{VQA} & \multicolumn{3}{c}{Captioning} & \multicolumn{1}{c}{Referring} & \multicolumn{2}{c}{Multitasking} & \multicolumn{2}{c}{Hallucinations}  \\
    Model & $v_{\theta}$ & $t_{\theta}$ & ScienceQA & \multicolumn{1}{c|}{GQA}  & NoCaps & COCO2017 & \multicolumn{1}{c|}{Flickr30k} & \multicolumn{1}{c|}{RefCOCO} & MMBench & \multicolumn{1}{c|}{MMMU} & POPE (F1) & POPE (R) \\
    & & & \texttt{test} & \texttt{test} & \texttt{val} & \texttt{val} & \texttt{test} & \texttt{val} & \texttt{en\_dev} & \texttt{val} & \texttt{all} & \texttt{all} \\
    \midrule
     LLaVA-1.6-7b  & CLIP   & Mistral   & \textbf{78.54} & \textbf{75.80} & 92.60 & 109.68 & 78.74 & 6.70 & \textbf{80.30} & 34.11 & 86.73 & 79.60\\
     w/ \texttt{SVP} (C) &CLIP & Mistral & 77.24 & 73.80 & 100.93 & 112.95 & 83.49 & 18.15 & 77.27 & 36.44 & \textbf{88.33} & \textbf{84.20} \\
     w/ \texttt{SVP} (CVQ) &CLIP & Mistral & \textbf{78.40} & 75.10 & \textbf{103.95} & \textbf{115.02} & \textbf{85.31} & \textbf{24.74} & 78.03 & \textbf{37.44} & \textbf{88.25} & \textbf{84.41} \\
     \midrule
     & & & \multicolumn{2}{c|}{\color{Gray}$\downarrow$  0.54 \%} &  \multicolumn{3}{c|}{\color{dartmouthgreen}$\uparrow$ 8.48 \%}  & \multicolumn{1}{c|}{\color{dartmouthgreen}$\uparrow$ 18.04} & \multicolumn{2}{c|}{\color{dartmouthgreen}$\uparrow$ 3.43 \%} & \multicolumn{2}{c}{\color{dartmouthgreen}$\uparrow$ 3.94 \%} \\
     \midrule
     LLaVA-1.6-13b & CLIP   & Vicuna   & 70.30 & \textbf{74.60} & 83.89 &  104.21 & 69.86 & \textbf{29.71} & \textbf{83.33} & 35.22 & 86.24 & 78.13\\
     w/ \texttt{SVP} (C) & CLIP & Vicuna   & \textbf{74.34} & \textbf{74.40} & 87.09 & 111.09 & 71.43 & 28.93 & 81.06 & \textbf{36.33} & \textbf{87.44} & 81.20\\
     w/ \texttt{SVP} (CVQ) & CLIP & Vicuna &  68.49 & 73.20 & \textbf{100.26} & \textbf{122.03} & \textbf{85.32} & 27.20 & 78.03 & 35.66 & \textbf{87.68} & \textbf{82.53}\\
     \midrule
     & & & \multicolumn{2}{c|}{\color{dartmouthgreen}$\uparrow$  2.65 \%} &  \multicolumn{3}{c|}{\color{dartmouthgreen}$\uparrow$ 19.58 \%}  & \multicolumn{1}{c|}{\color{Gray}$\downarrow$ 0.78} & \multicolumn{2}{c|}{\color{dartmouthgreen}$\uparrow$ 0.12 \%} & \multicolumn{2}{c}{\color{dartmouthgreen}$\uparrow$ 3.65 \%} \\
     \bottomrule
    \end{tabular}
    }
    \label{tab:main-results}
\end{table}

\begin{table}[ht!]
    \centering
    \caption{\tb{Captioning Performance} on COCO2014, NoCaps, COCO2017, and Flickr30k datasets (80k samples) using \texttt{lmms-eval}. Results compare LLaVA-1.6-7B/13B models with weighted-difference ($\Delta(q, p)$) and log-ratio ($S(q, p)$) scoring mechanisms. Performance measured by METEOR (M), ROUGE-L (R), and CIDEr (C); higher scores better. See~\ref{appx:details} for dataset details.}
    \resizebox{\textwidth}{!}{%
    \setlength{\tabcolsep}{4pt}
    \begin{tabular}{lc c c c c c c c c c c c c}
    \toprule
     & & \multicolumn{3}{c}{\texttt{COCO2014\_val}} & \multicolumn{3}{c}{\texttt{COCO2017\_val}} & \multicolumn{3}{c}{\texttt{NoCaps\_test}} & \multicolumn{3}{c}{\texttt{Flickr30k\_test}} \\ 
     Model & Score & M & R & C & M & R & C & M & R & C & M & R & C \\
    \midrule
    LLaVA-1.6-7b & - &  26.14 & 54.25 & 107.65 & 26.00 & 54.12 & 109.32 & 27.03 & 56.98 & 96.08 & 23.63 & 51.61 & 73.17 \\
    w/ \texttt{SVP} (C)   & $\Delta(q,p)$ & 28.74 & 56.69 & 111.98 & 28.74 & 56.69 & \tb{114.77} & 29.37 & 59.52 & \tb{104.79} & 25.62 & 53.25 & 75.98 \\
    w/ \texttt{SVP} (CVQ) & $\Delta(q,p)$ & \tb{29.26} & 56.62 & 111.38 & 29.24 & 56.67 & 114.72 & 30.07 & \tb{59.69} & 104.58 & \tb{26.34} & \tb{53.58} & \tb{77.68} \\
    w/ \texttt{SVP} (C)   & $S(q,p)$ & 28.64 & \tb{56.74} & \tb{112.45} & 28.57 & \tb{56.71} & 114.69 & 29.29 & 59.62 & 104.75 & 25.54 & 53.40 & 76.53 \\
    w/ \texttt{SVP} (CVQ) & $S(q,p)$ & 29.22 & 56.25 & 109.57 & \tb{29.25} & 56.34 & 113.08 & \tb{30.08} & 59.55 & 104.01 & 26.26 & 53.23 & 76.73\\
    \midrule
    LLaVA-1.6-13b & - & 24.67 & 52.03 & 99.39 & 24.72 & 52.23 & 102.04 & 25.44 & 54.93 & 88.13 & 22.21 & 48.78 & 66.68 \\
    w/ \texttt{SVP} (C)   & $\Delta(q,p)$ & 25.31 & 54.28 & 104.83 & 25.30 & 54.40 & 107.20 & 26.16 & 57.21 & 93.11 & 22.54 & 50.82 & 67.77\\
    w/ \texttt{SVP} (CVQ) & $\Delta(q,p)$ & 28.38 & \tb{56.71} & \tb{113.30} & \tb{28.49} & \tb{57.03} & \tb{117.23} & 28.94 & \tb{59.19} & \tb{102.32} & \tb{25.69} & \tb{53.61} & \tb{78.11} \\
    w/ \texttt{SVP} (C)    & $S(q,p)$ & 25.32 & 54.22 & 104.84 & 25.37 & 54.37 & 107.52 & 26.14 & 57.14 & 93.11 & 22.71 & 51.00 & 68.56 \\
    w/ \texttt{SVP} (CVQ)  & $S(q,p)$ & \tb{28.39} & 56.54 & 112.65 & 28.35 & 56.67 & 116.09 & \tb{28.96} & 59.14 & 101.93 & 25.59 & 53.25 & 77.00\\
     \bottomrule
    \end{tabular}
    }
    \label{tab:full-benchmark-appx}
\end{table}
\begin{table}[ht!]
    \centering
    \caption{\tb{Captioning Performance} on COCO2014, NoCaps, COCO2017, and Flickr30k datasets (80k samples) using \texttt{lmms-eval}. Comparing LLaVA-1.6-7B/13B models with weighted-difference ($\Delta(q, p)$) and log-ratio ($S(q, p)$) scoring. Evaluated using BLEU-1 to BLEU-4 (B1-B4); higher scores better.}
    \resizebox{\textwidth}{!}{%
    \setlength{\tabcolsep}{4pt}
    \begin{tabular}{lc c c c c c c c c c c c c c c c c}
    \toprule
     & & \multicolumn{4}{c}{\texttt{COCO2014\_val}} & \multicolumn{4}{c}{\texttt{COCO2017\_val}} & \multicolumn{4}{c}{\texttt{NoCaps\_test}} & \multicolumn{4}{c}{\texttt{Flickr30k\_test}} \\ 
     Model & Score & B4 & B3 & B2 & B1 & B4 & B3 & B2 & B1 & B4 & B3 & B2 & B1 & B4 & B3 & B2 & B1 \\
    \midrule
    LLaVA-1.6-7b  & - & 31.04 & 41.51 & 54.40 & 68.81 & 30.82 & 41.24 & 54.14 & 68.54 & 38.43 & 50.03 & 62.89 & 75.43 & 28.57 & 39.90 & 54.54 & 71.41 \\
    w/ \texttt{SVP} (C)   & $\Delta(q,p)$ & 32.29 & 44.25 & 59.33 & 76.16 & 32.61 & 44.50 & 59.44 & 76.09 & 41.05 & 54.12 & \tb{68.82} & 83.18 & 28.94 & 40.62 & 55.85 & 73.53 \\
    w/ \texttt{SVP} (CVQ) & $\Delta(q,p)$ & 31.69 & 43.50 & 58.46 & 75.52 & 32.01 & 43.72 & 58.53 & 75.53 & 40.88 & 53.78 & 68.49 & \tb{83.42} & 29.22 & 40.71 & 55.63 & 73.27 \\
    w/ \texttt{SVP} (C)   & $S(q,p)$ & \tb{32.75} & \tb{44.76} & \tb{59.86} & \tb{76.71} & \tb{32.82} & \tb{44.74} & \tb{59.78} & \tb{76.54} & \tb{41.15} & \tb{54.17} & 68.77 & 82.93 & \tb{29.59} & \tb{41.38} & \tb{56.68} & \tb{74.36} \\
    w/ \texttt{SVP} (CVQ) & $S(q,p)$ & 30.95 & 42.67 & 57.60 & 74.76 & 31.46 & 43.02 & 57.78 & 74.86 & 40.29 & 53.27 & 68.16 & 83.17 & 28.76 & 40.09 & 54.90 & 72.56 \\
    \midrule
    LLaVA-1.6-13b & - & 27.33 & 36.76 & 48.51 & 61.98 & 27.64 & 37.06 & 48.84 & 62.33 & 34.06 & 44.86 & 56.93 & 68.78   & 24.28 & 34.50 & 48.31 & 65.26 \\
    w/ \texttt{SVP} (C)   & $\Delta(q,p)$ & 29.97 & 39.65 & 51.34 & 63.79 & 29.96 & 39.65 & 51.37 & 63.76 & 37.28 & 48.33 & 59.97 & 70.31 & 27.15 & 37.88 & 51.83 & 67.78 \\
    w/ \texttt{SVP} (CVQ) & $\Delta(q,p)$ & \tb{33.65} & \tb{45.40} & \tb{59.99} & 76.45 & \tb{34.28} & \tb{45.90} & \tb{60.43} & \tb{76.71} & \tb{40.77} & \tb{53.66} & \tb{68.09} & \tb{82.25} & \tb{29.91} & \tb{41.92} & \tb{57.53} & \tb{75.55} \\
    w/ \texttt{SVP} (C) & $S(q,p)$ & 29.97 & 39.78 & 51.67 & 64.45 & 30.25 & 39.97 & 51.83 & 64.56 & 37.54 & 48.61 & 60.40 & 71.12 & 27.60 & 38.64 & 52.60 & 68.83 \\
    w/ \texttt{SVP} (CVQ) & $S(q,p)$ & 33.45 & 45.26 & 59.90 & \tb{76.47} & 34.00 & 45.59 & 60.10 & 76.50 & 40.35 & 53.24 & 67.81 & 82.17 & 29.40 & 41.39 & 57.03 & 75.18 \\
     \bottomrule
    \end{tabular}
    }
    \label{tab:full-benchmark-bleu-appx}
\end{table}

\paragraph{Captioning Tasks}
We conducted extensive captioning experiments using both 7B and 13B model architectures across three standard datasets: COCO2017, Flickr30k, and NoCaps (Fig.~\ref{bench:caption}). Our comprehensive evaluation, detailed in Tables~\ref{tab:full-benchmark-appx} and~\ref{tab:full-benchmark-bleu-appx}, spans four datasets and employs four widely-accepted metrics for assessing language generation and alignment quality. The evaluation encompasses over 80,000 samples, providing robust statistical evidence for our findings.

SVP demonstrates consistent superior performance across all datasets and metrics compared to existing methods. This comprehensive improvement underscores the effectiveness of our integrated sampling and feedback approach in enhancing image captioning capabilities. More fundamentally, these results validate our core hypothesis: strengthening vision-language alignment serves as a foundational principle for advancing VLM capabilities.

\clearpage
\paragraph{Referring Tasks}
\begin{wrapfigure}{r}{0.5\textwidth}
\centering
    \includegraphics[width=\linewidth]{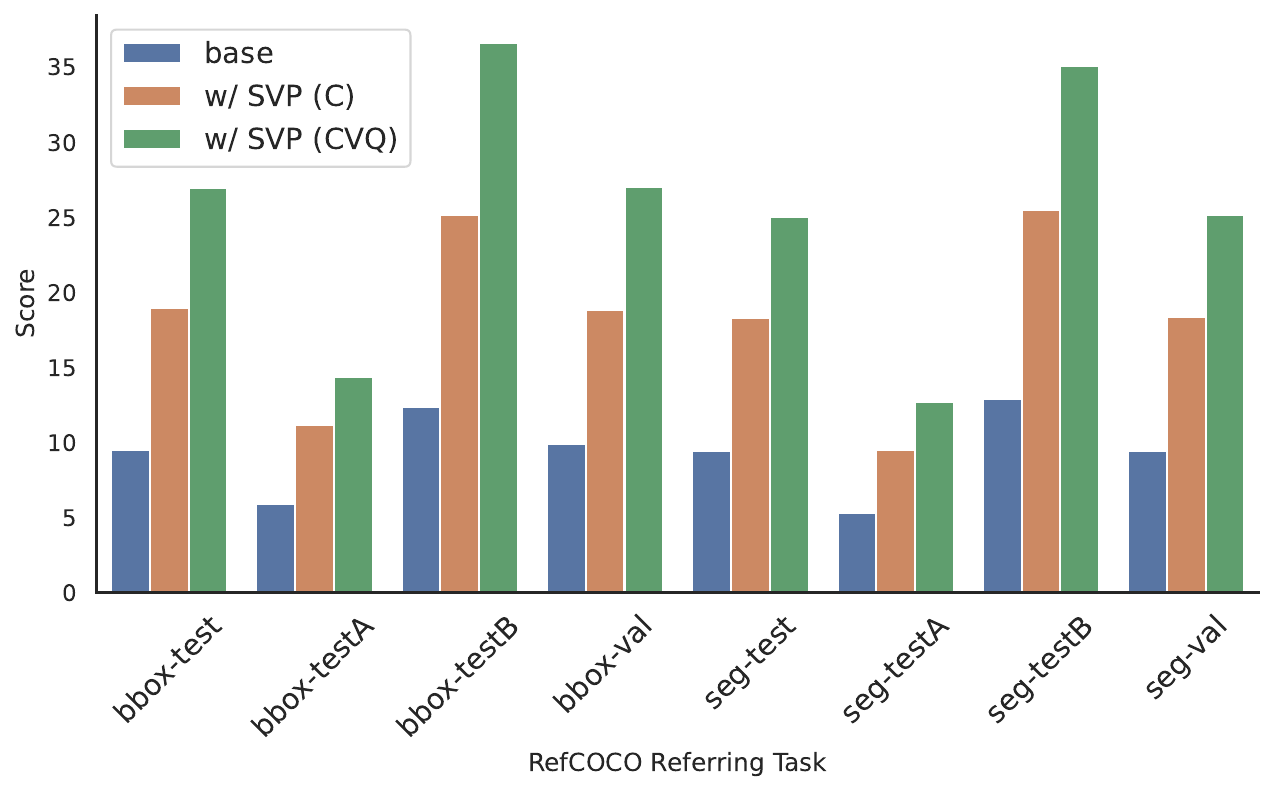}
    \caption{\textbf{Referring Expression Generation} on RefCOCO comparing base LLaVA-1.6-7b versus \texttt{SVP} (C) and \texttt{SVP} (CVQ) variants. CIDEr scores shown for detection (\texttt{bbox}) and segmentation (\texttt{seg}) on test/validation sets. \texttt{SVP} models outperform baseline without using bounding boxes. See Appx~\ref{referring-results} for RefCOCO+ and RefCOCOg results.}
    \label{fig:barplot-refcoco-results}
\end{wrapfigure}
We evaluate model performance on referring expression tasks, which require the VLM to generate descriptions for specific image regions (Fig.~\ref{fig:barplot-refcoco-results} and Appx~\ref{referring-results}). Our analysis compares four model variants: a baseline model, a model tuned without grounding (w/o $\vg$), a model incorporating visual grounding (w/ \texttt{SVP} (C)), and our full model with both grounding and visual queries (w/ \texttt{SVP} (CVQ)).

The results demonstrate that SVP substantially improves performance across all datasets and tasks. Most notably, SVP significantly enhances the base model's ability to understand and describe spatial relationships, particularly in cases where initial performance is poor. In fact, our enhanced models achieve performance levels approaching those of much larger 13B parameter models (Table~\ref{tab:main-results}).

A key insight emerges from these results: these improvements occur without direct access to grounding information (bounding boxes) during the adaptation phase. The grounding conditioning $\vg$ is utilized only during the "inner-loop" sampling to construct $q(\vz | \vc, \vg)$ (Fig.~\ref{fig:main-loop-fig}.$(iii)$), after which we adapt model parameters $\theta$ using only the refined visual projections $\vz$. This success in improving referring abilities without explicit grounding supervision suggests that enhanced modality alignment naturally leads to better spatial understanding in VLMs.
\begin{figure}[h]
    \centering
    \begin{minipage}[b]{0.32\textwidth}
        \centering
        \includegraphics[width=\linewidth]{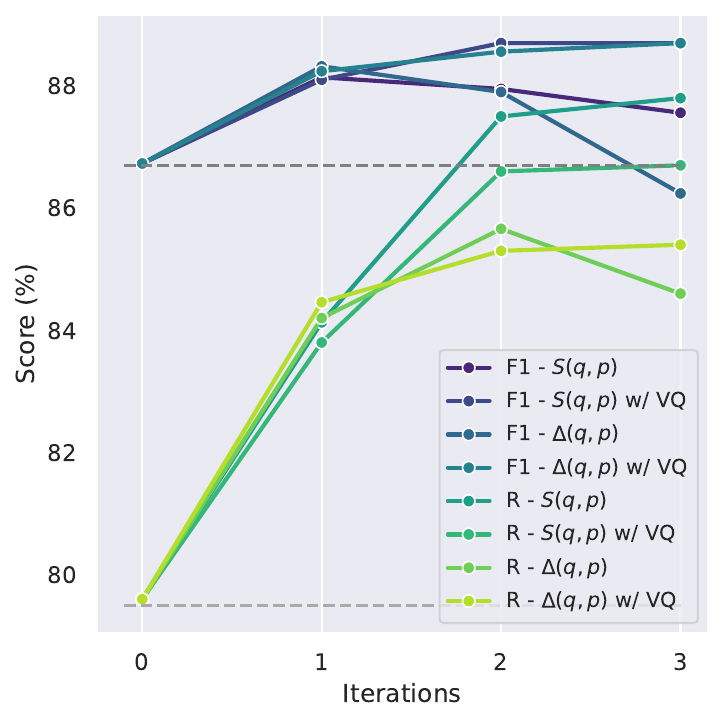}
    \caption{\tb{Iterations.} Hallucinations (F1) and object recall (R) results with $C=1000$ for $I=3$ iterations with $\Delta(q,p)$ and $S(q,p)$ scores.}
    \label{fig:iterations-recall}
    \end{minipage}
    \hfill
\begin{minipage}[b]{0.32\textwidth}
        \centering
        \includegraphics[width=\linewidth]{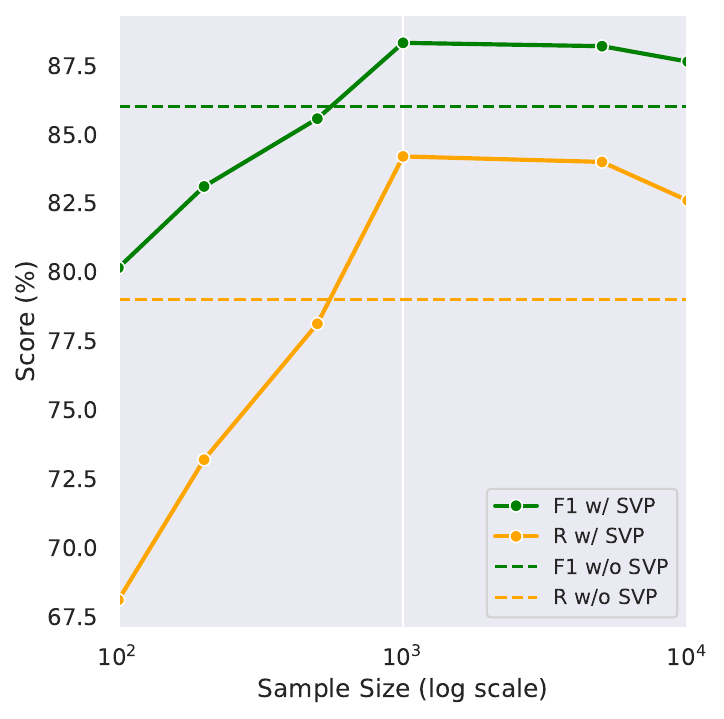}
    \caption{\tb{Sample Size.} Hallucinations (F1) and object recall (R) results with $I=1$ a single iteration increasing the sample size $C \in (0.1,0.2,0.5,1,5,10)k$.}
    \label{fig:sample size}
    \end{minipage}
\hfill
\begin{minipage}[b]{0.32\textwidth}
\centering
    \includegraphics[width=\linewidth]{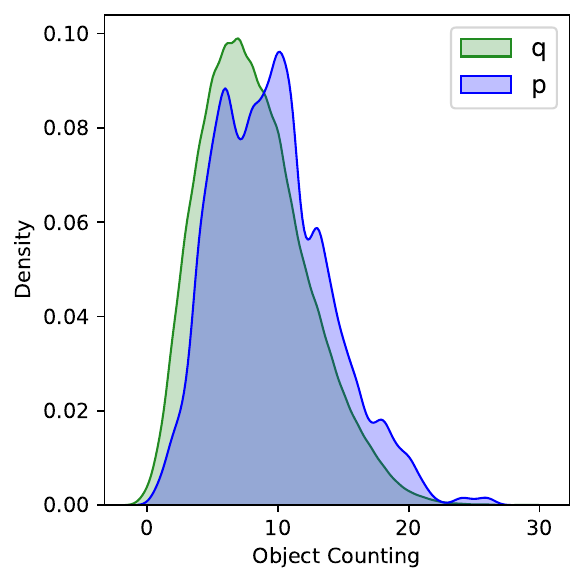}
    \caption{Distribution of groundable objects in captions from base model $p_{\theta}(\vz | \vc)$ and grounded model $q(\vz | \vc, \vg)$. \texttt{SVP} guided models generate fewer hallucinated objects.}
\label{fig:object-recall-main}
\end{minipage}
\end{figure}

\paragraph{Hallucination and Object Recall}
\begin{table}[htbp]
\begin{minipage}{0.48\textwidth}
\centering
    \caption{\textbf{Component Ablation.} Performance comparison of LLaVA-1.6-7b variants after one adaptation iteration: base model, fine-tuning without feedback, sampling with grounding (no scoring), grounding with scoring, and full \texttt{SVP} (grounding, scoring, visual queries). Results provide evidence of the importance of the \texttt{SVP}'s components for model performance.
    }
    \resizebox{\linewidth}{!}{%
    \setlength{\tabcolsep}{4pt}
    \begin{tabular}{l c c c c c  c c c c}
    \toprule
    Model & Grounding & Scoring & VQ & RefCOCO & Flickr30k & MMMU & POPE \\
    \midrule
     LLaVA     & - & - & - & 6.70 & 78.74 & 34.11 & 86.73 \\
     w/o \texttt{SVP} & \xmark & \xmark & \xmark  & 3.01 & 79.03 & 35.55 & 87.21\\
     w/ \texttt{SVP} & \cmark & \xmark & \xmark & 9.98 & 78.67 & 35.77 & 86.92 \\
     w/ \texttt{SVP} & \cmark & \cmark & \xmark & 18.15 & 83.49 & 36.44 & \textbf{88.33} \\
     w/ \texttt{SVP} & \cmark & \cmark & \cmark & \textbf{24.74} & \textbf{85.31} & \textbf{37.44} & 88.25 \\
    \bottomrule
    \end{tabular}
    }
    \label{tab:ablation-results}
\end{minipage}
\hfill
\begin{minipage}{0.48\textwidth}
    \centering
    \caption{
    \textbf{Preference Ablation}. Comparison between \texttt{SVP} and DPO~\cite{rafailov2024direct} for LLaVA-7b-OV with Qwen2 language model (higher is better). While DPO requires a learned reward model or human preference pairs, \texttt{SVP} uses only a small grounding model for feedback ($C=2000$, $K=10$, top $10\%$). Results show that DPO, while effective for general preference alignment, does not achieve the visual-language alignment gains of \texttt{SVP}.
    }
    \resizebox{\linewidth}{!}{%
    \setlength{\tabcolsep}{4pt}
    \begin{tabular}{l l c  c c c c c}
    \toprule
    Model   & Samples & SciQA & NoCaps & RefCOCO & MMBench & POPE \\
    \midrule
    w/ DPO   & $\geq 9.4k$  & 79.25 &  112.51 & 13.60 & 85.60 & \textbf{86.24}\\
    w/ \texttt{SVP} (C) & $\approx 2k$ & \textbf{83.89}   & \textbf{120.23} & \textbf{15.75} & \textbf{86.36} & 85.78\\
    \bottomrule
    \end{tabular}
    }
    \label{tab:ablation-dpo}
\end{minipage}
\end{table}

\begin{figure}[htbp]
\begin{minipage}{.48\textwidth}
    \centering
    \captionof{table}{
Text-to-Image alignment scores using LLaVA-1.6-7b and \texttt{SVP} VLMs at inference time without tuning. While typically used for evaluating AI-generated images, we compute ITM~\cite{li2023blip} and ImageReward~\cite{xu2024imagereward} to assess \emph{AI-generated captions for real images}. Though not standard metrics for vision-language alignment, these scores offer additional insight into text-image correspondence. Higher scores indicate better alignment.
}
\resizebox{\linewidth}{!}{%
\setlength{\tabcolsep}{4pt}
\begin{tabular}{lcccc}
\toprule
Model & Size & ITMScore (BLIP2) $\uparrow$ & ImageReward $\uparrow$\\
\midrule
     w/o \texttt{iSVP} & 7b  & 0.83  & 0.47  \\
     w/ \texttt{iSVP}  & 7b  & \textbf{0.89}  & \textbf{0.49}  \\
\midrule
     w/o \texttt{iSVP} & 13b & 0.82  & 0.44  \\
     w/ \texttt{iSVP}  & 13b  & \textbf{0.87} & \textbf{0.46} \\
\bottomrule
\end{tabular}
}
\label{tab:t2i-alignment}
\end{minipage}
\hfill
\begin{minipage}{.48\textwidth}
\centering
\begin{minipage}{.48\textwidth}
    \includegraphics[width=\linewidth]{img/flux/000000253742.jpg}
    \caption{Input image from COCO2017.}
\end{minipage}
\hfill
\begin{minipage}{.48\textwidth}
    \includegraphics[width=\linewidth]{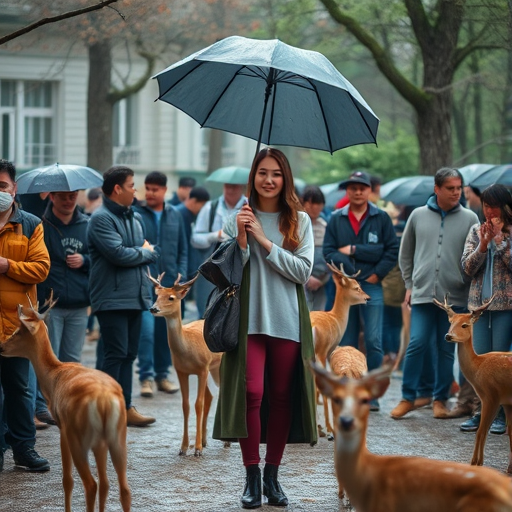}
    \caption{Text-to-Image generation using \texttt{iSVP} response.}
\end{minipage}
\caption{FLUX-schnell~\cite{flux} text-to-image generation using \texttt{iSVP} caption generation (\texttt{"A woman holding an umbrella stands among a group of people and deer"}) as input.}
\label{fig:flux-2s}
\end{minipage}
\end{figure}

We evaluate our model's hallucination rate (Tables~\ref{tab:hallucination-f1-results} and~\ref{tab:hallucination-acc-results}) and object recall (Figs.~\ref{bench:hallucination} and~\ref{fig:iterations-recall}), where object recall measures the model's ability to capture visual elements in its textual output. Our comparison includes HA-DPO~\cite{zhao2023beyond}, the leading DPO~\cite{rafailov2024direct} variant for hallucination reduction, and CSR~\cite{zhou2024calibrated}, an iterative self-rewarding VLM mechanism. For CSR, we evaluate both single-iteration performance and the best result across iterations $K \in [1:5]$.

SVP demonstrates substantial improvements across most model variants on the POPE dataset. With the 7B model, SVP raises the F1 score from 86.7\% to 88.3\%, achieving performance comparable to models five times larger (\ref{appx:model_size}). Similarly, the 13B model shows improvement from 86.2\% to 87.5\%.

Most impressively, when running SVP for three iterations with our scoring mechanism (Eq.~\ref{eq:score-kl}), object recall improves dramatically from 79\% to over 87\% (Fig.~\ref{fig:iterations-recall}). These results provide strong evidence that enhancing modality alignment through self-captioning and grounding feedback effectively reduces hallucinations without requiring specialized fine-tuning. This validates our core hypothesis while demonstrating SVP's ability to significantly improve the model's factual accuracy and reliability.

\paragraph{Ablations}
We conduct comprehensive ablation studies to analyze SVP's components and behavior. First, we examine the individual contributions of grounding, scoring, and visual queries (Table~\ref{tab:ablation-results}). We then investigate the impact of key hyperparameters: the number of iterations $I$ (Fig.~\ref{fig:iterations-recall}, Appx~\ref{fig:iterations}) and sample size $C$ (Fig.~\ref{fig:sample size}).
For scoring mechanisms, we evaluate both $\Delta(q,p)$ and $S(q,p)$ on the full captioning benchmark (Tables~\ref{tab:full-benchmark-appx} and~\ref{tab:full-benchmark-bleu-appx}). We also compare \texttt{SVP} against DPO using Qwen2~\cite{yang2024qwen2} as the language model on a subset of our benchmark (Table~\ref{tab:ablation-dpo}).
Additionally, we explore iSVP, a variant designed for inference-time adaptation without parameter tuning (Table~\ref{tab:t2i-alignment}, Fig.~\ref{fig:flux-2s}). Finally, we quantify the set of groundable objects for captions generated by guided versus prior distributions (Fig.~\ref{fig:object-recall-main}).

\section{Related Work}
\label{appx:related_work}

\paragraph{Improving Vision-Language Models}
Researchers have investigated explicit grounding in VLMs, primarily to address hallucinations~\cite{wan2024contrastive,favero2024multi}, with less focus on developing general paradigms for improving vision-language alignment. A common strategy involves incorporating grounding annotations into training data~\cite{peng2024grounding} for vision-centric VLMs~\cite{beyer2024paligemma, yuan2021florence, you2023ferret, zhang2025llava}. 

However, this annotation process is costly, time-consuming, and prone to errors. For instance, directly generating coordinate tokens as output is sample-inefficient, requiring billions of annotations even for small VLMs to develop a competitive detector~\cite{yuan2021florence}. While explicit supervision during fine-tuning can enhance alignment between visual and linguistic representations~\cite{liu2024improved, sun2023aligning}, these train-time methods necessitate large amounts of high-quality visual-text data and are resource-intensive to scale with human annotations.

Train-time techniques like Reinforcement Learning from Human Feedback (RLHF~\cite{christiano2017deep,ouyang2022training}) and Direct Preference Optimization (DPO~\cite{rafailov2024direct}), primarily used for aligning LLMs with human preferences, can be adapted to align VLM text outputs with visual inputs~\cite{zhou2024aligning,sun2023aligning,wang2024enhancing}. These approaches incorporate feedback and preferences during post-training but are limited by the need for reward signals~\cite{sun2023aligning}, curated preference pairs~\cite{zhu2024self,zhou2024aligning}, and AI feedback~\cite{wang2024enhancing}.

Test-time methods~\cite{wan2024contrastive}, such as Visual Contrastive Decoding~\cite{leng2024mitigating} and Multi-Modal Mutual-Information Decoding~\cite{favero2024multi}, aim to improve grounding at inference by leveraging differences between vision-conditional and unconditional models, without altering the model architecture or training. 
Woodpecker~\cite{yin2023woodpecker} proposes a five-step inference procedure to mitigate hallucination. While somewhat effective, these methods often require memory-intensive and computationally expensive inference, as well as model-specific heuristics, which limits their generalization and usability.

\paragraph{Grounding in Vision-Language Models}
Visual grounding can be conceptualized as the dual of text-image alignment. When viewed as a mechanism to elicit and organize information within Vision-Language Models (VLMs), it represents a form of alignment between visual and textual modalities, encompassing both representation and generation aspects.

The concept of grounding has deep roots in cognitive sciences~\cite{kiefer2013grounding,barsalou2008grounded,anderson2010neural,glenberg2002grounding}. In the context of computer vision, visual grounding can be seen as an extension of the classic closed-set detection problem~\cite{girshick2014rich,carion2020end,redmon2016you,zhang2022dino}. 

Traditional object detection tasks involve regressing bounding box coordinates and assigning class labels to regions within an input image. While leveraging curated benchmark datasets~\cite{lin2014microsoft} has led to rapid improvements in precision and speed, this approach has been constrained by predefined class sets. Scaling to a larger number of classes and adapting to varying detection granularities have proven challenging~\cite{gupta2019lvis,dave2021evaluating}.

Visual grounding inverts this paradigm by using the set of classes as input and employing a vision-language model to assign bounding boxes to each element in the input. This concept can be further generalized to accommodate captions, descriptions, and various forms of textual input. Contrastive models such as GLIP~\cite{li2022grounded} and GroundingDINO~\cite{liu2023grounding} offer flexible, generalized detection models that enhance spatial understanding~\cite{yin2023woodpecker} and serve as foundations for a wide range of tasks.
Moreover, auto-regressive VLMs have been developed to perform grounding and referring tasks~\cite{yuan2021florence, you2023ferret, peng2024grounding,tong2024cambrian}, further expanding the capabilities of these models in bridging visual and linguistic information.

\paragraph{Self-improvement in Vision and Language Models}
Self-improving autonomous learners have been a long standing goal of the AI field~\cite{schmidhuber1987evolutionary, schmidhuber1999general}. In the context of Vision-Language Models (VLMs), self-improvement can be conceptualized as a form of self-play~\cite{silver2016mastering,silver2017mastering}, where the model enhances its performance through sampling and external feedback mechanisms~\cite{anthony2017thinking}.
The advent of Large Language Models (LLMs)~\cite{brown2020language,achiam2023gpt} has necessitated novel approaches to self-improvement, given the challenges in defining explicit feedback for natural language trajectories. 

Reinforcement Learning from Human Feedback (RLHF)~\cite{ouyang2022training} and Reinforcement Learning from AI Feedback (RLAIF)~\cite{bai2022constitutional} have emerged as prominent mechanisms. These methods score samples from the base model and select preferred outputs based on specific criteria, such as human preferences in chat interactions. Both approaches learn preference or reward models from human or AI feedback, and these concepts have been successfully adapted to VLMs~\cite{sun2023aligning,favero2024multi}.

Further developments in this field include using rewards for ranking~\cite{dong2023raft} and implicitly specifying preferences through positive and negative pairs~\cite{rafailov2024direct}. Alignment can also be achieved through AI distillation~\cite{sudalairaj2024lab, chiang2023vicuna} and self-refinement techniques~\cite{kang2024self, kang2024self2, wang2022self,sun2024principle}.

A recent class of algorithms for self-improvement involves iterative processes~\cite{zelikman2022star, zelikman2024quiet,anthony2017thinking, gulcehre2023reinforced} that leverage feedback to enhance downstream tasks and reasoning chains~\cite{wei2022chain} in LLMs. Moreover, feedback can be incorporated at inference time~\cite{madaan2024self} and even utilize the model's own capabilities as evaluator~\cite{yao2022react,shinn2024reflexion}.
These methods can be seen as instantiating meta-learning algorithms.

Meta-learning~\cite{schaul2010metalearning,hospedales2020meta,finn2017model}, often described as learning to learn, plays a crucial role in the development of self-improving AI systems. This approach aims to create models that can adapt quickly to new tasks by leveraging knowledge from previously learned tasks~\cite{schmidhuber1987evolutionary, schmidhuber1999general}. In the context of VLMs and LLMs, meta-learning techniques have been explored to enhance model adaptability and generalization across diverse domains. For instance, few-shot in-context learning methods~\cite{brown2020language,xie2021explanation,akyurek2022learning} demonstrate how large models can rapidly adapt to new tasks with minimal task-specific examples.

\section{Conclusions and Limitations}
We present \texttt{SVP}, a novel method that leverages self-captioning and grounding feedback to enhance VLMs without requiring additional annotations. Our approach significantly improves captioning quality, referring expression generation, hallucination control, and object recall while maintaining strong performance on VQA and multitasking benchmarks. These results demonstrate \texttt{SVP}'s potential to unlock latent VLM capabilities, advancing toward more robust real-world applications.
However, \texttt{SVP} has notable limitations: it requires VLMs capable of in-context learning, needs multiple samples per input, and depends on grounding model quality. The method may not benefit tasks without spatial components or those requiring specialized knowledge, such as VQA. Additionally, without injecting new information, its applicability to knowledge-intensive tasks remains uncertain without external data.

\small
\clearpage
\bibliographystyle{abbrv}
\bibliography{biblio.bib}

\appendix
\normalsize
\clearpage
\clearpage
\section{Sampling-based Visual Projection Workflow}
\label{appx:workflow}
\begin{figure}[ht!]
\centering
\begin{contextbox}[width=0.49\textwidth,box align=top,before=,after=\hfill]{Visual Input and Text Instruction.}
\begin{promptbox}{}
\textbf{Instruction:} 
Please describe the content of this image as detailed as possible.
\\
\\
\\
\\
\\
\\
\\
    \begin{center}
    \includegraphics[width = 0.45\linewidth, keepaspectratio]
    {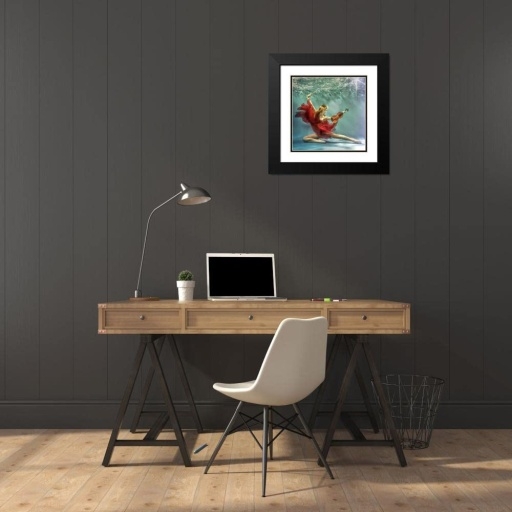}
    \quad
    \includegraphics[width = 0.45\linewidth, keepaspectratio]{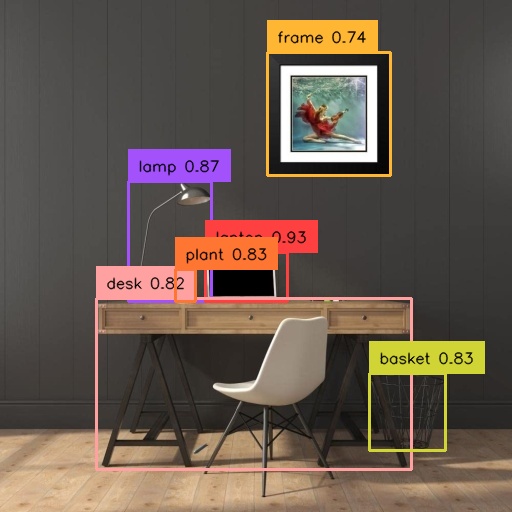}  
    \end{center}
    
\tcblower

The image depicts a modern home office setup. At the center of the scene is a wooden desk with a sleek, minimalist design. The desk features a flat surface with a built-in drawer and a small shelf on the left side. \hle{On the right side of the desk, there is a laptop open and ready for use}.

Above the desk, mounted on the wall, is a framed artwork of a ballerina in a dynamic pose, adding a touch of elegance and movement to the space. The artwork is encased in a black frame that contrasts with the wall's dark gray color.

\hle{To the left of the desk, there is a floor lamp with a modern design, providing additional lighting}. The lamp has a tall, slender base and a curved arm with a white shade.

On the desk, there is a small plant in a white pot, adding a touch of greenery and life to the space. The plant is placed on the left side of the desk, near the laptop.
Underneath the desk, there is a black wire basket, which is likely used for cable management and storage. 
\\
\\
\hle{Note: missing chair, missing background.}
\\
\\
\\
\textbf{Grounded Objects}: \texttt{desk (0.82). laptop (0.93). artwork (0.64). frame (0.74). lamp (0.87). plant (0.83). pot (0.68). basket (0.83).}

\end{promptbox}
\end{contextbox}%
\begin{contextbox}[width=0.49\textwidth,box align=top,before=,after=]{Grounding}
\begin{promptbox}{(Posterior Sampling)}
\textbf{Instruction:} 
Please describe the content of this image as detailed as possible.
\\
\textbf{Context:}
Use the following context. The list of numbers are bounding boxes for the objects.
Check if object and bounding boxes connections are correct. Check if there are additional objects in the image and describe them.

format: \texttt{cxcywh}. \\
boxes:\texttt{
lamp   [0.33,0.47,0.16,0.23],
laptop [0.48,0.54,0.16,0.09],
desk   [0.49,0.75,0.61,0.33],
plant  [0.36,0.56,0.04,0.06],
frame  [0.64,0.22,0.24,0.24],
basket [0.80,0.80,0.15,0.15].
}
    \begin{center}
    \includegraphics[width = 0.45\linewidth, keepaspectratio]{img/intro/input_image.jpg}  
    \quad
    \includegraphics[width = 0.45\linewidth, keepaspectratio]{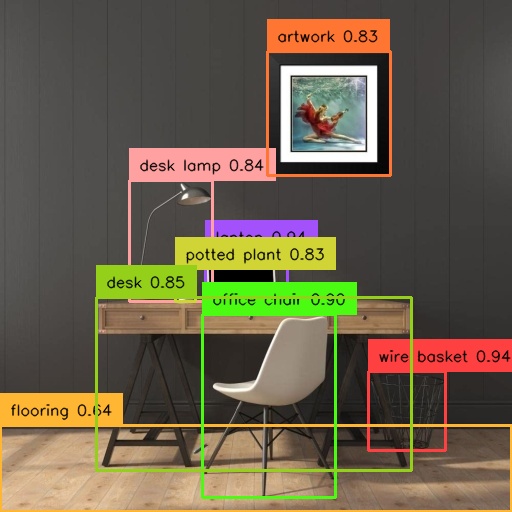}    
    \end{center}
    
\tcblower

The image depicts a modern workspace setup against a \hl{dark gray wall with horizontal wood paneling}. On the wall, there is a framed artwork featuring a figure in a red dress with a blue background, possibly a painting or a print. Below the artwork, there is a desk with a light wood finish. \hl{The desk has a rectangular shape with a flat top and is supported by four black metal legs.}

On the desk, there is a laptop with a black screen and a silver body, positioned towards the center. \hl{To the left of the laptop, there is a white desk lamp with a curved neck and a silver base}, turned off. Next to the lamp, there is a small white potted plant with green leaves.

\hl{In front of the desk, there is a white office chair with a modern design, featuring a curved backrest and armrests. The chair is positioned facing the desk, with the seat slightly pulled out.}

To the right of the desk, there is a black wire basket with a handle, placed on the floor. \hl{The flooring appears to be a light-colored wooden plank.} The overall lighting in the room is soft and ambient, suggesting a calm and focused work environment.

\textbf{Grounded Objects:} \texttt{artwork (0.83). desk (0.85). laptop (0.94). desk lamp (0.84). potted plant (0.83). office chair (0.90). wire basket (0.94). flooring (0.64).}
\end{promptbox}
\end{contextbox}

\caption{\textbf{Sampling VLMs with and without Grounding Feedback.}
Incorporating grounding feedback helps VLMs to focus on factual information and better describe the details in the input image. We use GroundingDINO~\cite{liu2023grounding}, an open-set grounding model, to obtain the conditioning information. When the predicted bounding boxes overlap above a certain threshold, we select the box with the highest score, following a standard non-maximum-suppression approach.
By leveraging this grounding feedback, the model is better able to specify the entities and relationships between the objects in the image, leading to an improved parsing of the visual information. This results in more accurate and detailed descriptions, such as identifying a \texttt{desk lamp} instead of a \texttt{floor lamp}, mentioning an \texttt{office chair}, describing the \texttt{flooring} in the background, and differentiating between an \texttt{artwork} and a simple \texttt{frame}, or a \texttt{potted plant} and a generic \texttt{plant}. More visualizations in~\ref{appx:examples}.
}
\label{fig:intro-figure}
\end{figure}

\clearpage
\section{Visual Grounding and Text-Image Alignment}
\label{appx:text-to-image}
A key question in vision-language modeling is whether and how vision-language alignment relates to text-to-image generation capabilities.
Visual grounding serves as the dual of text-image alignment, functioning as a mechanism to structure cross-modal information within VLMs. This alignment forms a critical foundation in both LLMs and VLMs, aiming to create a unified representational space for effective multi-modal reasoning. The process typically involves multiple stages: pre-training on large datasets, task-specific fine-tuning, and advanced techniques like preference tuning and contrastive learning.
Strong modality alignment is crucial for VLMs to effectively integrate visual and textual information. When properly aligned, models can better process, understand, and generate coherent multi-modal responses, leading to improved performance across diverse applications.

\begin{figure}[ht!]    
\centering
\begin{subfigure}{0.48\linewidth}
    \includegraphics[width=\linewidth]{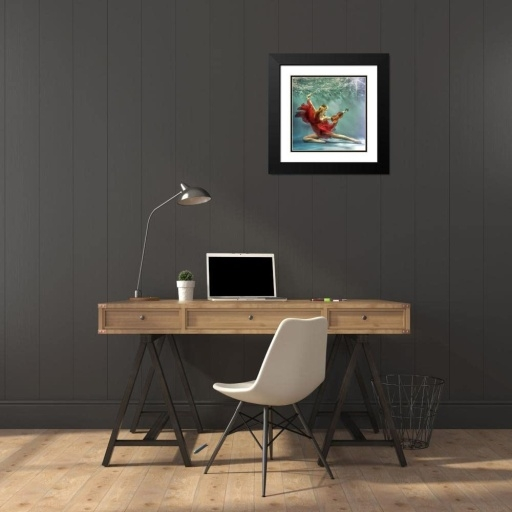}
    \caption{Input image
    }
    \label{fig:input}
\end{subfigure}

\begin{subfigure}{0.48\linewidth}
    \includegraphics[width=\linewidth]{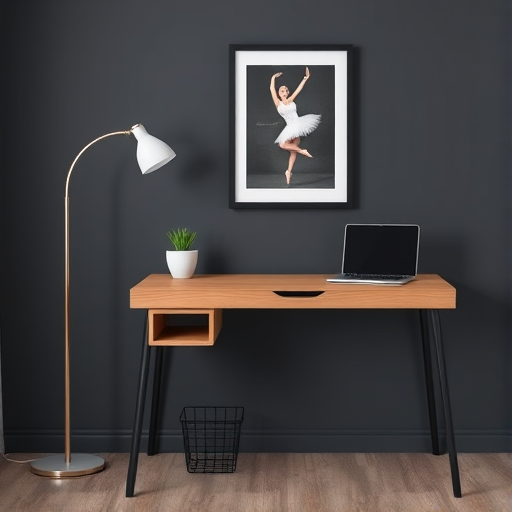}
    \caption{Text-to-Image generation using base VLM response - $\vz \sim p_{\theta}(\vz | \vc)$. See left side~\ref{fig:intro-figure}.
    }
    \label{fig:prior-flux}
\end{subfigure}
    \hfill
\begin{subfigure}{0.48\linewidth}
    \includegraphics[width=\linewidth]{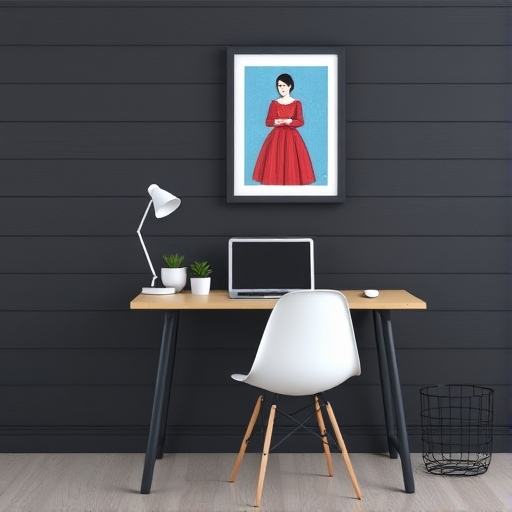}
    \caption{Text-to-Image generation using grounded VLM response - $\vz \sim q(\vz | \vc, \vg)$.
    See right side~\ref{fig:intro-figure}.
    }
    \label{fig:posterior-flux}
\end{subfigure}
\caption{FLUX-schnell~\cite{flux} text to image generation using the original VLM response (left) and the response leveraging grounding (right) as input. We generated a single image without multiple attempts or selective filtering. The comparison clearly illustrates that the grounding-enhanced response produces more accurate and reliable generation outcomes.}
    \label{fig:flux}
\end{figure}

\begin{figure}    
\centering
\begin{subfigure}{0.48\linewidth}
    \includegraphics[width=\linewidth]{img/flux/000000466567.jpg}
    \caption{Input image from \texttt{coco2017\_cap\_val\_lite}. Image id: \texttt{000000466567}. Target Captions (provided as ground truth): 
    ["A tree with a donut as an ornament", 
    "A plastic tree with a doughnut hanging by a strip of red ribbon. ", 
    "A Christmas ornament is a donut with a squirrel on it.", 
    "A doughnut hanging from a Christmas tree as a decoration.", 
    "a donut being used as an ornament for a chistmas tree"]
    }
    \label{fig:input-1}
\end{subfigure}

\begin{subfigure}{0.48\linewidth}
    \includegraphics[width=\linewidth]{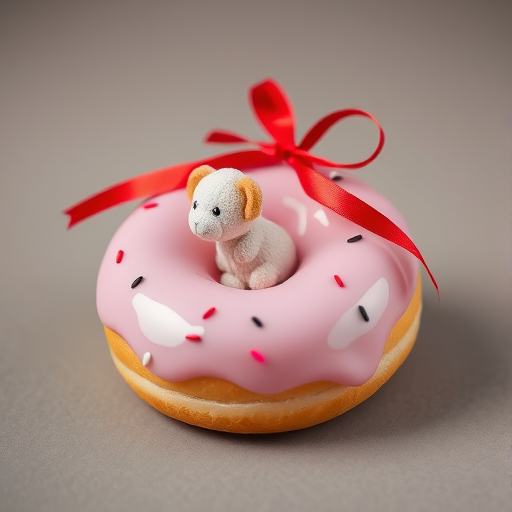}
    \caption{Text-to-Image generation using base VLM response - $\vz \sim p_{\theta}(\vz | \vc)$: 
    \texttt{"A donut with a red ribbon and a small toy animal on it" \\} for image (a).
    }
    \label{fig:prior-flux-1}
\end{subfigure}
    \hfill
\begin{subfigure}{0.48\linewidth}
    \includegraphics[width=\linewidth]{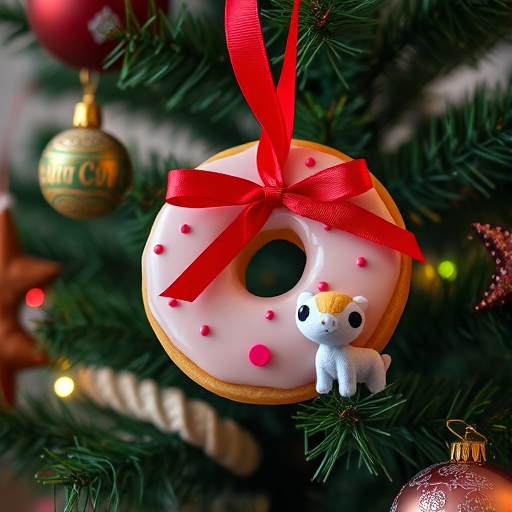}
    \caption{Text-to-Image generation using grounded VLM response - $\vz \sim q(\vz | \vc, \vg)$: 
    \texttt{"A donut with a red ribbon and a small toy animal on a Christmas tree"} for image (a).
    }
    \label{fig:posterior-flux-1}
\end{subfigure}
\caption{FLUX-schnell~\cite{flux} text to image generation using the base VLM response (left) and the response using \texttt{iSVP} (right) as input.
We generated a single image without multiple attempts or selective filtering. }
    \label{fig:flux-1}
\end{figure}

\begin{figure}    
\centering
\begin{subfigure}{0.48\linewidth}
    \includegraphics[width=\linewidth]{img/flux/000000253742.jpg}
    \caption{Input image from \texttt{coco2017\_cap\_val\_lite}. Image id: \texttt{000000253742}. Target Captions (provided as ground truth): 
    ["A woman standing next to a herd of animals.", "a woman holding an umbrella at the park", 
    "A woman standing in the rain with an umbrella with a herd of deer behind her.", 
    "On a rainy day at the zoo umbrellas are frequently seen.", 
    "Several people holding umbrellas and standing next to deer."]
    }
    \label{fig:input-2}
\end{subfigure}

\begin{subfigure}{0.48\linewidth}
    \includegraphics[width=\linewidth]{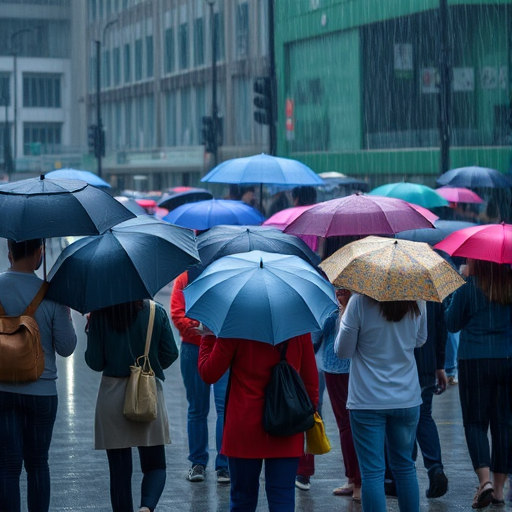}
    \caption{Text-to-Image generation using base VLM response - $\vz \sim p_{\theta}(\vz | \vc)$:
    \texttt{"A group of people holding umbrellas and standing in the rain"} for image (a).
    }
    \label{fig:prior-flux-2}
\end{subfigure}
    \hfill
\begin{subfigure}{0.48\linewidth}
    \includegraphics[width=\linewidth]{img/flux/generated_image_2_svp.png}
    \caption{Text-to-Image generation using grounded VLM response - $\vz \sim q(\vz | \vc, \vg)$: 
    \texttt{"A woman holding an umbrella stands among a group of people and deer"} for image (a).
    }
    \label{fig:posterior-flux-2}
\end{subfigure}
\caption{FLUX-schnell~\cite{flux} text to image generation using the base VLM response (left) and the response using \texttt{iSVP} (right) as input. We generated a single image without multiple attempts or selective filtering. }
    \label{fig:flux-2}
\end{figure}

\clearpage
\section{SVP Algorithms}

\begin{algorithm}[H]
\caption{Sampling-based Visual Projection (SVP) w/ log-ratio based scoring $S(q,p)$}
\label{alg:svp-s}
\begin{algorithmic}[1]
\Require
    \State Base VLM $p_\theta(\vz_p | \vc)$
    \State Grounding model $g(\vz, \vc)$
    \State Scoring function $S(q,p)$
    \State Seed images $\mathcal{C} = \{\vc_c\}^{C}_{c=1}$
    \State Samples per image $K$, top-k ratio $k$
    \State Learning rate $\alpha$, iterations $I$, vocabulary size $V$, grounded sequence length $T$
\Ensure Updated model parameters $\theta_1 \leftarrow \theta$

\For{iteration $i=1$ to $I$}
    \State $\mathcal{D} \leftarrow \{\}$ \Comment{Initialize dataset}
    
    \For{each image $\vc \in \mathcal{C}$}
        \State $\vZ_q \leftarrow \{\}$ \Comment{Sample buffer}
        
        \For{$j=1$ to $K$}
            \State $\vz^{j}_p \sim p_{\theta_{i}}(\vz|\vc)$ \Comment{Sample from prior}
            \State $\vg_j \leftarrow g(\vz^{j}_p, \vc_v)$ \Comment{Grounding feedback}
            \State $\vz^{j}_q \sim q(\vz|\vc,\vg_j)$ \Comment{Sample with grounding}
            \State $\vZ_q \leftarrow \vZ_q \cup \{\vz^{j}_q\}$
        \EndFor
        
        \For{$\vz_q \in \vZ_q$}
            \State $S(q, p)_{\vz_q} \leftarrow \sum\nolimits_{t=1}^T \sum\nolimits_{v=1}^V w_{v,t}[\log q_{v,t} - \log p_{v,t}]$
            \State $s_q \leftarrow S(q, p_{\theta_{i}})_{\vz_q}$ \Comment{Score samples}
        \EndFor
        
        \State $S_k \leftarrow k\text{-th highest score in } \{s_q\}$
        \State $\vZ^{*} \leftarrow \{\vz_q : s_q \geq S_k\}$ \Comment{Select top-k}
        \State $\mathcal{D} \leftarrow \mathcal{D} \cup \{(\vc, \vz) : \vz \in \vZ^{*}\}$
    \EndFor
    
    \For{minibatch $B \subset \mathcal{D}$}
        \State $\mathcal{L}(\theta) \leftarrow -\frac{1}{|B|} \frac{1}{|k(\vc)|} \sum_{(\vc,\vz)\in B} \log p_\theta(\vz|\vc)$
        \State $\theta_i \leftarrow \theta - \alpha\nabla_\theta\mathcal{L}$ \Comment{Update parameters}
    \EndFor
    \State $\theta_{i + 1} = \theta_{i}$
\EndFor

\Return $p_{\theta_{I}}(\vz | \vc)$
\end{algorithmic}
\end{algorithm}

\begin{algorithm}[H]
\caption{Sampling-based Visual Projection (SVP) w/ weighted difference based scoring $\Delta(q,p)$}
\label{alg:svp-d}
\begin{algorithmic}[1]
\Require
    \State Base VLM $p_\theta(\vz_p | \vc)$
    \State Grounding model $g(\vz, \vc)$
    \State Scoring function $\Delta(q,p)$
    \State Seed images $\mathcal{C} = \{\vc_c\}^{C}_{c=1}$
    \State Samples per image $K$, top-k ratio $k$
    \State Learning rate $\alpha$, iterations $I$, vocabulary size $V$, grounded sequence length $T$
\Ensure Updated model parameters $\theta_1 \leftarrow \theta$

\For{iteration $i=1$ to $I$}
    \State $\mathcal{D} \leftarrow \{\}$ \Comment{Initialize dataset}
    
    \For{each image $\vc \in \mathcal{C}$}
        \State $\vZ_q \leftarrow \{\}$ \Comment{Sample buffer}
        
        \For{$j=1$ to $K$}
            \State $\vz^{j}_p \sim p_{\theta_{i}}(\vz|\vc)$ \Comment{Sample from prior}
            \State $\vg_j \leftarrow g(\vz^{j}_p, \vc_v)$ \Comment{Grounding feedback}
            \State $\vz^{j}_q \sim q(\vz|\vc,\vg_j)$ \Comment{Sample with grounding}
            \State $\vZ_q \leftarrow \vZ_q \cup \{\vz^{j}_q\}$
        \EndFor
        
        \For{$z_q \in Z_q$}
            \State $\Delta(q, p)_{\vz_q} = \sum\nolimits^{T}_{t=1} \sum\nolimits^{V}_{v=1} w^{q}_{v,t} \log q_{v,t} 
    - \sum\nolimits^{T}_{t=1} \sum\nolimits^{V}_{v=1} w^{p}_{v,t} \log {p_{\theta}}_{v,t}$ 
            \State $s_q \leftarrow \Delta(q, p)_{\vz_q}$ \Comment{Score samples}
        \EndFor
        
        \State $S_k \leftarrow k\text{-th highest score in } \{s_q\}$
        \State $\vZ^{*} \leftarrow \{\vz_q : s_q \geq S_k\}$ \Comment{Select top-k}
        \State $\mathcal{D} \leftarrow \mathcal{D} \cup \{(\vc, \vz) : \vz \in \vZ^{*}\}$
    \EndFor
    
    \For{minibatch $B \subset \mathcal{D}$}
        \State $\mathcal{L} \leftarrow -\frac{1}{|B|}\sum_{(\vc,\vz)\in B} \log p_\theta(\vz|\vc)$
        \State $\theta \leftarrow \theta - \alpha\nabla_\theta\mathcal{L}$ \Comment{Update parameters}
    \EndFor
    \State $\theta_{i + 1} = \theta_{i}$
\EndFor

\Return $p_{\theta_{I}}(\vz | \vc)$
\end{algorithmic}
\end{algorithm}

\clearpage
\section{Additional Experiments}
\label{appx:experiment_additional}

\subsection{Referring Tasks}
\begin{table}[ht!]
\centering
\caption{Evaluation of referring expression generation on various \texttt{RefCOCO}, \texttt{RefCOCO+}, and \texttt{RefCOCOg} datasets using LLaVA-1.6-7b. 
The experiment compares the performance of different models, including a base model, a model without visual grounding (w/o $\vg$), a model with Visual Projections (w/ \texttt{SVP} (C)), and a model with \texttt{SVP} and Visual Query (w/ \texttt{SVP} (CVQ) ). 
The performance is measured using the CIDEr score on bounding box (\texttt{bbox}) and segmentation (\texttt{seg}) referring task on the test and validation sets for each dataset. The results show that \texttt{SVP} models significantly outperform the base and w/o $\vg$ models, indicating the importance of visual grounding for referring tasks. Notice that the adapted models do not have access to the bounding boxes during fine-tuning.}
\begin{tabular}{l cc cc cc}
\toprule
& & & \multicolumn{2}{c}{$\Delta(q,p)$} & \multicolumn{2}{c}{$S(q,p)$} \\
     & base  & w/o $\vg$ & w/ \texttt{SVP} (C) & w/ \texttt{SVP} (CVQ) & w/ \texttt{SVP} (C) & w/ \texttt{SVP} (CVQ)     \\
     \midrule
&       &      &      && \\
\emph{RefCOCO}    &       &      &      &&             \\
\midrule
\texttt{bbox-test}  & 9.53  & 3.57 & 18.99 & 26.96 & 20.74  & 25.52 \\
\texttt{bbox-testA} & 5.91  & 1.59 & 11.14 & 14.37 & 12.33  & 14.00 \\
\texttt{bbox-testB} & 12.35 & 6.27 & 25.13 & 36.65 & 27.64  & 34.71 \\
\texttt{bbox-val}   & 9.93  & 3.95 & 18.84 & 27.01 & 21.07  & 25.76 \\
\texttt{seg-test}   & 9.46  & 3.70 & 18.27 & 25.02 & 19.68  & 23.89 \\
\texttt{seg-testA}  & 5.32  & 1.37 & 9.48  & 12.67 & 10.95  & 11.70 \\
\texttt{seg-testB}  & 12.92 & 6.44 & 25.49 & 35.08 & 26.61  & 33.28 \\
\texttt{seg-val}    & 9.44  & 4.02 & 18.35 & 25.15 & 19.60  & 23.95 \\
&       &      &      && \\
\emph{RefCOCO+}   &       &      &      &&           \\
\midrule
\texttt{bbox-testA} & 6.68  & 2.16 & 12.25 & 16.93 & 14.05 & 16.44 \\
\texttt{bbox-testB} & 10.98 & 6.21 & 23.31 & 33.02 & 25.46 & 30.98 \\
\texttt{bbox-val}   & 9.57  & 3.68 & 18.00 & 26.67 & 20.70 & 25.35 \\
\texttt{seg-testA}  & 5.98  & 1.86 & 10.74 & 13.97 & 12.30 & 13.56 \\
\texttt{seg-testB}  & 11.75 & 6.45 & 23.67 & 31.25 & 24.59 & 29.70 \\
\texttt{seg-val}    & 9.19  & 3.90 & 17.15 & 24.31 & 19.13 & 23.81 \\
&       &      &      && \\
\emph{RefCOCOg} &       &      &      &&                      \\
\midrule
\texttt{bbox-test} & 20.27 & 13.68 & 47.74 & 59.74 & 50.89 & 56.79 \\
\texttt{bbox-val}  & 19.70 & 12.16 & 47.69 & 59.65 & 50.73 & 56.81 \\
\texttt{seg-test}  & 18.76 & 12.90 & 45.23 & 54.39 & 47.51 & 51.18 \\
\texttt{seg-val}   & 18.77 & 12.55 & 45.45 & 54.01 & 46.93 & 50.77 \\
\bottomrule
\end{tabular}
\label{referring-results}
\end{table}

\clearpage
\subsection{Iteration Ablation}
\label{appx:iter_appx}
\begin{figure}[ht!]
\centering
\includegraphics[width=\textwidth,height=.8\textheight,keepaspectratio]{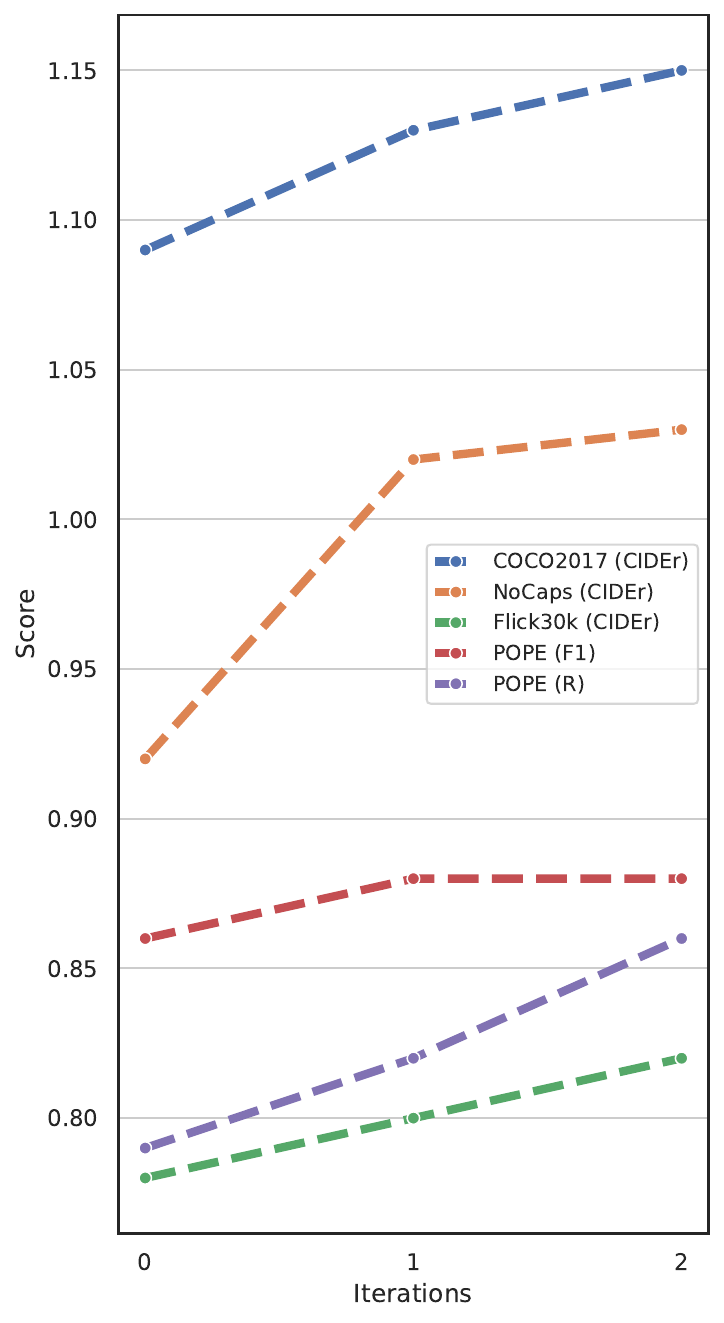}
    \caption{\texttt{SVP} effectively boosts captioning performance and reduces hallucinations on benchmark tasks using LLaVA-1.6-7b as base model. The second iteration of \texttt{SVP} adaptation leads to significant improvements compared to the initial round, underscoring the value of this technique for enhancing visual-language model capabilities. However, the gains tend to plateau after the second iteration, suggesting diminishing returns from further fine-tuning.
    }
    \label{fig:iterations}
\end{figure}

\clearpage
\subsection{Model Size Ablation}
\label{appx:model_size}
\begin{figure}[ht!]
\centering
\includegraphics[width=\textwidth,height=.8\textheight,keepaspectratio]{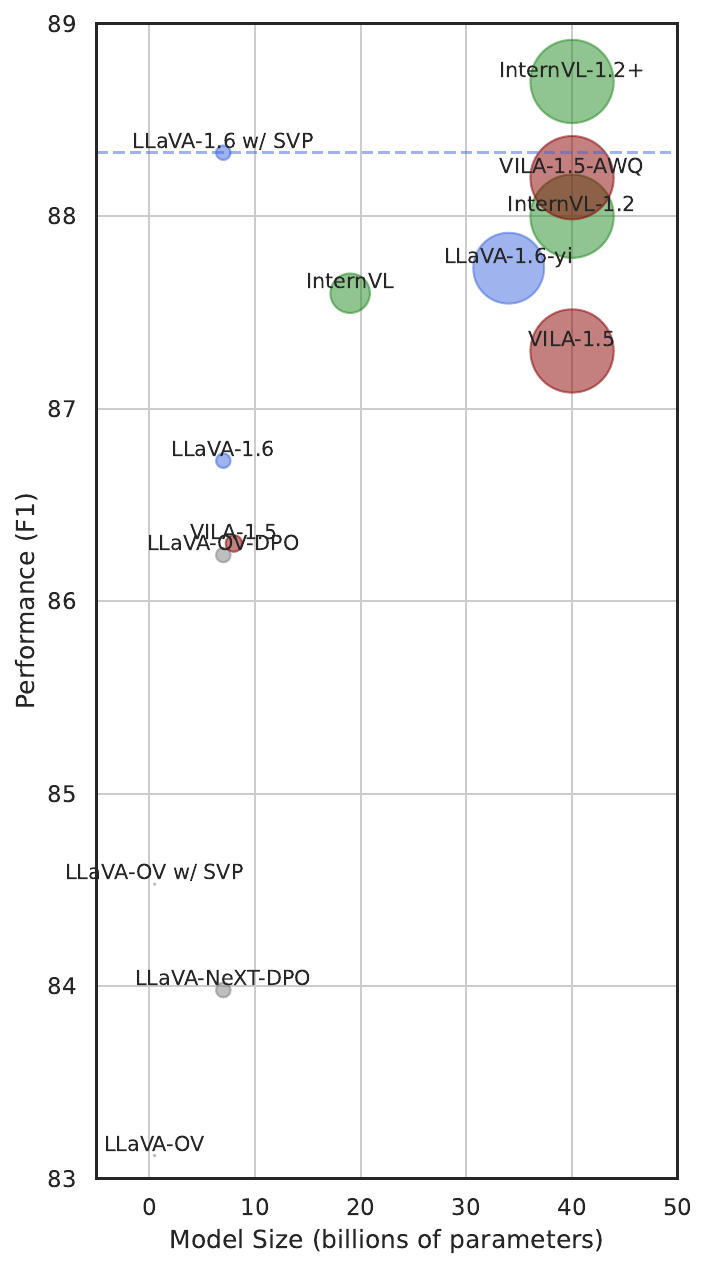}
    \caption{Model size comparison using the F1 metric on the POPE dataset.
    \texttt{SVP} improves the base model and achieves better or comparable performance with models five times larger.
    }
    \label{fig:model-size-appx}
\end{figure}

\clearpage
\subsection{Object Grounding Ablation}
\begin{figure}[ht!] 
\centering
\begin{subfigure}{0.32\linewidth}
    \includegraphics[width=\linewidth]{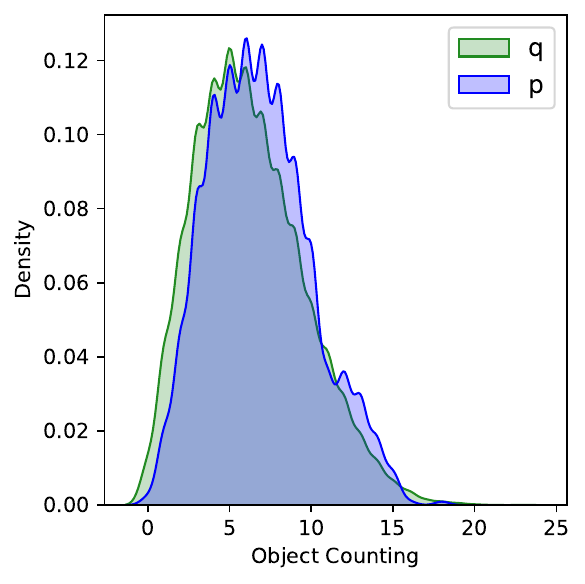}
    \caption{LLaVA-1.5-7b.}
\end{subfigure}
    \hfill
\begin{subfigure}{0.32\linewidth}
    \includegraphics[width=\linewidth]{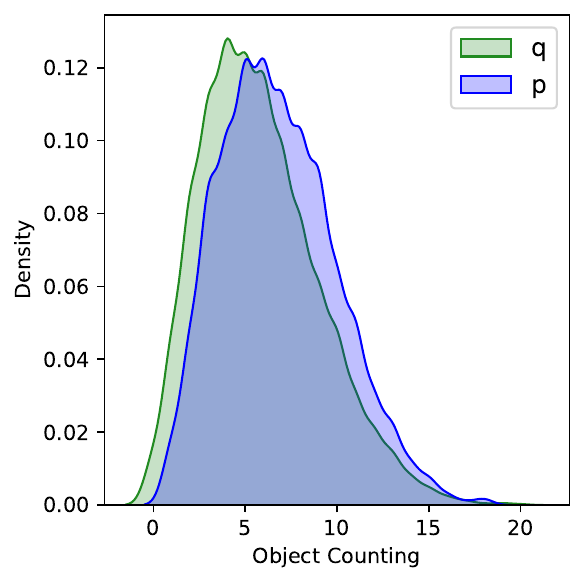}
    \caption{LLaVA-1.5-13b.}
\end{subfigure}
\end{figure}

\begin{figure}[ht!]    
\centering
\begin{subfigure}{0.32\linewidth}
    \includegraphics[width=\linewidth]{img/intro/counting_1p6_iter1.pdf}
    \caption{LLaVA-1.6-7b.}
\end{subfigure}
    \hfill
\begin{subfigure}{0.32\linewidth}
    \includegraphics[width=\linewidth]{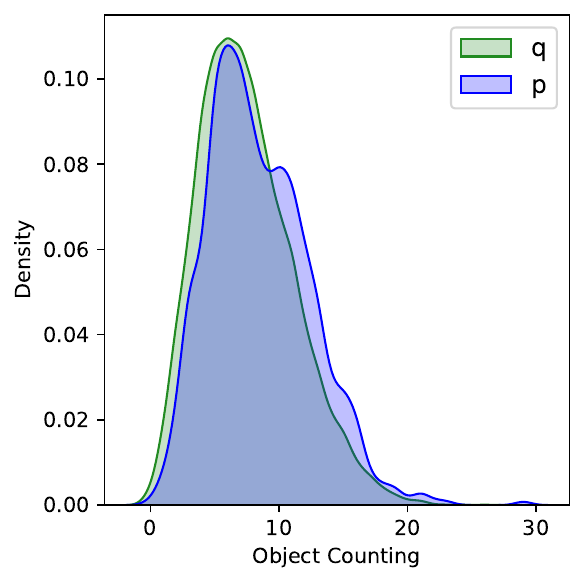}
    \caption{LLaVA-1.6-13b.}
\end{subfigure}
\end{figure}

\begin{figure}[ht!]  
\centering
\begin{subfigure}{0.32\linewidth}
    \includegraphics[width=\linewidth]{img/intro/counting_1p6_iter1.pdf}
    \caption{LLaVA-1.6-7b iteration 1.}
\end{subfigure}
    \hfill
\begin{subfigure}{0.32\linewidth}
    \includegraphics[width=\linewidth]{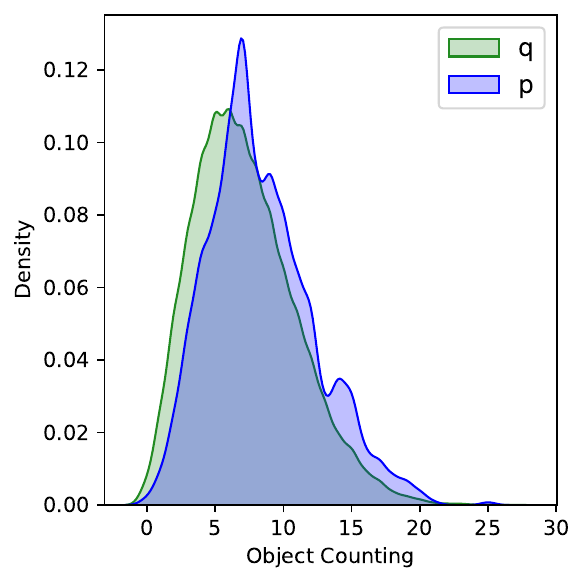}
    \caption{LLaVA-1.6-7b iteration 2.}
\end{subfigure}
\hfill
\begin{subfigure}{0.32\linewidth}
    \includegraphics[width=\linewidth]{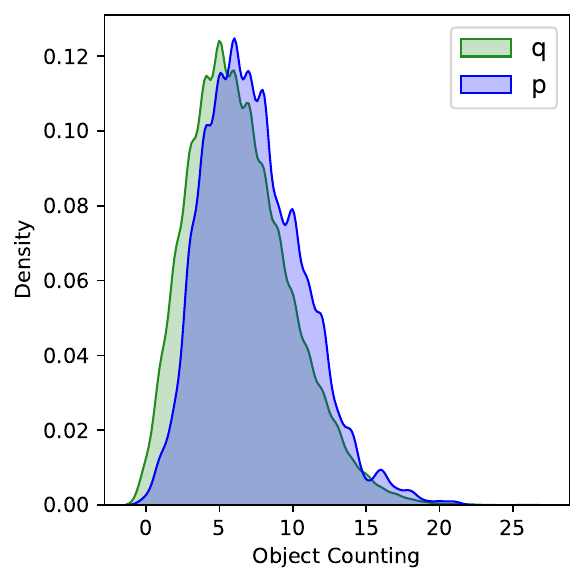}
    \caption{LLaVA-1.6-7b iteration 3.}
\end{subfigure}
\caption{Distribution of groundable objects in generated caption sampling the base model $p_{\theta}(\vz | \vc)$ and the grounded model $q(\vz | \vc, \vg)$. Models adapted with $\texttt{SVP}$ generate less groundable objects and have better object recall.}
\label{fig:object-recall-appx}
\end{figure}

\clearpage
\subsection{Score Ablation}
\label{appx:score-ablation}

\begin{figure}[ht!]   
\centering
\begin{subfigure}{0.48\linewidth}
    \centering
    \includegraphics[width=\linewidth]{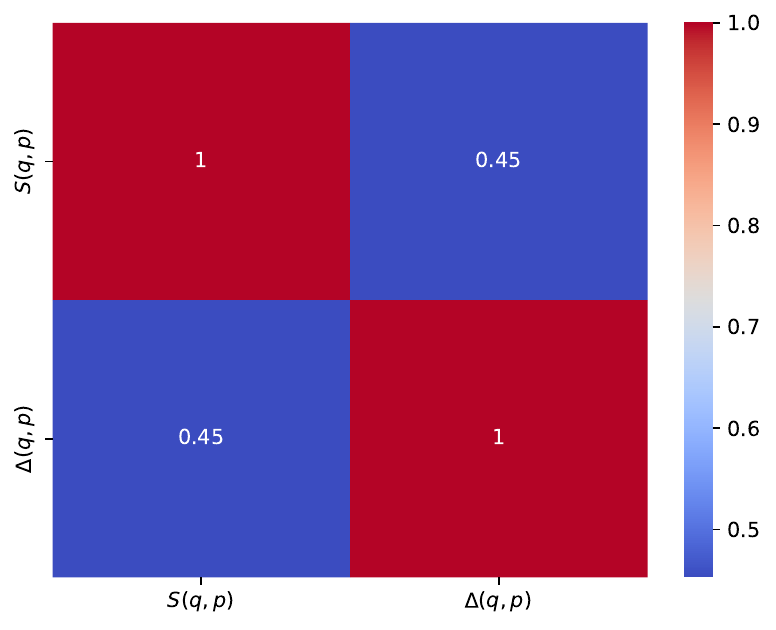}
    \caption{\textbf{Top1 Ranking Correlation} for weighted-difference $\Delta(q,p)$ and log-ratio $S(q,p)$ score using LLaVA-1.6-7b as base model.}
    \label{fig:score-corr-appx}
\end{subfigure}
    \hfill
\begin{subfigure}{0.35\linewidth}
    \includegraphics[width=\linewidth]{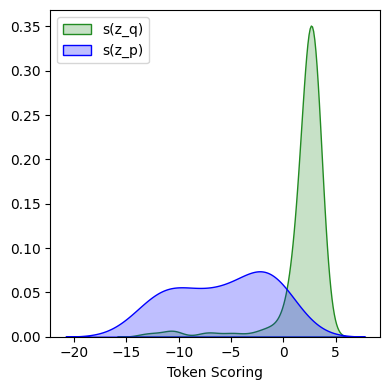}
    \caption{\textbf{Empirical Distribution} of sequence scores. Log-space representation of $S(q,p_{\theta})$ for sequence scoring. We see the scoring mechanism's effectiveness to differentiate between posterior samples $\vz_q$ (with grounding) and prior samples $\vz_p$ (without grounding).}
    \label{fig:token-scoring-appx}
\end{subfigure}
\end{figure}

\clearpage
\section{Training Objective Derivation}
\label{appx:objective}

We derive our visual-language alignment objective following two approaches: re-weighted maximum likelihood and greedy off-policy optimization. Assuming a deterministic output distribution $p(\vx | \vz, \vc) = d(\vz, \vc)$, we start with re-weighted maximum likelihood as a negated KL maximization:
\begin{equation}
\mathcal{F}_{\texttt{MLE}}(\vc; \theta) = 
    -\mathbb{KL}
    \left[q(\vz | \vc, \vg), p_{\theta}(\vz | \vc)
    \right] 
    = 
    \int q(\vz | \vc, \vg) \log p_{\theta}(\vz | \vc) d\vz
    - \int q(\vz | \vc, \vg) \log q(\vz | \vc, \vg) d\vz
\end{equation}

Taking the gradient with respect to $\theta$:
\begin{equation}
      \nabla_{\theta}\mathcal{F}_{\texttt{MLE}}(\vc; \theta) 
    = \nabla_{\theta} \int q(\vz | \vc, \vg) \log p_{\theta}(\vz | \vc) d\vz 
    = \int q(\vz | \vc, \vg) \nabla_{\theta} \log p_{\theta}(\vz | \vc) d\vz
\end{equation}

Approximating the expectation with $K$ samples from $\vz \sim q(\vz | \vc, \vg)$ and filtering using our scoring mechanism:
\begin{equation}
\nabla_{\theta} \mathcal{F}^{k(\vc)}_{\texttt{MLE}}(\vc; \theta) \approx
\dfrac{1}{|k(\vc)|}
\sum^{K}_{i=1}
\left[
\mathbbm{1}\{\vz^i : S(q(\vz^i | \vc, \vg), p_{\theta}(\vz^i | \vc)) \geq S_{k(\vc)}\} \nabla_{\theta} \log p_{\theta}(\vz^i | \vc)
\right]
\end{equation}

For the policy optimization approach, we begin with the standard on-policy REINFORCE estimator using our scoring mechanism $f(\vz)$ as reward:
\begin{equation}
    \mathcal{F}_{\texttt{RL-ON}}(\vc; \theta) = \bE_{p_{\theta}(\vz | \vc)}\left[f(\vz)\right] = \int p_{\theta}(\vz | \vc) f(\vz) d\vz
\end{equation}

The gradient for $\theta$ yields:
\begin{equation}
    \nabla_{\theta} \mathcal{F}_{\texttt{RL-ON}}(\vc; \theta) 
    = \nabla_{\theta} \int p_{\theta}(\vz | \vc) f(\vz) d\vz 
    = \int \nabla_{\theta} p_{\theta}(\vz | \vc) f(\vz) d\vz 
    = \int p_{\theta}(\vz | \vc)  \nabla_{\theta} \log p_{\theta}(\vz | \vc) f(\vz) d\vz. 
\end{equation}

To incorporate our guiding distribution $q$, we use importance sampling:
\begin{equation}
    \nabla_{\theta} \mathcal{F}^{q}_{\texttt{RL-ON}}(\vc; \theta) 
    = \int q(\vz | \vc, \vg) \dfrac{p_{\theta}(\vz | \vc)}{q(\vz | \vc, \vg)}  \nabla_{\theta} \log p_{\theta}(\vz | \vc) f(\vz) d\vz
\end{equation}

This is an unbiased estimator for the on-policy gradient leveraging the "off-policy" or behavioral/guiding distribution $q$. 
If now we approximating the expectation for $q$ with $K$ samples and filter using the score contained in $f(\vz)$, we can write:
\begin{equation}
    \nabla_{\theta} \mathcal{F}^{k(\vc)}_{\texttt{RL-OFF}}(\vc; \theta)  \approx
    \dfrac{1}{|k(\vc)|} \sum^{K}_{i=1} 
    \left[ \mathbbm{1}\{\vz^i : S(q(\vz^i | \vc, \vg), p_{\theta}(\vz^i | \vc)) \geq S_{k(\vc)}\}\right]
    \dfrac{p_{\theta}(\vz^i | \vc)}{q(\vz^i | \vc, \vg)}  \nabla_{\theta} \log p_{\theta}(\vz^i | \vc), 
\end{equation}
where we leverage the fact that $f(\vz^i) = \mathbbm{1}\{\vz^i : S(q(\vz^i | \vc, \vg), p_{\theta}(\vz^i | \vc)) \geq S_{k(\vc)}\}$. 

By construction we are only retaining samples with low importance ratio $p_{\theta}/q$. We are introducing bias focusing on samples that will improve vision-language alignment, and reducing the importance sampling estimator variance. 
Simplifying the previous gradient considering the importance ratio constant, we obtain the objective we maximize:
\begin{equation}
    \nabla_{\theta} \mathcal{\tilde F}^{k(\vc)}_{\texttt{RL-OFF}}(\vc; \theta)  \approx
    \dfrac{1}{|k(\vc)|} \sum^{K}_{i=1} 
    \left[ \mathbbm{1}\{\vz^i : S(q(\vz^i | \vc, \vg), p_{\theta}(\vz^i | \vc)) \geq S_{k(\vc)}\}\right]
    \nabla_{\theta} \log p_{\theta}(\vz^i | \vc)
\end{equation}

Both approaches yield equivalent gradients after approximations: $\nabla_{\theta} \mathcal{F}^{k(\vc)}_{\texttt{MLE}}(\vc; \theta) = \nabla_{\theta} \mathcal{\tilde F}^{k(\vc)}_{\texttt{RL-OFF}}(\vc; \theta)$. This equivalence provides a strong theoretical foundation for our method. We optimize this objective at each SVP iteration by averaging over a batch of visual inputs: $\mathcal{L}(\theta) = -1/|B| \sum^{C}_{c=1} \mathcal{F}(\vc; \theta)$.

\clearpage
\section{Prompting}
\label{appx:prompting}

\begin{tcolorbox}[
  width=\textwidth,
  colback=lightgray!10,  %
  colframe=Gray,        %
  title=System Prompt - Sampling,    %
  fonttitle=\bfseries    %
]
You are an AI visual-language assistant that can analyze images and helps writing detailed descriptions of images.

\texttt{<instruction>}

    Describe the scene and the objects in the image in details. 
    Describe the object attributes and positions. 
    Output only the descriptions of objects that are in the image.
    Use separate sentence for each object.

    Include details like object counts, position of the objects, relative position between the objects.

    Start your description with "In the image, ".

\texttt{</instruction>}
\end{tcolorbox}

\begin{tcolorbox}[
  width=\textwidth,
  colback=lightgray!10,  %
  colframe=Gray,        %
  title=System Prompt - Grounded Sampling,    %
  fonttitle=\bfseries    %
]
You are an AI visual-language assistant that can analyze images and helps writing detailed descriptions of images.

In addition, specific objects and object locations within the image are given, along with detailed coordinates inside \texttt{<context></context>}. 
These coordinates are in the form of bounding boxes, represented as \texttt{(x1, y1, x2, y2)} with floating numbers ranging from 0 to 1. 
These values correspond to the top left \texttt{x}, top left \texttt{y}, bottom right \texttt{x}, and bottom right \texttt{y}.

\texttt{<instruction>}

    Using the provided objects and bounding boxes inside \texttt{<context></context>}, describe the image. 
    
    Describe the scene and the objects in the image in details. 
    Describe the object attributes and positions. 
    Output only the descriptions of objects that are in the image.
    Use separate sentence for each object.

    Include details like object counts, position of the objects, relative position between the objects.
    
    \emph{Do not mention the bounding box coordinates. Utilize this data to explain the scene using natural language.}
    
    Start your description with "In the image, ".
    
\texttt{</instruction>}
\end{tcolorbox}

\begin{tcolorbox}[
  width=\textwidth,
  colback=lightgray!10,  %
  colframe=Gray,        %
  title=Base Prompt,    %
  fonttitle=\bfseries    %
]
Please generate a detailed and comprehensive description for the content of this image. Be precise.
\end{tcolorbox}

\begin{tcolorbox}[
  width=\textwidth,
  colback=lightgray!10,  %
  colframe=Gray,        %
  title=Grounded Prompt,    %
  fonttitle=\bfseries    %
]
\texttt{<context>}

\texttt{\{grounding\}}

\texttt{</context>}

Please generate a detailed and comprehensive description for the content of this image. Be precise.
\end{tcolorbox}

\clearpage
\section{Iterative Self-Improvement in Generative Models}
\label{sec:self-improvement}

Our method is inspired by recent advances in self-improvement in language modelling~\cite{zelikman2022star,dong2023raft,gulcehre2023reinforced}, where the focus has been in improving chain-of-though reasoning leveraging feedback. The Self-Taught Reasoner (STaR~\cite{zelikman2022star}) uses ground truth labels and rationalization as feedback~\cite{zelikman2022star} and fine-tunes the model on reasoning steps that generate the correct answer~\cite{hoffman2024training}. In~\ref{fig:star-prompt} we provide a comparison between \texttt{SVP} and rationalization in STaR, where the model tries to find the correct reasoning path given external feedback in the form or a ground truth response.

\begin{figure}[ht!]
\centering
    \begin{subfigure}[t]{0.47\linewidth} %
        \begin{tcolorbox}
        \begin{Verbatim}[commandchars=\\\{\}]
P: Where do you put your grapes
just before checking out?
Answer Choices:
(a) mouth
(b) grocery cart
(c) super market
(d) fruit basket
(e) fruit market
\textcolor{dartmouthgreen}{C: grocery cart (CORRECT)}
R: The answer should be the place
where grocery items are placed 
before checking out. 
Grocery cart makes the most 
sense for holding grocery items. 
        \end{Verbatim}
        \end{tcolorbox}
    \caption{Rationalization in STaR. We sample from $q(\vz | \vc, \vy)$, where $\vy$ is the ground truth label provided as context $C$. The response $\vz$ is then leveraged to build $p(\vy | \vc, \vz)$ or a deterministic decoding $d(\vc, \vz)$.}
    \label{fig:star}
    \end{subfigure}%
    \quad
    \begin{subfigure}[t]{0.47\linewidth} %
        \begin{tcolorbox}
        \begin{Verbatim}[commandchars=\\\{\}]
P: <IMAGE> Please describe 
the content of this image.
\textcolor{dartmouthgreen}{C:} 
\textcolor{dartmouthgreen}{lamp   [0.33,0.47,0.16,0.23],}
\textcolor{dartmouthgreen}{laptop [0.48,0.54,0.16,0.09],}
\textcolor{dartmouthgreen}{desk   [0.49,0.75,0.61,0.33],}
\textcolor{dartmouthgreen}{plant  [0.36,0.56,0.04,0.06],}
\textcolor{dartmouthgreen}{frame  [0.64,0.22,0.24,0.24],}
\textcolor{dartmouthgreen}{basket [0.80,0.80,0.15,0.15].}
R: The image depicts a modern 
work setup. On the wall, 
there is a framed artwork.
Below the frame, there is a desk. 
On the desk, there is a laptop. 
        \end{Verbatim}
        \end{tcolorbox}
    \caption{Grounded sampling in SVP. We sample from $q(\vz | \vc, \vg)$, where $\vg$ is the grounding information provided as context $C$. The response $\vz$ is then leveraged to build $p(\vx | \vc, \vz)$ or a deterministic decoding $d(\vc, \vz)$.
    }
    \label{fig:SVP}
    \end{subfigure}
    
    \caption{Comparison between rationalization in STaR~\cite{zelikman2022star} and posterior sampling in \texttt{SVP}.}
\label{fig:star-prompt}
\end{figure}

\clearpage

\section{DPO Derivation}
\label{appx:dpo}

The DPO loss comparing policy $\pi_{\theta}$ to reference $\pi_{\text{ref}}$ is:
\begin{equation}
\mathcal{L}_{\text{DPO}}(\pi_\theta, \pi_{\text{ref}}) = -\mathbb{E}_{(\vx,\vy_w,\vy_l)\sim \mathcal{D}}[\log\sigma(\beta~\delta r_{\theta})],
\end{equation}
where $\vx$ is the input prompt, $\vy_w$ and $\vy_l$ are preferred and dis-preferred responses, $\sigma(z)$ is the sigmoid function, and $\beta=1$ for simplicity. $\delta r_{\theta}$ represents the log-probability ratio difference between winning and losing samples:
\begin{equation}
\delta r_{\theta} = \log\frac{\pi_\theta(\vy_w|\vx)}{\pi_{\theta}(\vy_l|\vx)} = \log\frac{\pi_\theta(\vy_w|\vx)}{\pi_{\text{ref}}(\vy_w|\vx)} - \log\frac{\pi_\theta(\vy_l|\vx)}{\pi_{\text{ref}}(\vy_l|\vx)} = r_w(\theta) - r_l(\theta)
\end{equation}

\paragraph{Preference Feedback and Optimal Policy}
In determining the shape of DPO's implicit reward $r_{w}(\theta)$, we can draw insights from the standard PPO formulation used in RLHF. The RLHF framework integrates reinforcement learning with human preferences through three key components: 
(i) a reward model $s_{\psi}$ (typically parametric) that encodes human preference labels; 
(ii) a generative policy model $\pi_{\theta}$ that can be sampled and improved through reward feedback;
(iii) a reference model $\pi_{\texttt{ref}}$ that provides stability during learning. This framework is expressed mathematically as:
\begin{equation}
\mathcal{F}_{\texttt{PPO}} = \bE_{\pi_{\theta}(\vz | \vc)} \left[s_{\bar \psi}(\vz, \vc) - \gamma \log \dfrac{\pi_{\theta}(\vz | \vc)}{\pi_{\texttt{ref}}(\vz | \vc)}\right],
\end{equation}
where $\vz$ represents a textual continuation for a given prompt $\vc$ (either visual or textual). For this regularized policy optimization problem, it can be demonstrated that the optimal policy takes the form:
\begin{equation}
\pi^{*}_{\theta}(\vz | \vc) \propto \pi_{\texttt{ref}}(\vz | \vc) \exp \left(\dfrac{s_{\bar\psi}(\vz, \vc)}{\gamma}\right).
\end{equation}
When we isolate the reward term $s_{\bar \psi}$, we find that this formulation aligns with the DPO framework, differing only by constant terms.

\paragraph{Gradient Derivation}
Applying the chain rule to find $\nabla_\theta \mathcal{L}_{\text{DPO}}$:
\begin{equation}
- \nabla_\theta \mathcal{L}_{\text{DPO}} = \mathbb{E} \left[\dfrac{\sigma'(\delta r_{\theta})}{\sigma(\delta r_{\theta})} ~\nabla_{\theta} \delta r_{\theta}\right]
\end{equation}
Using $\sigma'(z) = \sigma(z)(1 - \sigma(z))$:
\begin{equation}
\nabla_\theta \mathcal{L}_{\text{DPO}} = -\mathbb{E}\left[(1-\sigma(\delta r_{\theta})) \cdot  \nabla_\theta \delta r_{\theta}\right]
\end{equation}
The gradient of $\delta r_{\theta}$ simplifies to:
\begin{align}
\nabla_\theta \delta r_{\theta} = \nabla_\theta \log\pi_\theta(\vy_w|\vx) - \nabla_\theta \log\pi_\theta(\vy_l|\vx)
\end{align}
The final DPO gradient is:
\begin{equation}
\nabla_\theta \mathcal{L}_{\text{DPO}} = 
-\beta~\mathbb{E}\left[(1-\sigma(\delta r_{\theta}))[\nabla_\theta \log\pi_\theta(\vy_w|\vx) - \nabla_\theta \log\pi_\theta(\vy_l|\vx)]\right]
\end{equation}
This formulation optimizes preferences using a re-weighted maximum likelihood without requiring a separate reward model or RL training.

\paragraph{Preference Optimization and Vision-Language Alignment}
The application of SVP for vision-language alignment shares conceptual similarities with DPO, though with notable distinctions. A key difference lies in our scoring mechanisms' approach to negative samples. While our scores can effectively guide ranking and top-K selection, a small difference between guided and prior distributions doesn't necessarily indicate an undesirable sample. Low scores may simply reflect cases where grounding information offers minimal improvement, such as when the base model's response is already grounded in the visual input, or the response does not require the visual information. Consequently, in SVP, the gradient information from negative samples provides limited value.

DPO employs a gradient re-weighting scheme based on how much a model's preference predictions deviate from ground truth. This acts as a soft sample selection mechanism, giving greater weight to more informative training examples. Our approach parallels this concept, though we implement a hard sample selection strategy that emphasizes examples offering greater vision-language alignment information.

\clearpage
\section{Qualitative Examples}
\label{appx:examples}

Here we provide additional visualizations and examples to illustrate our method. 
The images used for captioning and visual queries were selected from the \texttt{COCO2014} training set~\cite{lin2014microsoft}, while referring examples were sourced from the \texttt{RefCOCO} dataset~\cite{kazemzadeh2014referitgame}, specifically the \texttt{val\_lite} set~\cite{zhang2024lmms}.
The examples are presented with minimal post-processing, limited to the removal of special characters and correction of obvious grammatical errors. Both models were prompted using the protocols outlined in~\ref{appx:prompting}. All hyper-parameters remained consistent across both models. 
The examples compare output of LLaVA-1.6 models sampling from the base model (w/o \texttt{SVP}) and sampling from the adapted model (w/ \texttt{SVP}).
These comparisons provide qualitative insights into the performance differences between the base and adapted models.

\subsection{Captioning Tasks}

\begin{figure*}[ht!]
\centering
\begin{examplebox}[width=0.48\textwidth,box align=top,before=,after=\hfill]{Example A}
\begin{vlmbox}{}
\textbf{Instruction:} 
Describe the content of this image as detailed as possible.
    \begin{center}
    \includegraphics[width=\linewidth, keepaspectratio]{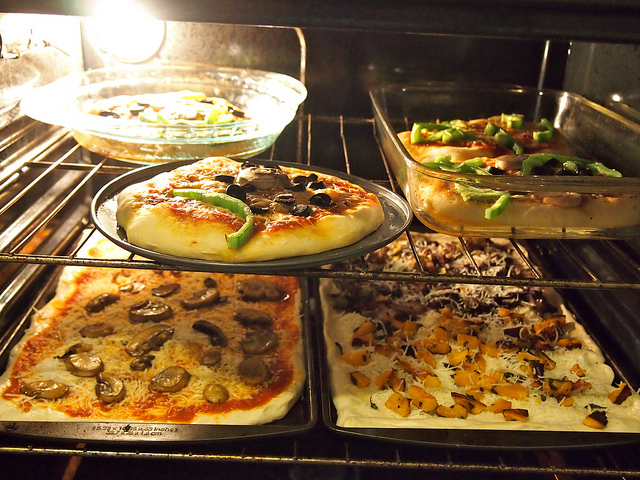}
    \end{center}
    
\tcblower

In the image, there are \hle{four pizzas} in an oven. 
The oven is filled with racks, each holding a pizza. 
The pizzas are of different types, with various toppings visible. 
The oven is illuminated, casting a warm glow on the pizzas. 
The pizzas are arranged in two rows, with \hle{two pizzas in each row}. 
The pizzas are cooking, as indicated by their position in the oven. 
The oven is a standard kitchen appliance, with a door that is open to reveal 
the pizzas inside. The pizzas are the main focus of the image, 
with their colors and textures contrasting against the dark interior of the oven. 
The image captures a moment of culinary preparation, 
with the pizzas in various stages of cooking.
\end{vlmbox}%
\begin{svpbox}{}
\tcblower

In the image, there is a well-lit oven with a black interior. 
Inside the oven, there are \hl{three pizzas on the top rack and two pizzas 
on the bottom rack}.
On the top rack, the first pizza is on the left side of the oven. 
It has a golden brown crust and is topped with green peppers and mushrooms. 
The second pizza is on the right side of the oven. 
It also has a golden brown crust and is topped with green peppers and black olives. 
On the bottom rack, the first pizza is on the left side of the oven. 
It has a golden brown crust and is topped with mushrooms. 
The second pizza is on the right side of the oven.

The light inside the oven illuminates the pizzas, highlighting their toppings and crusts. 
The pizzas appear to be freshly baked, as their crusts are golden brown and their toppings are evenly distributed. 
The overall scene suggests a home or restaurant kitchen where pizzas are being prepared.
\end{svpbox}
\end{examplebox}%
\begin{examplebox}[width=0.48\textwidth,box align=top,before=,after=\hfill]{Example B}
\begin{vlmbox}{}
\textbf{Instruction:} 
Describe the content of this image as detailed as possible.
    \begin{center}
    \includegraphics[width=\linewidth, keepaspectratio]{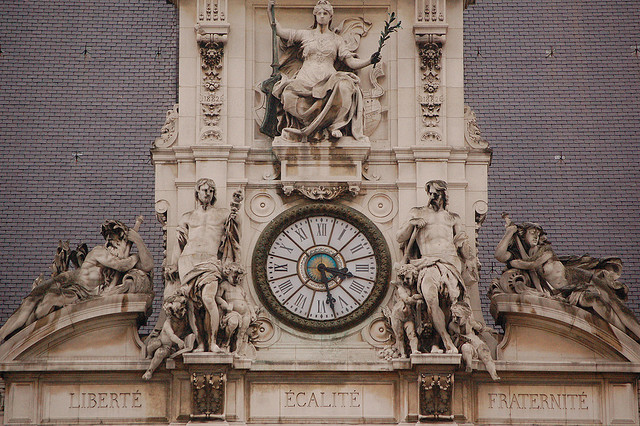}
    \end{center}
    
\tcblower

In the image, there is a large clock mounted on a building. 
The clock face is white with black numbers and hands. 
The building itself is adorned with intricate carvings and statues. 
There are three statues visible in the image. 
One statue is of a woman holding a shield, another is of a man holding a spear, 
and the third is of a man holding a scroll. 
The carvings and statues are detailed and appear to be made of stone. 
The building has a gray roof, and the clock is 
positioned in the center of the image. 
The statues are positioned on either side of the clock, 
with the woman on the left, the man with the spear in the middle, 
and the man with the scroll on the right. 
The overall scene is a blend of architectural and artistic elements.
\end{vlmbox}%
\begin{svpbox}{}
\tcblower

In the image, there is a grand clock tower that is ornate and intricate. 
The tower is made of stone and is topped with a statue of a woman holding a sword. 
This statue is located at the top of the tower, near the center. 
The tower also features a large clock face on the front, 
which is situated below the statue. 
The clock face is white with black numbers and hands. 
Below the clock face, there are two smaller statues of people. 
These statues are positioned on either side of the clock face, 
with the one on the left being slightly larger than the one on the right. 
The tower is set against a gray brick background, 
which contrasts with the white stone of the tower. 
The overall scene suggests a sense of grandeur and history.
\end{svpbox}
\end{examplebox}
\end{figure*}

\clearpage
\begin{figure*}
\centering
\begin{examplebox}[width=0.48\textwidth,box align=top,before=,after=\hfill]{Example C}
\begin{vlmbox}{}
\textbf{Instruction:} 
Describe the content of this image as detailed as possible.
    \begin{center}
    \includegraphics[width=\linewidth, keepaspectratio]{img/examples/COCO_train2014_000000193281.jpg}
    \end{center}
    
\tcblower

In the image, there are three objects: a pink frisbee, a pair of red sneakers, and \hl{a beer bottle}. The pink frisbee is located in the foreground, with its star-shaped design prominently displayed. The red sneakers are positioned in the background, with one shoe slightly in front of the other. The beer bottle is also in the background, standing upright and slightly \hle{to the right of the sneakers}. The objects are arranged in a way that suggests a casual, outdoor setting, possibly a park or a beach. The frisbee and sneakers are the main focus of the image, while the beer bottle adds a touch of leisure to the scene.

\end{vlmbox}%
\begin{svpbox}{}
\tcblower

In the image, there is a casual outdoor setting with a pair of \hl{red and white Adidas shoes}. The shoes are placed on a grassy area, with one shoe closer to the camera and the other slightly behind it. \hl{The Adidas logo is prominently displayed on both pairs of shoes}. 
A pink frisbee with a blue star design is also present in the scene. It is positioned in the foreground, with the shoes in the background. The frisbee appears to be resting on the grass, and its bright color stands out against the natural backdrop.The overall scene suggests a leisurely outdoor activity, possibly a game of frisbee, with the Adidas shoes indicating a sporty or active lifestyle. The grassy area and the casual arrangement of the shoes and frisbee give the image a relaxed and informal vibe.
\end{svpbox}
\end{examplebox}%
\begin{examplebox}[width=0.48\textwidth,box align=top,before=,after=\hfill]{Example D}
\begin{vlmbox}{}
\textbf{Instruction:} 
Describe the content of this image as detailed as possible.
    \begin{center}
    \includegraphics[width=\linewidth, keepaspectratio]{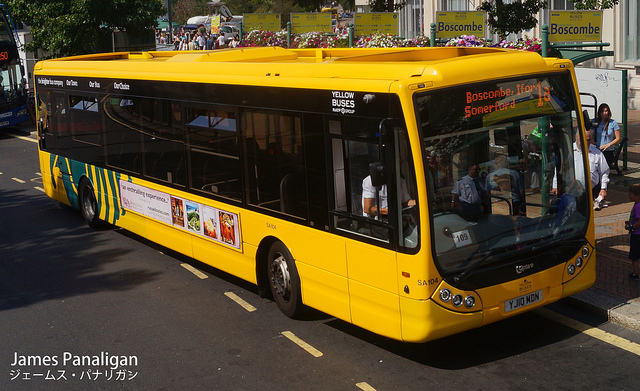}
    \end{center}
    
\tcblower

In the image, there is a \hl{large yellow bus} \hle{parked on the side of a street}. The bus is positioned on the right side of the image and is the most prominent object in the scene. \hle{It is a double-decker bus, which is characterized by its two levels of seating}. The bus is parked next to a sidewalk, which is visible on the right side of the image. There are several people in the image, some of whom are standing near the bus. One person is standing on the sidewalk, while others are standing on the street. \hle{There is also a car parked on the left side of the image}, which is partially visible. The bus is parked in front of a building, which is located on the left side of the image. The building is partially visible, with only the top part of it visible in the image. The bus is also parked near a tree, which is located on the left side of the image. The tree is partially visible, with only the top part of it visible in the image. The overall scene suggests a typical day in a city, with people going about their daily routines and public transportation being a common sight.
\end{vlmbox}%
\begin{svpbox}{}
\tcblower

In the image, a \hl{large yellow bus} dominates the scene, traveling down a city street. The bus is vibrant and appears to be in motion, heading towards the right side of the frame. Its size and bright color make it the main focus of the image. \hl{In the background, another bus can be seen, albeit much smaller and less prominent}. It's positioned on the left side of the image, providing a sense of depth and perspective to the scene. On the right side of the image, there's a large advertisement. It's not clear what the advertisement is for, but its size and placement suggest it's a significant feature of the urban landscape. The setting appears to be a typical city street. Trees can be seen in the background, adding a touch of nature to the urban environment. There are also several people scattered throughout the image, going about their day, which contributes to the everyday, relaxed atmosphere of the scene. The overall composition of the image suggests a bustling city life, with public transportation playing a significant role. The buses, people, and advertisement all come together to paint a picture of a vibrant, active cityscape.
\end{svpbox}
\end{examplebox}
\end{figure*}

\clearpage
\begin{figure*}
\centering
\begin{examplebox}[width=0.48\textwidth,box align=top,before=,after=\hfill]{Example C}
\begin{vlmbox}{}
\textbf{Instruction:} 
Describe the content of this image as detailed as possible.
    \begin{center}
    \includegraphics[width=\linewidth, keepaspectratio]{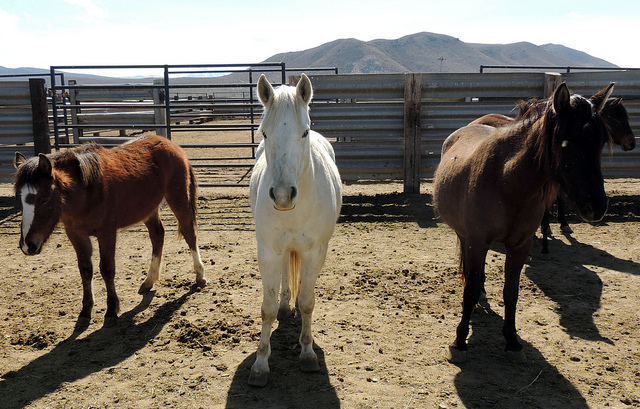}
    \end{center}
    
\tcblower

In the image, there are three horses standing in a dirt enclosure. 
The horse on the left is brown with a white blaze on its face. 
It is standing with its head turned to the side, \hle{looking towards the camera}. 
The horse in the middle is white with a brown spot on its head. 
It is standing with its \hle{head turned to the side}, looking towards the camera. 
The horse on the right is brown with a white blaze on its face. 
It is standing with its head turned to the side, \hle{looking towards the camera}. 
The enclosure is made of metal fencing and there is a mountain visible 
in the background.
\end{vlmbox}%
\begin{svpbox}{}
\tcblower

In the image, there are three horses standing in a dirt enclosure. 
The horse on the left is brown with a white stripe running down its nose. 
Its head is \hl{slightly turned to the left}. 
The horse in the middle is white and stands with its head facing forward. 
The horse on the right is black and its head is turned to the right. 
The three horses are standing next to each other, separated by a metal fence. 
In the background, there is a clear blue sky and a mountain range. 
The overall scene is a serene outdoor setting.
\end{svpbox}
\end{examplebox}%
\begin{examplebox}[width=0.48\textwidth,box align=top,before=,after=\hfill]{Example D}
\begin{vlmbox}{}
\textbf{Instruction:} 
Describe the content of this image as detailed as possible.
    \begin{center}
    \includegraphics[width=\linewidth, keepaspectratio]{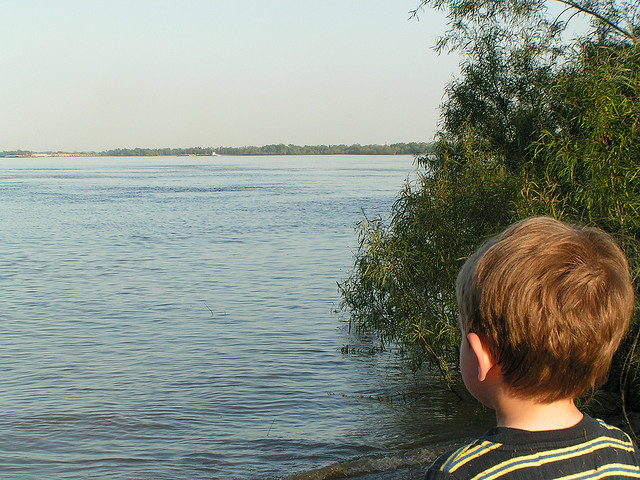}
    \end{center}
    
\tcblower

In the image, there is a young boy standing on the shore of a body of water. 
The boy is wearing a striped shirt and has blonde hair. 
He is looking out towards the water, which is calm and reflects the light. 
In the distance, there is a boat on the water, and the horizon is visible. 
The boy is standing in \hle{front of a tree with green leaves}, 
which is located on the shore. 
The overall scene suggests a peaceful day by the water.
\end{vlmbox}%
\begin{svpbox}{}
\tcblower

In the image, a young boy is standing by a body of water. 
He is wearing a black striped shirt. 
His back is turned to the camera, and he is looking out at the water. 
The boy appears to be standing on a grassy area near the water's edge. 
The water in the image is calm and appears to be a lake or a large river. 
It is a clear day, and the sky is visible in the background. 
There are no other people or objects in the image. 
The boy is the only person present. 
The relative position of the boy to the water is such that \hl{he is standing on the right side of the image}. 
He is standing closer to the foreground than the water, which is further back in the image. 
The boy is also standing to the \hl{right of the water}.
\end{svpbox}
\end{examplebox}
\end{figure*}

\clearpage
\begin{figure*}
\centering
\begin{examplebox}[width=0.48\textwidth,box align=top,before=,after=\hfill]{Example C}
\begin{vlmbox}{}
\textbf{Instruction:} 
Describe the content of this image as detailed as possible.
    \begin{center}
    \includegraphics[width=\linewidth, keepaspectratio]{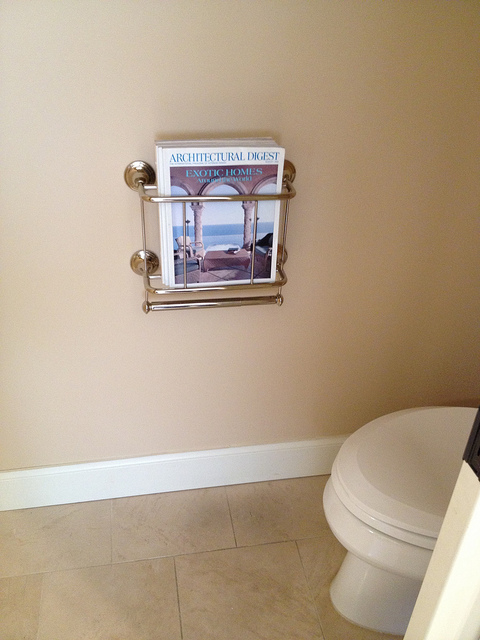}
    \end{center}
    
\tcblower

In the image, there is a bathroom setting. 
The main object is a white toilet, which is positioned in the lower right 
corner of the image. 
The toilet is situated on a beige tiled floor. 
Above the toilet, there is a metal rack mounted on the wall. 
The rack holds \hle{a single magazine}, which is titled "Architectural Digest". 
The magazine is displayed prominently on the rack, 
indicating that it is the main focus of the rack's purpose. 
The overall scene suggests a well-organized and aesthetically 
pleasing bathroom space, with the magazine rack providing a functional and 
decorative element. The white toilet contrasts with the beige tiles, 
creating a clean and modern look. The magazine on the rack adds a touch 
of sophistication and suggests an interest in architecture or design.
\end{vlmbox}%
\begin{svpbox}{}
\tcblower

In the image, there is a bathroom setting with a \hl{white toilet located towards the right side of the frame}. 
Above the toilet, there is a magazine rack mounted on the wall. 
The rack holds \hle{two magazines}, one titled \hl{"ARCHITECTURAL DIGEST"} and 
the other \hl{"EXOTIC HOMES"}. 
The "ARCHITECTURAL DIGEST" magazine 
is positioned slightly higher than the "EXOTIC HOMES" magazine, 
and both are centrally located within the rack. The "EXOTIC HOMES" magazine 
is situated directly below the "ARCHITECTURAL DIGEST" magazine. 
The bathroom walls are beige, and the overall scene suggests a quiet and 
relaxed atmosphere.
\end{svpbox}
\end{examplebox}%
\begin{examplebox}[width=0.48\textwidth,box align=top,before=,after=\hfill]{Example D}
\begin{vlmbox}{}
\textbf{Instruction:} 
Describe the content of this image as detailed as possible.
    \begin{center}
    \includegraphics[width=\linewidth, keepaspectratio]{img/examples/COCO_train2014_000000437510.jpg}
    \end{center}
    
\tcblower

In the image, there are \hle{three giraffes} standing in a field. 
The giraffe on the \hl{left is bending its neck down towards the ground}, 
possibly to reach for food or water. 
The giraffe in the middle is standing upright, with its \hl{neck extended upwards}. 
The giraffe on the right is also standing upright, but its \hle{neck is bent down towards
 the ground}. All three giraffes are facing the same direction, 
 suggesting they might be looking at something in the distance. 
 The field they are standing in appears to be dry and dusty, 
 with no visible vegetation. \hle{There are no other objects or animals in the image}.
\end{vlmbox}%
\begin{svpbox}{}
\tcblower

In the image, there are \hle{three giraffes} standing in a field. 
The giraffes are tall and majestic, with their long necks reaching upwards. 
The giraffe on the left is \hl{slightly bending its neck, possibly to reach for something}. 
The giraffe in the middle is standing upright, 
and the giraffe on the right is also bending its neck. 
In the \hl{center of the image, there is a tall pole}. 
The pole has a \hl{feeder attached to it}, which is likely providing food for the giraffes. 
The pole is surrounded by the field, which appears to be dry and brown.
The giraffes are spread out across the image, 
with the giraffe on the left being closer to the pole than the other two. 
The giraffe on the \hl{right is the furthest away from the pole}. 
The giraffes are the main focus of the image, with their long necks and distinctive 
patterns standing out against the backdrop of the field.

\end{svpbox}
\end{examplebox}
\end{figure*}

\clearpage
\subsection{Referring Tasks}

\begin{figure*}[ht!]
\centering
\begin{examplebox}[width=0.48\textwidth,box align=top,before=,after=\hfill]{Example C}
\begin{vlmbox}{}
\textbf{Instruction:} 
Provide a short description for this region.
    \begin{center}
    \includegraphics[width=\linewidth, keepaspectratio]{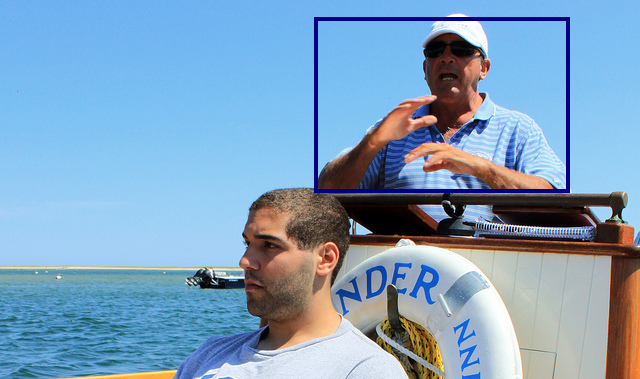}
    \end{center}
    
\tcblower

\hle{The image shows a man sitting in the foreground, looking to the side with a serious expression.} Behind him, there is another man standing on a boat, gesturing with his hands. \hle{The boat has a lifebuoy with the word "UNDER" on it}, and the man standing is wearing a white cap and sunglasses. \hle{The background features a clear blue sky and calm water, with a few boats visible in the distance. The overall setting suggests a leisurely day out on the water.}
\end{vlmbox}%
\begin{svpbox}{}
\tcblower

\hl{A man wearing a white hat and sunglasses.}
\end{svpbox}
\end{examplebox}%
\begin{examplebox}[width=0.48\textwidth,box align=top,before=,after=\hfill]{Example D}
\begin{vlmbox}{}
\textbf{Instruction:} 
Provide a short description for this region.
    \begin{center}
    \includegraphics[width=\linewidth, keepaspectratio]{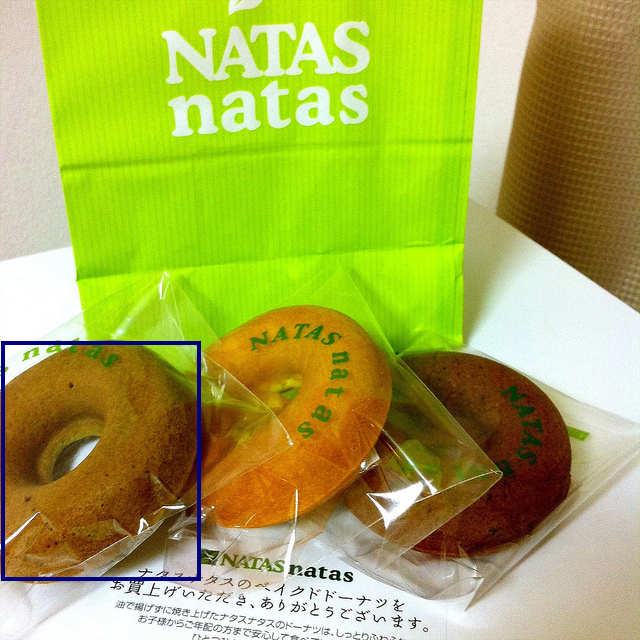}
    \end{center}
    
\tcblower

The image shows a bag of donuts with the brand name "NATAS" printed on it. The donuts are visible through the clear plastic packaging, and there are \hle{three donuts in total. The donuts appear to be of a similar type, possibly a ring donut, and they are placed on a white surface. The background is blurred, but there is a hint of a beige or light brown object to the right of the bag. The text on the bag is in Japanese, and the donuts have a hole in the center, which is typical for ring donuts.}
\end{vlmbox}%
\begin{svpbox}{}
\tcblower

\hl{A donut in a bag.}
\end{svpbox}
\end{examplebox}
\end{figure*}

\begin{figure*}
\centering
\begin{examplebox}[width=0.48\textwidth,box align=top,before=,after=\hfill]{Example C}
\begin{vlmbox}{}
\textbf{Instruction:} 
Provide a short description for this region.
    \begin{center}
    \includegraphics[width=\linewidth, keepaspectratio]{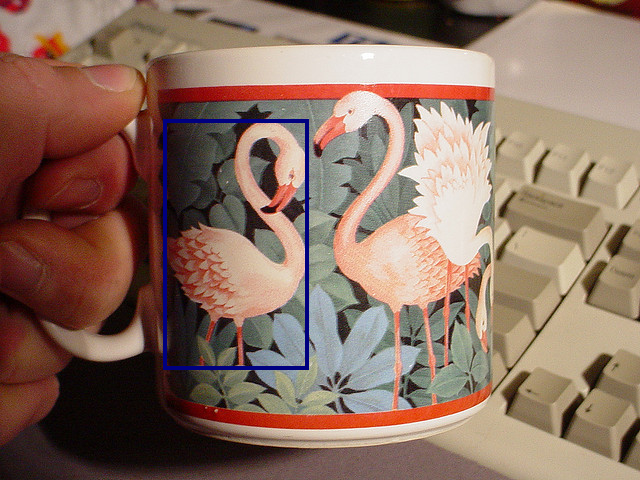}
    \end{center}
    
\tcblower

\hle{The image shows a person holding} a mug with a flamingo design. \hle{The mug features two flamingos}, one pink and one white, surrounded by green foliage and blue flowers. The design is colorful and playful, with the flamingos depicted in a stylized manner. \hle{The mug has a white handle and is placed in front of a computer keyboard, suggesting that the person might be using the computer while enjoying their drink. The background is blurred, but it appears to be an indoor setting with a desk or table.}
\end{vlmbox}%
\begin{svpbox}{}
\tcblower

A white coffee mug with \hl{a flamingo design.}
\end{svpbox}
\end{examplebox}%
\begin{examplebox}[width=0.48\textwidth,box align=top,before=,after=\hfill]{Example D}
\begin{vlmbox}{}
\textbf{Instruction:} 
Provide a short description for this region.
    \begin{center}
    \includegraphics[width=\linewidth, keepaspectratio]{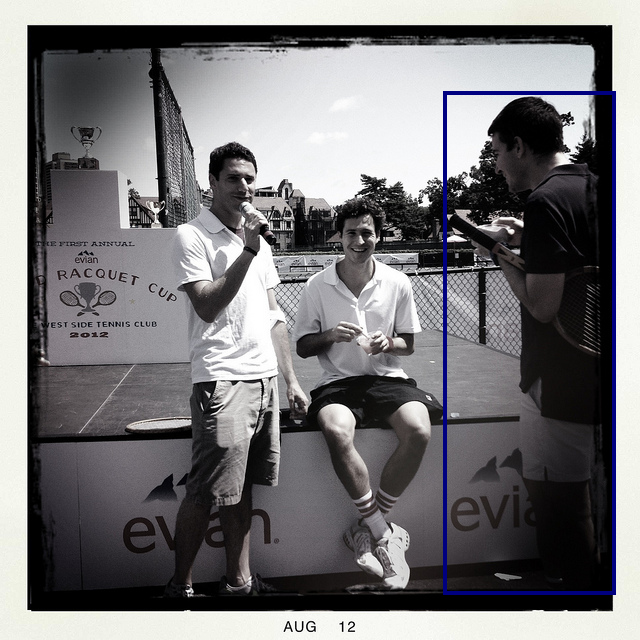}
    \end{center}
    
\tcblower

The image is a black and white photograph capturing a moment on a tennis court. In the foreground, there are two men standing on the court. \hle{The man on the left is holding a tennis racket} and appears to be speaking into a microphone, suggesting he might be a commentator or announcer. \hle{He is wearing a white shirt and shorts, which is typical attire for a tennis player or official}. \hle{The man on the right is seated on the edge of the court, holding a tennis racket and smiling}, possibly a player or a coach. He is wearing a white shirt and black shorts. \hle{In the background, there is a fence separating the court from the spectator area, and beyond the fence, there are trees and buildings, indicating that the court is part of a larger sports complex or club}. The sky is clear, suggesting it might be a sunny day. The image has a vintage or retro feel due to the black and white filter and the style of the clothing and the court's design. The text "THE FIRST ANNUAL WEST SIDE TENNIS CLUB" and "EVIAN RACQUET CUP 2012" are visible, indicating that this is an event sponsored by Evian, a brand known for its water products. The date "AUG 12" is also visible, which is likely the date of the event.The overall atmosphere of the image is casual and relaxed, with the focus on the tennis court and the people involved in the event.
\end{vlmbox}%
\begin{svpbox}{}
\tcblower

\hl{A man holding a tennis racket.}
\end{svpbox}
\end{examplebox}
\end{figure*}

\begin{figure*}
\centering
\begin{examplebox}[width=0.48\textwidth,box align=top,before=,after=\hfill]{Example C}
\begin{vlmbox}{}
\textbf{Instruction:} 
Provide a short description for this region.
    \begin{center}
    \includegraphics[width=\linewidth, keepaspectratio]{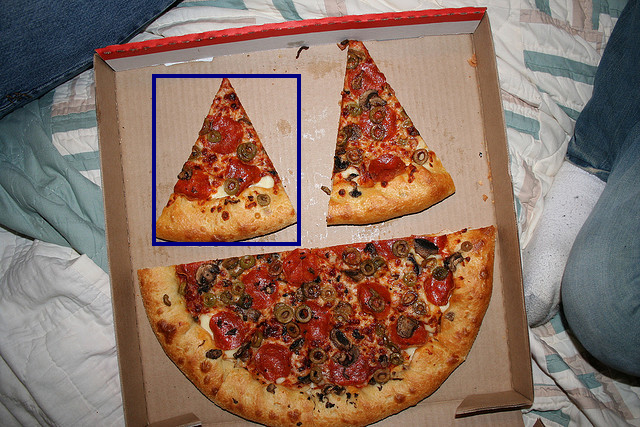}
    \end{center}
    
\tcblower

\hle{The pizza is in a box.}
\end{vlmbox}%
\begin{svpbox}{}
\tcblower

\hl{A slice of pizza.}
\end{svpbox}
\end{examplebox}%
\begin{examplebox}[width=0.48\textwidth,box align=top,before=,after=\hfill]{Example D}
\begin{vlmbox}{}
\textbf{Instruction:} 
Provide a short description for this region.
    \begin{center}
    \includegraphics[width=\linewidth, keepaspectratio]{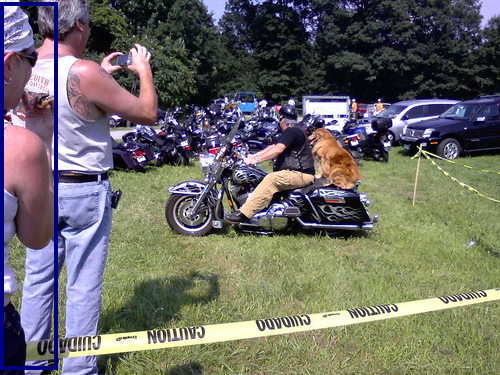}
    \end{center}
    
\tcblower

\hle{This is a photo of a man riding a motorcycle with a dog sitting on the back}. The man is wearing a black helmet and a black vest, and the dog appears to be a golden retriever. \hle{The man is also wearing a black tank top and blue jeans}. The motorcycle is black and has a large engine. The setting appears to be an outdoor event or gathering, as there are other motorcycles parked in the background and a yellow police tape is visible in the foreground. The man is taking a photo of the scene with his cell phone.
\end{vlmbox}%
\begin{svpbox}{}
\tcblower

\hle{A man taking a picture} of a dog sitting on a motorcycle.
\end{svpbox}
\end{examplebox}
\end{figure*}

\clearpage
\subsection{Visual Queries}

\begin{figure*}[ht!]
\centering
\begin{contextbox}[width=0.48\textwidth,box align=top,before=,after=\hfill]{Visual Input and Text Instruction.}
\begin{promptbox}{}
\textbf{Instruction}: What activity could potentially be enjoyed near the area shown in the image? 

    \begin{center}
    \includegraphics[width = \linewidth, keepaspectratio]{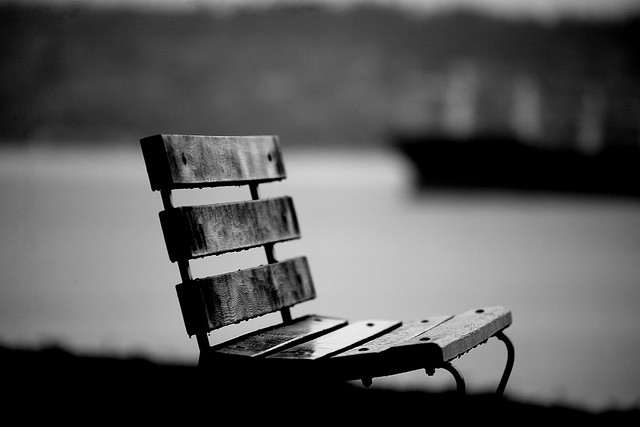}
    \end{center}
    
\tcblower

The image shows a wooden bench situated near a body of water, possibly a lake or a river. Given the serene and peaceful setting, one could potentially enjoy various activities such as:
Sitting on the bench to relax, read a book, or simply enjoy the view of the water and the surrounding landscape.
Taking a leisurely walk or jog along the water's edge, enjoying the tranquility and the natural beauty of the area.
Engaging in photography, capturing the picturesque scenery and the reflections on the water's surface.
Bird watching, as the area might be home to various bird species.
Fishing, if the water body is a lake or a river with fish populations.
Picnicking with friends or family, enjoying a meal outdoors in the fresh air and the calming presence of the water.
These activities are suitable for the setting depicted in the image, offering a chance to unwind and connect with nature.

\end{promptbox}
\end{contextbox}%
\begin{contextbox}[width=0.48\textwidth,box align=top,before=,after=\hfill]{Visual Input and Text Instruction.}
\begin{promptbox}{}
\textbf{Instruction}: What are some factors contributing to the difficulty of this activity?

    \begin{center}
    \includegraphics[width = \linewidth, keepaspectratio]{img/examples/COCO_train2014_000000522534.jpg}
    \end{center}
    
\tcblower

The image shows a person skiing on a steep, snow-covered slope. The difficulty of this activity can be attributed to several factors:
Terrain: The steepness of the slope increases the challenge of maintaining balance and control while skiing. Steeper slopes require more advanced skiing techniques and greater physical strength.
Snow Conditions: The type of snow can greatly affect the difficulty of skiing. In the image, the snow appears to be well-packed, which can make it harder to carve turns and maintain speed.
Visibility: The image shows a cloudy day, which can reduce visibility and make it more difficult to navigate the slope.
Equipment: The skier is wearing a backpack, which can add extra weight and affect balance and maneuverability.
Weather: The cloudy day suggests that the weather conditions might be challenging, with the possibility of poor visibility, cold temperatures, or even snowfall.
Skill Level: The skier's skill level plays a significant role in the difficulty of the activity. More advanced skiers are better equipped to handle steep slopes and challenging

\end{promptbox}
\end{contextbox}

\end{figure*}

\clearpage
\section{Details}
\label{appx:details}

\subsection{Datasets}

\begin{table}[ht!]
    \centering
    \caption{Datasets utilized in \texttt{SVP}. We use COCO2014 images as conditioning for building visual projections, using self-captioning and grounding feedback. 
    VP: visual projection.
    VQA: visual question answering. REG: referring expression generation.
    We use \texttt{lmms-eval}~\cite{zhang2024lmms} for all the evaluations. The \texttt{lite} splits as proposed in~\cite{zhang2024lmms}.
    }
    \begin{tabular}{l ccc}
    \toprule
    Dataset & Task & Split & $N$ \\
    \midrule
    \emph{Sampling} \\
    COCO2014~\cite{lin2014microsoft} & VP & \texttt{train} & 100:10000\\
    \midrule
    \\
    \emph{Evaluation} \\
    ScienceQA~\cite{saikh2022scienceqa}       & VQA & \texttt{test} & 4241 \\
    GQA~\cite{hudson2019gqa}                  & VQA &   \texttt{lite} & 500 \\
    \midrule
    COCO2017~\cite{lin2014microsoft}          & Captioning &   \texttt{val\_lite} & 500\\
    Flickr30k~\cite{plummer2015flickr30k}     & Captioning &  \texttt{test\_lite} & 500\\
    NoCaps~\cite{agrawal2019nocaps}           & Captioning &  \texttt{val\_lite} & 500 \\
    \midrule
    COCO2014~\cite{lin2014microsoft}      & Captioning & \texttt{val}   & 40504 \\
    COCO2017~\cite{lin2014microsoft}      & Captioning & \texttt{val}   & 5000  \\
    Flickr30k~\cite{plummer2015flickr30k} & Captioning &  \texttt{test} & 31783 \\
    NoCaps~\cite{agrawal2019nocaps}       & Captioning &  \texttt{val}  & 4500  \\
    \midrule
    RefCOCO~\cite{kazemzadeh2014referitgame}  & REG &  \texttt{val\_lite} & 500 \\
    \midrule
    RefCOCO~\cite{kazemzadeh2014referitgame}  & REG & \texttt{val} & 8811   \\
    RefCOCO~\cite{kazemzadeh2014referitgame}  & REG & \texttt{test} & 5000   \\
    RefCOCO~\cite{kazemzadeh2014referitgame}  & REG & \texttt{testA} & 1975  \\
    RefCOCO~\cite{kazemzadeh2014referitgame}  & REG & \texttt{testB} & 1810  \\
    RefCOCO+~\cite{kazemzadeh2014referitgame} & REG & \texttt{val}  & 3805  \\
    RefCOCO+~\cite{kazemzadeh2014referitgame} & REG & \texttt{testA} & 1975 \\
    RefCOCO+~\cite{kazemzadeh2014referitgame} & REG & \texttt{testB} & 1798 \\
    RefCOCOg~\cite{kazemzadeh2014referitgame} & REG & \texttt{val} & 7573   \\
    RefCOCOg~\cite{kazemzadeh2014referitgame} & REG & \texttt{test} & 5023  \\
    \midrule
    MMBench~\cite{liu2025mmbench}             & Multitasking   & \texttt{en\_dev\_lite} & 500\\
    MMMU~\cite{yue2024mmmu}                   & Multitasking   & \texttt{val} & 900\\
    \midrule
    POPE~\cite{li2023evaluating}              & Hallucinations & \texttt{adv} & 3000\\
    POPE~\cite{li2023evaluating}              & Hallucinations & \texttt{pop} & 3000\\
    POPE~\cite{li2023evaluating}              & Hallucinations & \texttt{random} & 3000\\
    \bottomrule
    \end{tabular}
    \label{tab:datasets}
\end{table}

\clearpage
\subsection{Experiments}
\begin{table}[ht!]
    \centering
    \caption{Hyper-parameters for the main experiments.}
    \resizebox{\textwidth}{!}{%
    \begin{tabular}{l c c c c c}
    \toprule
    & \texttt{LLaVA-1.5-13b} & \texttt{LLaVA-1.6-7b} & \texttt{LLaVA-1.6-13b} & \texttt{LLaVA-OV-0.5b} & \texttt{LLaVA-OV-7b} \\
    \midrule
    \emph{Sampling} \\
    images   & 1000 & 1000 & 1000 & 2000 & 2000     \\
    iterations   &  1 &  1 &  1 &  1 &  1           \\
    prompt-version  & \texttt{llava\_v1} & \texttt{mistral\_instruct} & \texttt{llava\_v1}  & \texttt{qwen\_1\_5} & \texttt{qwen\_1\_5} \\ 
    sample-batch & 20 & 20 & 20 & 10 & 10 \\
    samples/image   &  20 & 20 & 20 & 10 & 10       \\
    top $k$         & 0.2 & 0.2 & 0.2 & 0.1  & 0.1  \\
    \midrule
    \emph{Training} \\
    accelerators & A100 & A100 & A100 & A100 & A100 \\
    deepspeed & w/ ZeRO-2 &  w/ ZeRO-3 &  w/ ZeRO-3 &  w/ ZeRO-3 &  w/ ZeRO-3 \\
    epochs  & 1    & 1    & 1    & 3    & 3    \\
    grad-acc & 1 & 1 & 1 & 2 & 2 \\
    learning-rate      & $2e^{-4}$ & $2e^{-4}$ & $2e^{-4}$ & $1e^{-5}$ & $1e^{-5}$ \\
    lora    & w/   & w/   & w/   & w/ and w/o & w/ and w/o \\  
    lora-$\alpha$  & 256 & 16 & 256 & 16 & 16 \\
    lora-r  & 128 & 64 & 128 & 64 & 64 \\
    lr-schedule & cos & cos & cos & cos & cos \\
    max-tokens & 2048 & 2048 & 2048 & 1024 & 1024 \\
    mix-precision & w/ & w/ & w/ & w/ & w/ \\
    optimizer & AdamW & AdamW & AdamW & AdamW & AdamW \\
    samples & 4000:8000 & 4000:8000 & 4000:8000 & 2000 & 2000 \\ 
    text-encoder & \texttt{Vicuna-13b-v1.5} & \texttt{Mistral-7b-Instruct-v0.2} & \texttt{Vicuna-13b-v1.5} & 
    \texttt{Qwen2-0.5b} & \texttt{Qwen2-7b} \\
    train-batch & 16 & 16 & 16 & 4 & 4\\
    vision-encoder & \texttt{CLIP-L/14} & \texttt{CLIP-L/14} & \texttt{CLIP-L/14} & \texttt{SigLIP-SO/14} & 
    \texttt{SigLIP-SO/14} \\
    warm-up-rate  & 0.03 & 0.03 & 0.03 & 0.03 & 0.03 \\
    \bottomrule
    \end{tabular}
    }
    \label{tab:settings}
\end{table}

\end{document}